\documentclass[journal]{IEEEtran}

\usepackage{times}
\usepackage{epsfig}
\usepackage{graphicx}
\usepackage{amsmath}
\usepackage{amssymb}
\usepackage{algorithm}
\usepackage{algcompatible}
\usepackage{url}

\DeclareMathAlphabet{\mathcal}{OMS}{cmsy}{m}{n}

\usepackage{amsthm}
\usepackage{bm} 
\usepackage{amsfonts}
\usepackage{latexsym}
\usepackage{multirow}
\usepackage{subfigure}
\usepackage[T1]{fontenc} 
\usepackage{graphicx} 
\usepackage{mathptmx} 
\usepackage{subfigure}
\usepackage{fancyhdr}
\usepackage{float}
\usepackage{color}
\usepackage{hhline}
\usepackage{epstopdf}
\usepackage{ctable}
\usepackage[percent]{overpic}

\usepackage{rotating}

\newcommand{\x}{\textbf{x}}
\newcommand{\y}{\textbf{y}}
\newcommand{\z}{\textbf{z}}

\newcommand{\h}{\textbf{h}}
\newcommand{\n}{\textbf{n}}
\newcommand{\X}{\textbf{X}}
\newcommand{\bL}{\textbf{L}}
\usepackage{calligra}
\DeclareMathAlphabet{\mathcalligra}{T1}{calligra}{m}{n}

\newcommand{\newdeltax}{\delta_{\textbf{x}}}
\newcommand{\Cx}{\textbf{C}_{\textbf{x}}}
\newcommand{\Ch}{\textbf{C}_{\textbf{z}}}

\newcommand{\newgamma}{\bm{\gamma}}
\newcommand{\newH}{\textbf{H}}
\newcommand{\newlambda}{\bm{\lambda}}
\newcommand{\newLambda}{\bm{\Lambda}}
\newcommand{\newxi}{\xi}

\definecolor{hotpink}{rgb}{1.0, 0.41, 0.71}

\algnewcommand\INITIALIZATION{\item[\textbf{Initialization.}]}%
\algnewcommand\ITERATION{\item[\textbf{Iterative steps.}]}%

\begin{document}
\graphicspath{figures}

\title{Unrolled Variational Bayesian Algorithm for Image Blind Deconvolution}

\author{Yunshi Huang,
        Emilie Chouzenoux,~\IEEEmembership{Senior Member, IEEE}
        and~Jean-Christophe Pesquet,~\IEEEmembership{Fellow,~IEEE}
				\thanks{The authors acknowledge support from the \emph{Agence Nationale de la Recherche} of France under MAJIC (ANR-17-CE40-0004-01) project, and from the European Research Council Starting Grant MAJORIS ERC-2019-STG-850925.}
				}

\markboth{Journal of \LaTeX\ Class Files,~Vol.~XX, No.~X, Month~Year}%
{Shell \MakeLowercase{\textit{et al.}}: Bare Demo of IEEEtran.cls for IEEE Journals}

\maketitle

\begin{abstract}
In this paper, we introduce a variational Bayesian algorithm (VBA) for image blind deconvolution. Our generic framework incorporates smoothness priors on the unknown blur/image and possible affine constraints (e.g., sum to one) on the blur kernel. One of our main contributions is the integration of VBA within a neural network paradigm, following an unrolling methodology. The proposed architecture is trained in a supervised fashion, which allows us to optimally set two key hyperparameters of the VBA model and lead to further improvements in terms of resulting visual quality. Various experiments involving grayscale/color images and diverse kernel shapes, are performed. The numerical examples illustrate the high performance of our approach when compared to state-of-the-art techniques based on optimization, Bayesian estimation, or deep learning. \end{abstract}


\IEEEpeerreviewmaketitle

\section{Introduction}
\label{sec:intro}
Image blind deconvolution problem appears in many fields of image processing such as astronomy \cite{Murtagh2007}, biology \cite{Holmes2006} and medical imaging \cite{Michailovich2007}.
Given a degraded, blurred and noisy image, the aim is to restore a clean image along with an estimate of the blur kernel. Blind deconvolution is a severely ill-posed problem as there exists an infinite number of pairs (image/blur) that lead to the same observed image. Blind deconvolution methods adopt either a sequential identification process \cite{Carasso2001}, or a joint estimation approach \cite{Levin2009}. In the former, the blur kernel is identified first, possibly through a calibration step \cite{Bell16,Yunshi2021,Lefort2019}. Then the unknown image is inferred using a non-blind image restoration method. In the latter, the blur kernel and unknown image are simultaneously estimated. Since the problem is highly ill-posed, it is mandatory to incorporate prior knowledge on the sought unknowns. The retained prior strongly influences the choice for the solver. Let us distinguish three main classes of joint blind deconvolution approaches. A first option consists of formulating the problem as the minimization of a cost function gathering a data fidelity term (e.g., least-squares discrepancy) and penalties/constraints acting on the image and kernel variables. In such a way, it is quite standard to impose normalization and sparsity enhancing constraints on the kernel coefficients to avoid scale ambiguity inherent to the blind deconvolution model \cite{Komodakis2012,Kotera2013,Krishnan2011}. One can also easily impose the smoothness of the image, by adopting total-variation based regularization \cite{Chan1998}. Several other efficient choices have been proposed in the literature, along with suitable iterative optimization methods to solve the resulting problems \cite{Levin2006,Shan2008,Joshi2008,Abboud2019,Bolte2010}. The main advantage of this family of methods is probably its flexibility. But this comes at the price of heavy parameter tuning. The second option is to resort to a Bayesian formulation to express the model and a priori knowledge on the variables. The estimates are then defined from the estimation of the moments (typically, the mean) of a posterior distribution given the observed data and prior. As this typically involves the evaluation of intractable integrals, sampling \cite{Robert04,luengo2020survey} or approximation \cite{Evans95} strategies are used. Markov chain Monte Carlo (MCMC) methods have been widely used for blind deconvolution involving 1D sparse signals \cite{Rosec2003,Ge2011,Kail}, but it is up to our knowledge scarcely employed in large-scale problems \cite{Bishop2008}, probably for computational time reasons. Another family of approach consists in adopting the so-called variational Bayesian approximation paradigm \cite{Foxtutorial,David2016}. Then, a simpler (usually separable) approximation to the posterior is built through the minimization of a suitable divergence. This approach leads to fast Bayesian-based algorithms, whose great performance has been assessed in the context of non-blind \cite{Yorsa2017,Zheng6990542} and blind \cite{Babacan2009} image restoration. Bayesian-based techniques usually require less parameters than optimization-based ones. Moreover, they can provide higher-order moments estimates, such as covariance matrices, which are of high interest for assessing probabilistically the uncertainty of the results. However, dealing with complex noise models and priors in such methods may be tricky, and the algorithms may be quite computationally heavy. A recent trend is to insert optimization-based steps in Bayesian sampling/approximation methods for a more versatile modality and faster computations. See, for example, \cite{Yorsa2017,Pereyra2016, Marnissi2020} for applications of such ideas in the context of large-scale image processing. The third category of methods is more recent, as the references only go back to the last decade~\cite{Kupyn2018,Dong2016,Albluwi2018}. These methods rely on the deep learning methodology. More precisely, a supervised learning strategy is adopted to learn (implicitly) some prior information on the image/kernel from a so-called training set. A highly non-linear and multi-layers architecture is built, and its parameters (i.e., neuron weights) are estimated by back-propagation to minimize a given loss function associated with the task at hand (e.g., image visual quality). Several recent works propose neural network architectures dedicated to the problem of image blind deconvolution. Let us mention DeblurGAN \cite{Kupyn2018}, based on conditional generative adversarial networks and a multi-component loss function, SRCNN \cite{Dong2016} and its extended version, DBSRCNN \cite{Albluwi2018}, relying on a CNN architecture, SelfDeblur \cite{Ren2020} combining an optimization-based method with two GAN networks. These methods can reach very good performance, as long as the training set is large and representative enough. Moreover, they are well suited to GPU-based implementation. However, they have traditionally suffered from lack of interpretability and robustness \cite{Bietti2018}. An emerging set of methods, in the field of inverse problems in signal/image processing, performs algorithm unrolling \cite{Vishal2021}, where an iterative method (e.g., an optimization algorithm) is unrolled as layers of a neural network. The reduced set of parameters of this network are learnt by supervised training. Promising results have been obtained in the context of image deconvolution in \cite{Li2020, Bertocchi2019,Pesquet2020}. Theoretical results assessing the stability and robustness of unrolling techniques can be found in \cite{Pesquet2020,Patrick2020,Emilie2021}. These methods are also closely related to plug-and-play techniques where a trained neural network is employed as the denoiser \cite{Pesquet2020, Kamilov2021}.

In this paper, we propose a novel approach for blind image deconvolution, that aims at gathering the best of the three aforementioned methods. We first introduce a variational Bayesian algorithm (VBA) enhanced by optimization-based ideas from \cite{Yorsa2017}, with the advantages to cope with a large set of priors on the kernel and the image, and to present a reduced computational cost. Then, we apply the unrolling paradigm to create a deep neural network architecture, where VBA iterations are integrated as layers. This allows us to (i) learn the hyperparameters (in particular, the noise level) of VBA in an automatic supervised fashion, (ii) improve further the quality of the results by choosing a dedicated loss in the training phase, (iii) implement the method by taking full advantage of possible GPU resources, thus considerably reducing the processing time during the test phase.  In contrast to standard deep learning methods for blind deconvolution, all these benefits come along with a preservation of the interpretability of the method, thanks to the unrolling technique. Let us emphasize that variational Bayesian methods often appear in deep learning context. Indeed, they are backbones of variational autoencoders \cite{Kingma2014} and also constitute methods of choice for training Bayesian neural networks \cite{Jospin2020}. However, up to our knowledge, our work is the first to investigate the unrolling of a variational Bayesian technique.

The rest of our paper is organized as follows. In Section \ref{sec:VBA}, we introduce the image degradation model and introduce our Bayesian modeling, and provide the background for deriving our algorithm. Section \ref{sec:proposed} explicitly describes the iterative updates of the proposed VBA. The unrolling of VBA is presented in Section \ref{sec:unfoldedVBA}. Numerical results, including comparisons with various methods, are presented in Section \ref{sec:experiments}. Section \ref{sec:conclusion} concludes this paper.

\section{Problem statement}
\label{sec:VBA}

\subsection{Observation model}\label{se:obsemod}

We focus on the restoration of an original image $\widetilde{\x}\in\mathbb{R}^{N}$, from a degraded version of it $\y\in\mathbb{R}^{N}$, related to $\widetilde{\x}$ according to the following model:
\begin{equation}
{\y=\widetilde{\textbf{H}}\widetilde{\x}+\n.} \label{eq:model1}
\end{equation}
Hereabove, $\n\in\mathbb{R}^{N}$ models some additive random perturbation on the observation. Moreover, {$\widetilde{\textbf{H}} \in\mathbb{R}^{N\times N}$} is a linear operator modeling the effect of 
a blur kernel {$\widetilde{\h}\in\mathbb{R}^{M}$}. In this work, we focus on the generalized blind deconvolution problem where
the matrix associated with a given kernel $\h=[h_{1},\ldots,h_{M}]^\top$ reads
\begin{equation}
\textbf{H}=\sum^{M}_{m=1} h_{m}\textbf{S}_{m}, \label{eq:H_first}
\end{equation}
with $\{\textbf{S}_{1},\ldots,\textbf{S}_{M}\}$ is a set of $M$ known sparse $N\times N$ real-valued matrices. This model allows to retrieve the standard image deblurring model, as a special case when $\textbf{H}$ identifies with a 2D discrete convolution matrix with suitable padding. The considered problem amounts to retrieving an estimate {($\widehat{\x},\widehat{\h}$)} of the pair of variables {$(\widetilde{\x},\widetilde{\h})$} given $\y$. Due to the ill-posedness of this inverse problem, assumptions are required on the sought image / kernel and on the noise statistics to reach satisfying results. In the sequel, we will assume that $\n$ is a realization of an additive Gaussian noise with zero mean and standard deviation $\sigma$. In the remainder of the paper, it will be convenient to set $\beta =\sigma^{-2}$. Furthermore, 
we introduce a linear equality constraint on the blur kernel estimate $\h$.
A general expression of such a constraint is as follows:
\begin{equation}
\h=\textbf{T}\z+\textbf{t},
\label{h}
\end{equation}
where
$\textbf{T} = (T_{m,p})_{1\le m \le M,1 \le p \le P}\in\mathbb{R}^{M\times P}$ is a  matrix of rank $P\in \{1,\ldots,M\}$ and $\textbf{t}=[t_{1},\dots,t_M]^\top\in\mathbb{R}^{M}$ is some vector of predefined constants. Vector $\z=[z_{1},\ldots,z_{P}]^\top\in\mathbb{R}^{P}$ becomes the new unknown of the problem, along with the image $\x$. A typical linear equality constraint in such context is the sum-to-one constraint, i.e. $\sum_{m=1}^M h_m = 1$. Other examples will be provided in the experimental section. We can thus rewrite \eqref{eq:H_first} as
\begin{equation}\label{e:decHKp}
 \textbf{H}=\sum^{P}_{p=1}z_{p}\textbf{K}_{p}+\textbf{K}_{0}= \mathcal{H}(\z),
\end{equation}
with 
\begin{equation}
(\forall p \in \{1,\ldots,P\}) \quad \textbf{K}_{p} = \sum_{m=1}^{M} T_{m,p}\textbf{S}_{m} \in\mathbb{R}^{N\times N}, 
\end{equation}
and
\begin{equation}
\textbf{K}_{0} = \sum_{m=1}^{M}t_{m}\textbf{S}_{m} \in\mathbb{R}^{N\times N}.
\end{equation}
%
%

\subsection{Hierarchical Bayesian Modeling}\label{se:hierar}

Let us now introduce the hierarchical Bayesian model on which our VBA method will be grounded. 

\subsubsection{Likelihood}
First, we express the likelihood $p(\y|\x,\z)$ of the observed data, given the unknowns $(\x,\z)$. Since the noise is assumed to be Gaussian distributed, the likelihood can be expressed as follows:
\begin{align}
p(\y|\x,\z)
&=\beta^{\frac{N}{2}}\textrm{exp}\left(-\frac{\beta}{2}||\y-\mathcal{H}(\z)\x||^{2}\right),
\end{align}
where we recall that $\beta$ denotes the inverse of the noise variance. 
\subsubsection{Prior}
\label{eq:secprior}
As already mentioned, it is necessary to incorporate suitable prior knowledge on the sought quantities to limit the problem {ill-posedness}. We here consider a wide range of sparsity enhancing prior for the image $\x$, by adopting the generic model,  
\begin{equation}
p(\x|\gamma)\propto\gamma^{\frac{N}{2\kappa}}\exp\Big(-\gamma\sum_{j=1}^{J}||\textbf{D}_{j}\x||^{2\kappa}\Big),\label{eq:priorx}
\end{equation}
with $\kappa \in (0,1]$ a scale parameter and $(\textbf{D}_{j})_{1\leq j\leq J}\in(\mathbb{R}^{S\times N})^{J}$ 
both assumed to be known. For instance, an isotropic total variation prior is obtained by setting {$\kappa = 1/2$, $S = 2$, $J=N$ and for every $j \in \{1,\ldots,N\}$, $\textbf{D}_{j}\x =[[\nabla^h \x]_{j},[\nabla ^v {\x}]_{j}] \in \mathbb{R}^2$ gathers the horizontal and vertical gradients of $\x$ at pixel $j$.} Other relevant choices are discussed in \cite{Yorsa2017}. Hereabove, $\gamma>0$ is a regularization hyperparameter that we incorporate in our hierarchical model. We assume a Gamma distribution on $\gamma$,
\begin{equation}
p(\gamma)\propto\gamma^{\alpha-1}\exp(-\eta\gamma),
\end{equation}
where $\alpha\geq 0$ and $\eta\geq 0$ are the (known) shape and inverse scale parameters of the Gamma distribution. Such choice for the hyperparameter is rather standard in the context of Bayesian image restoration. 
\\
Regarding the blur $\h$, we adopt the so-called SAR model, successfully used for Bayesian-based blind deconvolution in {\cite{Babacan2009}}. The model relies on the following Gaussian model, 
\begin{equation}
p(\h|\xi)\propto \xi^{\frac{M}{2}}\exp\Big(-\frac{\xi}{2}||\textbf{A}(\h-\textbf{m})||^{2}\Big),
\label{hprior}
\end{equation}
where $\textbf{A}\in\mathbb{R}^{Q\times M}$ with $Q \in \mathbb{N}\setminus\{0\}$ denotes a matrix of rank $M$. $\textbf{m}\in\mathbb{R}^{M}$ 
is the mean of the underlying Gaussian distribution, and  $\xi>0$ is such that $\xi \textbf{A}^\top \textbf{A}$ is its inverse covariance matrix. 
If $\h$ follows this distribution, the projection of $\h$ onto the affine space defined by \eqref{h}
is also Gaussian as well as the vector $\z$ associated with each projected vector. More precisely,
$\z$ follows a Gaussian distibution with mean  $\bm{\mu}=\textbf{T}^{-1}(\textbf{m}-\textbf{t})$
and covariance matrix $\xi^{-1} \textbf{T}^{-1} (\textbf{A}^\top \textbf{A})^{-1} (\textbf{T}^{-1})^\top$
where $\textbf{T}^{-1}$ is the left inverse of $\textbf{T}$, i.e. $\textbf{T}^{-1} = (\textbf{T}^\top \textbf{T})^{-1}\textbf{T}^\top$.
This yields the following prior for the variable of interest $\z$:
\begin{equation}
p(\z|\xi)\propto \xi^{\frac{P}{2}}\exp\Big(-\frac{\xi}{2}(\z-\bm{\mu})^\top\bL(\z-\bm{\mu})\Big),
\end{equation}
where $\bL = \textbf{T}^\top \textbf{T}(\textbf{T}^\top  \big(\textbf{A}^\top \textbf{A})^{-1} \textbf{T}\big)^{-1} \textbf{T}^\top \textbf{T}$.
We will consider $(\bL,\bm{\mu})$ to be predefined by the user, so as to be adapted to the sought properties of the blur kernel to estimate. The hyperparameter $\xi$ will be {learned} during a training phase, as we will explain in Section \ref{sec:unfoldedVBA}. 

\subsubsection{Hierarchical model}
Let us assume that $(\x,\gamma)$ and $\z$ are mutually independent.
According to Bayes formula, the posterior distribution of the {unknowns} $\Theta=(\x,\z,\gamma)$ given the observed data $\y$ is defined as
\begin{equation}
p(\Theta|\y)\propto p(\y|\x,\z)p(\x|\gamma)p(\z|\xi)p(\gamma), \label{eq:post}
\end{equation}
where the four factors on the right side have been defined above. 

\subsection{Variational Bayesian Inference}

The Bayesian inference paradigm seeks for solving the blind restoration problem through the exploration of the posterior $p(\Theta|\y)$. Typically, one would be interested in the posterior mean, its covariance, or its modes (i.e., maxima). Let us make \eqref{eq:post} explicit:
\begin{align}
&p(\Theta|\y)\propto\textrm{exp}\left(-\gamma\sum_{j=1}^{J}||\textbf{D}_{j}\textbf{x}||^{2\kappa}-\frac{\beta}{2}||\y-\mathcal{H}(\z)\x||^{2}\right)\nonumber\\
&\times\gamma^{\frac{N}{2\kappa}+\alpha-1}\exp(-\eta\gamma)\xi^{\frac{P}{2}}
\exp\Big(-\frac{\xi}{2}(\z-\bm{\mu})^\top\bL(\z-\bm{\mu})\Big).
\label{eq:post2}
\end{align}
Unfortunately, neither $p(\Theta|\y)$, nor its moments (e.g., mean, covariance), nor its mode positions have a closed form. In particular $p(\y)$, which acts as a normalization constant, cannot be calculated analytically. We thus resort to the variational Bayesian framework to approximate this distribution by a more tractable one, denoted by $q(\Theta)$, for which the estimators are easier to compute. The approximation is computed with the aim to minimize the Kullback-Leibler (KL) divergence between the target posterior and its approximation, which amounts to determining
\begin{align}
q^{\textrm{opt}}(\Theta)&=\textrm{argmin}_{q}\  \mathcal{KL}(q(\Theta)||p(\Theta|\y)),\nonumber\\
&=\textrm{argmin}_{q}\  \int q(\Theta)\textrm{ln}\left(\frac{q(\Theta)}{p(\Theta|\y)}\right)\textrm{d}\Theta,
\label{min_problem}
\end{align}
where the equality holds only when $q(\Theta)=p(\Theta|\y)$. In order to make the solution of the above minimization problem tractable, a typical strategy is to make use of a variational Bayesian algorithm (VBA) based on a so-called mean field approximation of the posterior, combined with an alternating minimization procedure. 

The mean field approximation reads as a factorized structure $q(\Theta) =\prod_{r=1}^{R}q_{r}(\Theta_{r})$, which is assumed for the distribution $q$. Each of the $R$ factors are then obtained by minimizing the KL divergence by iterative update of a given factor $q_{r}$ while holding the others unchanged. This procedure takes advantage of the property that the minimizer of the KL divergence with respect to each factor can be expressed as
\begin{align}
(\forall r\in\{1,...,R\})  \quad  q_{r}^{\textrm{opt}}(\Theta_{r})\propto \exp\left(<\textrm{ln} p(\y, \Theta)>_{\prod_{i\neq r} q_{i}^{\textrm{opt}}(\Theta_{i})}\right) \label{eq:KLstandard}
\end{align}
where $<\cdot>_{\prod_{i\neq r}q_{i}(\Theta_{i})}=\int \, \prod_{i\neq r} q_{i}(\Theta_{i})\textrm{d}\Theta_{i}$. Here, we will consider the following factorization:
\begin{equation}
q(\Theta)=q_{\X}(\x)q_{\textbf{Z}}(\z)q_{\Gamma}(\gamma).
\end{equation}

In Section \ref{sec:proposed}, we describe {the steps of VBA} for this particular choice. Due to the intricate form of the chosen prior on the image, we introduce an extra approximation step, relying on a majoration-minimization (MM) strategy, reminescent from \cite{Yorsa2017}. In addition, we propose a strategy to reduce the time complexity of VBA, so as to deal with medium to large size images. As we will emphasize, the method requires the setting of two cumbersome hyperparameters, namely the regularization weight $\xi$ and the noise level $\beta$. Then, in Section~\ref{sec:unfoldedVBA}, we show how to unroll the VBA method as a neural network structure, so as to learn the parameters $(\xi,\beta)$ in a supervised fashion. 




\section{VBA for blind image deconvolution}
\label{sec:proposed}

We now describe our proposed implementation of the VBA when applied to the approximation to the posterior {in \eqref{eq:post2}}. We first present an MM-based procedure to handle the complicated form of the prior term on variable $\x$. Then, we give the explicit expressions of the updates performed in the alternating minimization method.

\subsection{MM-based approximation}
Let us focus on the prior term in~\eqref{eq:priorx}. This distribution is difficult to deal with as soon as $\kappa$ is different from $1$ (in which case a Gaussian distribution is retrieved). We thus propose to construct a surrogate for the prior on $\x$. We use the tangent inequality for concave functions, which yields the following majorant function for the $\ell_{\kappa}$-function with $\kappa\in(0,1]$:
\begin{equation}
(\forall u>0)(\forall v\geq0)\quad v^{\kappa}\leq (1-\kappa)u^{\kappa}+\kappa u^{\kappa-1}v.
\end{equation}
Let us introduce the vector of auxiliary positive variables $\bm{\lambda} = (\lambda_{j})_{1\leq j\leq J}$.
From the previous inequality,  we then deduce the following majorant function for the negative logarithm of the prior distribution:
\begin{equation}
(\forall \x \in \mathbb{R}^N) \quad \gamma\sum_{j=1}^{J}||\textbf{D}_{j}\x||^{2\kappa}\leq \sum_{j=1}^{J}
F_{j}(\textbf{D}_{j}\x,\lambda_{j};\gamma),
\label{eq:majprior}
\end{equation}
where, for every $j\in\{1,\ldots,J\}$,
\begin{equation}
F_{j}(\textbf{D}_{j}\x,\lambda_{j};\gamma)=\gamma\frac{\kappa||\textbf{D}_{j}\x||^{2}+(1-\kappa)\lambda_{j}}{\lambda_{j}^{1-\kappa}}.
\end{equation}
This majorant function can be understood as a Gaussian lower bound on the prior distribution on $\x$, which will appear more tractable in the VBA implementation. We will also show that the update of the auxiliary variables remains rather simple, thus not impacting the complexity of the whole procedure. 

In a nutshell, using \eqref{eq:post2}, and \eqref{eq:majprior}, we obtain the following inequality:
\begin{equation}
p(\Theta|\y)\geq \mathcal{F}(\Theta|\y;\bm{\lambda}) \label{eq:majpost}
\end{equation}
where the lower bound on the posterior distribution is
\begin{align}
&\mathcal{F}(\Theta|\y;\bm{\lambda})=\nonumber\\
&C\gamma^{\frac{N}{2\kappa}}\exp\left(-\frac{\beta}{2}||\y-\mathcal{H}(\z)\x||^{2}-F(\x,\bm{\lambda};\gamma)\right)p(\gamma)p(\z|\xi).
\end{align}
Hereabove we have introduced  the shorter notation 
\begin{equation}
F(\x,\bm{\lambda};\gamma)=\sum^{J}_{j=1}F_{j}(\textbf{D}_{j}\x,\lambda_{j};\gamma)
\end{equation}
and $C$ is a multiplicative constant independent from $\Theta$. Inequality~\eqref{eq:majpost} leads to the following majorization of the $\mathcal{KL}$ divergence involved in \eqref{min_problem}:
\begin{equation}
\mathcal{KL}(q(\Theta)||p(\Theta|\y))\leq \mathcal{KL}(q(\Theta)||\mathcal{F}(\Theta|\y;\bm{\lambda})).
\label{KL}
\end{equation}
By minimizing the upper bound in \eqref{KL} with respect to $\bm{\lambda}$, we can keep it as tight as possible, so as to guarantee the good performance of the VBA. To summarize, we propose to solve Problem~\eqref{min_problem} through the following four iterative steps:
\begin{enumerate}
	\item Minimizing $\mathcal{KL}(q(\Theta)||\mathcal{F}(\Theta|\y;\bm{\lambda}))$ w.r.t. $q_{\X}(\x)$.
	\item Minimizing the upper bound $\mathcal{KL}(q(\Theta)||\mathcal{F}(\Theta|\y;\bm{\lambda}))$ in \eqref{KL} w.r.t. $q_{\textbf{Z}}(\z)$.
	\item Update the auxiliary variables $(\lambda_{j})_{1 \leq j \leq J}$ to minimize $\mathcal{KL}(q(\Theta)||\mathcal{F}(\Theta|\y;\bm{\lambda}))$.
	\item Minimizing $\mathcal{KL}(q(\Theta)||\mathcal{F}(\Theta|\y;\bm{\lambda}))$ w.r.t. $q_{\Gamma}(\gamma)$.
\end{enumerate}
Subsequently, at a given iteration $k$ of the proposed algorithm, the corresponding estimated variables will be indexed by $k$.

\subsection{VBA updates}
Let us now describe the four steps of the proposed VBA, starting from a given iteration $k$ associated with the current approximated distributions $q^k_{\X}(\x), q^k_{\textbf{Z}}(\z)$, and $q^k_{\Gamma}(\gamma)$, and the auxiliary parameter estimate $\bm{\lambda}^{k}$. We also denote by $(\x^{k},\z^{k},\gamma^k)$ the estimates of  the means of $q^k_{\X}$, $q^k_{\textbf{Z}}$, and $q^k_{\Gamma}$, and $(\Cx^{k},\Ch^{k})$ the covariance estimates for $q^k_{\X}$ and $q^k_{\textbf{Z}}$.


\subsubsection{Update of $q_{\X}(\x)$}
By definition,
\begin{equation}
q_{\X}^{k+1}(\x)\\
=\textrm{argmin}_{q_{\X}} \mathcal{KL}(q_{\X}(\x)q^{k}_{\Gamma}(\gamma)q^{k}_{\textbf{Z}}(\z)||\mathcal{F}(\Theta|\y;\bm{\lambda}^{k})).
\end{equation}
The standard solution provided by \eqref{eq:KLstandard} remains valid, by replacing the joint distribution by a lower bound 
chosen proportional to $\mathcal{F}(\Theta|\y;\bm{\lambda}^{k})$:
\begin{align}
q_{\x}^{k+1}(\x)& \propto\exp\left(<\textrm{ln}\,\mathcal{F}(\x,\z,\gamma\mid \y; \bm{\lambda}^{k})>_{q^{k}_{\Gamma}(\gamma),q^{k}_{\textbf{Z}}(\z)}\right)\nonumber\\
&\propto\exp\left(\int\int\textrm{ln}\,\mathcal{F}(\x,\z,\gamma\mid\y;\bm{\lambda}^{k})q^{k}_{\Gamma}(\gamma)q^{k}_{\textbf{Z}}(\z)\textrm{d}\gamma\textrm{d}\z\right).
\end{align}
By decomposing the different terms and using \eqref{e:decHKp},
\begin{align}
q_{\x}^{k+1}(\x)\propto &\exp\Biggl\{-\frac{1}{2}\x^\top\Biggr(\beta  \Big(\mathbb{E}_{q_{\textbf{Z}}^{k}(\z)}(\textbf{H})^\top \mathbb{E}_{q_{\textbf{Z}}^{k}(\z)}(\textbf{H})\nonumber\\
&+\sum^{P}_{p=1}\sum^{P}_{q=1}\textbf{e}^\top_{p}\textrm{cov}_{q^{k}_{\textbf{Z}}(\z)}(\z)\textbf{e}_{q}\textbf{K}^\top_{p}\textbf{K}_{q}\Big)\nonumber\\
&+2\mathbb{E}_{q_{\Gamma}^{k}(\gamma)}(\gamma)\textbf{D}^\top\Lambda^{k}\textbf{D}\Biggr)\x+\beta \x^\top \mathbb{E}_{q_{\textbf{Z}}^{k}(\z)}(\textbf{H})^\top\y\Biggl\}
\end{align}
where 
{
\begin{align}
\mathbb{E}_{q_{\textbf{Z}}^{k}(\z)}(\textbf{H})&=\sum^{P}_{p=1}\textbf{e}^\top_{p}\mathbb{E}_{q^{k}_{\textbf{Z}}(\z)}(\z)\textbf{K}_{p}+\textbf{K}_{0},\\
\textbf{D} & = [\textbf{D} _{1}^\top,\ldots,\textbf{D} _{J}^\top]^\top, \label{eq:matrixD}
\end{align}
}
$\Lambda^k$ is the block diagonal matrix whose diagonal elements are $(\kappa(\lambda_{j}^{k})^{\kappa-1}\textbf{I}_{S})_{1\leq j\leq J}$, and 
$(\textbf{e}_{1},\ldots,\textbf{e}_{P})$ is the canonical basis of $\mathbb{R}^P$.
We thus obtain a Gaussian distribution:
\begin{equation}
q_{\textbf{X}}^{k+1}(\x) = \mathcal{N}(\x;\check{\textbf{x}}^{k+1},\check{\Cx}^{k+1}),
\end{equation}
parametrized by
\begin{align}
 {(\check{\Cx}^{k+1})^{-1}} 
=\;& 
\beta  \Biggr((\newH^{k})^\top\newH^{k}
+\sum^{P}_{p=1}\sum^{P}_{q=1}\textbf{e}^\top_{p}{\Ch^{k}}\textbf{e}_{q}\textbf{K}^{\top}_{p}\textbf{K}_{q}\Biggr)\nonumber\\
&+2\newgamma^{k}\textbf{D}^{\textrm{T}}\newLambda^{k}\textbf{D},
\label{Cx}\\
\check{\textbf{x}}^{k+1}
=\;& \beta {\check{\Cx}^{k+1}}(\newH^{k})^\top\y,
\label{mx}
\end{align}
with {$\newH^{k}=\mathcal{H}(\z^{k})$}. 

In image restoration applications, dimension $N$ can be rather large (typically greater than $10^6$ variables), so that the storage of the full covariance matrix $\check{\Cx}^{k+1}$ is neither desirable nor usually possible. We thus propose to resort to a diagonal approximation to this matrix when required, so that the update finally reads:
\begin{equation}
q_{\textbf{X}}^{k+1}(\x) = \mathcal{N}(\x;{\x^{k+1}},\Cx^{k+1}),
\end{equation}
with
\begin{align}
\Cx^{k+1}&=\textrm{Diag}\left(\newdeltax^{k+1}\right)\label{Cx_hat}\\
{\x^{k+1}}&=\textrm{CG}\Big({(\check{\Cx}^{k+1})^{-1}},\beta (\newH^{k})^\top \y\Big),
\label{mx_hat}
\end{align}
where $\newdeltax^{k+1} \in \mathbb{R}^N$ is the vector of the inverses of 
the diagonal elements of {$(\check{\Cx}^{k+1})^{-1}$}, and $\textrm{CG}(\texttt{A,b})$ denotes the application of a linear conjugate gradient solver to the linear system $\texttt{Ax=b}$.

\subsubsection{Update of $q_{\textbf{Z}}(\z)$}
According to the VBA principle, 
\begin{equation}
q_{\textbf{Z}}^{k+1}(\z)\\
=\textrm{argmin}_{q_{\textbf{Z}}}  \mathcal{KL}(q^{k+1}_{\X}(\x)q^{k}_{\Gamma}(\gamma)q_{\textbf{Z}}(\z)||\mathcal{F}(\Theta|\y;\bm{\lambda}^{k})).
\end{equation}
Using \eqref{eq:KLstandard} and the previously introduced bound $\mathcal{F}(\Theta\mid\y;\bm{\lambda}^{k})$, we have
\begin{align}
q_{\textbf{Z}}^{k+1}(\z) 
&\propto\exp\left(\int\int\textrm{ln}\,\mathcal{F}(\x,\z,\gamma\mid\y;\bm{\lambda}^{k})q^{k}_{\Gamma}(\gamma)q^{k+1}_{\X}(\x)\textrm{d}\gamma\textrm{d}\x\right).
\end{align}
Replacing the involved quantities by their expression yields
\begin{multline}
q_{\textbf{Z}}^{k+1}(\z) 
\propto \exp\Biggl\{-\frac{1}{2}\z^\top\left(\beta \textbf{B}^{k+1}+\xi \bL\right)\z
\\
+\z^\top\left(\beta \textbf{a}^{k+1}+\xi \bL\bm{\mu}\right)\Biggl\},
\end{multline}
where $\textbf{a}^{k+1} = (a^{k+1}_{p})_{1 \leq p \leq P} \in\mathbb{R}^{P}$ and $\textbf{B}^{k+1}= (B^{k+1}_{p,q})_{1 \leq p,q \leq P}\in\mathbb{R}^{P\times P}$ 
are such that, for every $(p,q)\in \{1,\ldots,P\}^2$,\\
\begin{align}
a^{k+1}_{p}=&\mathbb{E}_{q_{\textbf{X}}^{k+1}(\textbf{x})}(\textbf{x})^\top\textbf{K}^\top_{p}\y
- \mathbb{E}_{q_{\textbf{X}}^{k+1}}(\textbf{x}^\top\textbf{K}^\top_{p}\textbf{K}_{0}\textbf{x})
\nonumber\\
& = (\x^{k+1})^\top\textbf{K}^\top_{p}\textbf{y}-\textbf{B}^{k+1}_{p,0},\\
B^{k+1}_{p,q}& =\mathbb{E}_{q_{\textbf{X}}^{k+1}}(\textbf{x}^\top\textbf{K}^\top_{p}\textbf{K}_{q}\textbf{x})
\nonumber\\
& = \textrm{trace}\left(\textbf{K}_{p}\Cx^{k+1}\textbf{K}^{\top}_{q}\right)+({\x^{k+1}})^\top\textbf{K}^\top_{p}\textbf{K}_{q}\x^{k+1}
\label{matrixB_1}
\end{align}
with
\begin{align}
B^{k+1}_{p,0}& =\mathbb{E}_{q_{\textbf{X}}^{k+1}}(\textbf{x}^\top\textbf{K}^\top_{p}\textbf{K}_{0}\textbf{x})
\nonumber\\
& = \textrm{trace}\left(\textbf{K}_{p}\Cx^{k+1}\textbf{K}^{\top}_{0}\right)+({\x^{k+1}})^\top\textbf{K}^\top_{p}\textbf{K}_{0}\x^{k+1}.
\label{matrixB_2}
\end{align}
Thus, the update for the distribution $q_{\textbf{Z}}$ reads
\begin{equation}
q^{k+1}_{\textbf{Z}}(\z) = \mathcal{N}(\z;{\z^{k+1}},\Ch^{k+1}),
\end{equation}
with 
\begin{align}
{(\Ch^{k+1})^{-1}}&=\beta \textbf{B}^{k+1}+\xi \bL,
\label{Cz}\\
{\z^{k+1}}&={\Ch^{k+1}}\left(\beta  \textbf{a}^{k+1}+\xi \bL\bm{\mu}\right).
\label{mz}
\end{align}

\subsubsection{Update of $\bm{\lambda}$}
Let us now express the update of the auxiliary variable. We aim at finding
\begin{equation}
\bm{\lambda}^{k+1}\\
=\textrm{argmin}_{\bm{\lambda}}  \mathcal{KL}(q^{k+1}_{\textbf{X}}(\x)q^{k}_{\Gamma}(\gamma)q^{k+1}_{\textbf{Z}}(\z)||\mathcal{F}(\Theta|\y;\bm{\lambda})).
\end{equation}
This amounts to finding, for every $j \in \{1,\ldots,J\}$,
\begin{align}
\lambda^{k+1}_{j}&=\textrm{argmin}_{\lambda_{j}\in[0,+\infty)}\int q^{k+1}_{\X}(\textbf{x})q^{k}_{\Gamma}(\gamma)q^{k+1}_{\textbf{Z}}(\z)\nonumber\\
&\qquad\quad\times\textrm{log}\frac{q^{k+1}_{\X}(\x)q^{k}_{\Gamma}(\gamma)q^{k+1}_{\textbf{Z}}(\z)}{\mathcal{F}(\Theta|\y,\bm{\lambda})}\textrm{d}\Theta,\nonumber\\
&=\textrm{argmin}_{\lambda_{j}\in[0,+\infty)}\sum^{J}_{j=1}\int\int q^{k+1}_{\X}(\x)q^{k}_{\Gamma}(\gamma)\nonumber\\
&\qquad\quad\times F_{j}(\textbf{D}_{j}\x,\lambda_{j};\gamma)\textrm{d}\x\textrm{d}\gamma,\nonumber\\
&=\textrm{argmin}_{\lambda_{j}\in[0,+\infty)}\frac{\kappa \mathbb{E}_{q^{k+1}_{\x}(\x)}\left[||\textbf{D}_{j}\x||^{2}\right]+(1-\kappa)\lambda_{j}}{\lambda_{j}^{1-\kappa}}.
\end{align}
The explicit solution to the above minimization problem yields the following update:
\begin{align}
\lambda^{k+1}_{j}&=\mathbb{E}_{q^{k+1}_{\x}(\x)}\left[||\textbf{D}_{j}\x||^{2}\right]\nonumber\\
& = ||\textbf{D}_{j}\x^{k+1}||^{2}+\textrm{trace}\left(\textbf{D}^{\textrm{T}}_{j}\textbf{D}_{j}\Cx^{k+1}\right).
\label{lambda}
\end{align}

\subsubsection{Update of $q_{\Gamma}(\gamma)$}
Finally, the update related to the hyperparameter $\gamma$ is expressed as
\begin{equation}
q_{\Gamma}^{k+1}(\gamma)\\
=\textrm{argmin}_{q_{\Gamma}}  \mathcal{KL}(q^{k+1}_{\textbf{X}}(\x)q_{\Gamma}(\gamma)q^{k+1}_{\textbf{Z}}(\z)||\mathcal{F}(\Theta|\y;\bm{\lambda}^{k+1})).
\end{equation}
Using \eqref{eq:KLstandard}, we have
\begin{multline}
q_{\Gamma}^{k+1}(\gamma)
\propto\exp\Biggr(\int\int\textrm{ln}\,\mathcal{F}(\x,\z,\gamma\mid\y;\bm{\lambda}^{k+1})
\\
\times q^{k+1}_{\X}(\x)q^{k+1}_{\textbf{Z}}(\z)\,\textrm{d}\x\textrm{d}\z\Biggr).
\end{multline}
The above integral has the following closed form expression:
\begin{align}
q_{\Gamma}^{k+1}(\gamma)&\propto\gamma^{\frac{N}{2\kappa}+\alpha-1}\exp(-\eta\gamma)\nonumber\\
&\times\exp\left(-\gamma\sum^{J}_{j=1}\frac{\kappa\, \mathbb{E}_{q^{k+1}_{\x}(\x)}\left[||\textbf{D}_{j}\textbf{x}||^{2}\right]+(1-\kappa)\lambda^{k+1}_{j}}{(\lambda^{k+1}_{j})^{1-\kappa}}\right).
\end{align}
It thus follows from \eqref{lambda} that the update of $q_{\Gamma}$ is
\begin{align}
q_{\Gamma}^{k+1}(\gamma) = \Gamma(d,b^{k+1}),
\end{align}
that is the Gamma distribution with parameters
\begin{align}\label{e:dbk}
d=\frac{N}{2\kappa}+\alpha,\ \ b^{k+1}=\sum^{J}_{j=1}(\lambda^{k+1}_{j})^{\kappa}+\eta.
\end{align}
The mean of $q_{\Gamma}^{k+1}$ is finally given by
\begin{equation}
\gamma^{k+1} = \frac{d}{b^{k+1}}.
\label{gamma}
\end{equation}
Note that parameter $d$ is not iteration dependent and can thus be precomputed from the beginning of the VBA. 

\subsection{Overview of VBA}
Algorithm \ref{alg:VBA} provides a summary of the resulting VBA for solving the blind deconvolution problem introduced in Section~\ref{sec:VBA}. We also specify our initialization strategy. More practical details about the latter will be discussed in the experimental section. 
As a result, the optimal posterior distributions for both variables $\x$ and $\z$ will be approximated as Gaussian distributions, while the one for hyperparameter $\gamma$ is approximated by a Gamma distribution. In particular,  after $K$ iterations, it is direct to extract from VBA outputs an estimate for the posterior mean of the image and the kernel, through variable {$\x^{K}$} and {$\textbf{T}\z^{K} + \textbf{t}$}. The associated covariance matrices are given by $\Cx^K$ and {$\textbf{T}\Ch^{K}\textbf{T}^\top$}.
These matrices can be useful to perform uncertainty quantification of the results. The VBA also allows us to estimate easily the hyperparameter $\gamma$ involved in the image prior. Nonetheless, it appears difficult to find an efficient manner to estimate the hyperparameter $\xi$ using a variational Bayesian approach, as this value highly fluctuates from one image/kernel pair to the other so that a simple prior modeling of does not appear obvious. Moreover, the VBA requires the knowledge of the noise level, through the parameter $\beta$. This is limitating, and one might prefer to have this quantity estimated in an automatic manner. Thus, we propose in the next section, to resort to a supervised learning strategy to learn both $\xi$ and $\beta$ along the iterates of VBA, in the spirit of recent works \cite{Bertocchi2019} on the unrolling (also called unfolding) 
of iterative algorithms. 



\begin{algorithm}

{
\caption{VBA approach for image blind deconvolution}\label{alg:VBA}
  \begin{algorithmic}[1]
\INITIALIZATION Set hyperparameters $(\xi,\beta,\alpha,\eta)$. Define initial values for $(\x^0,\Cx^0,\z^{0}, \Ch^{0})$. Compute $\newlambda^{0}$ and $\gamma^0$ using \eqref{lambda} and \eqref{gamma}, respectively.
\ITERATION For $k=0,1,\ldots,K$:
\State Update the mean {$\x^{k+1}$} and the covariance matrix $\Cx^{k+1}$ of $q_{\X}^{k+1}(\x)$ using  \eqref{Cx_hat}-\eqref{mx_hat}.
\State Update the mean {$\z^{k+1}$} and the covariance matrix $\Ch^{k+1}$ of $q_{\textbf{Z}}^{k+1}(\z)$ using \eqref{Cz}-\eqref{mz}.
\State Update $\lambda^{k+1}_{j}$ using \eqref{lambda}, for every $j\in\{1,\ldots,J\}$.
\State Update the mean $\gamma^{k+1}$ of $q_{\Gamma}^{k+1}(\gamma)$ using \eqref{e:dbk}-\eqref{gamma}.
 \end{algorithmic}
 }
 \label{algo:VBA}
\end{algorithm}


\section{Supervised learning of VBA hyperparameters}
\label{sec:unfoldedVBA}

\subsection{Overview}
We introduce a supervised learning strategy to estimate the hyperparameter $\xi$ and  {the inverse of the noise variance $\beta$}, that are required to run VBA. We adopt the so-called \emph{unrolling} (or unfolding) methodology \cite{Vishal2021}. The idea is to view each iteration of an iterative algorithm as one layer of a neural network structure. Each layer can be parametrized by some quantities that are learned from a training database so as to minimize a task-oriented loss function. The advantage of the unrolling approach is threefold: (i) each layer {mimics} one iteration of the algorithm and thus it is highly interpretable, (ii) the choice of the loss is directly related to the task {at the end}, which is beneficial for the quality of the results, (iii) once trained, the network can be applied easily and rapidly on a large set of test data without any further tuning. In particular, its implementation can make use of GPU-accelerated frameworks. Several recent examples in the field of image processing have shown the benefits of unrolling \cite{Gilton2021,Li2021,Tolooshams2021,Nan2020} when compared to standard black-box deep learning techniques or more classical restoration methods based on Bayesian or optimization tools. {Let us in particular mention the works \cite{Li2020,Li2019} for the application of unrolling in the context of blind image restoration}.

\begin{figure}[h]
\centering
\includegraphics[width = 9cm]{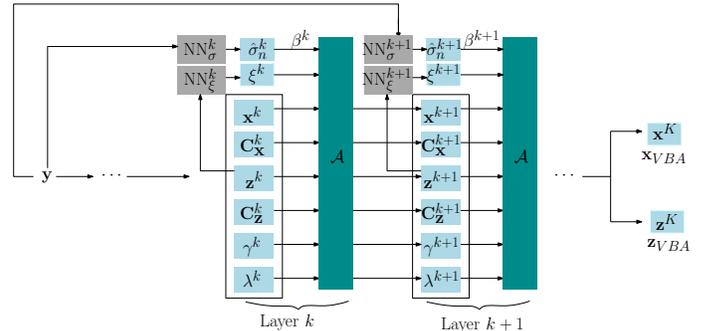}  
\caption{\footnotesize Architecture of \emph{unfoldedVBA} network.}
\label{fig:unfoldedVBA}
\end{figure}

Let us now specify the unrolling procedure in the context of VBA. Let $K >0$ be the number of iterations of the VBA described in Algorithm \ref{algo:VBA}, thus corresponding to $K$ layers of a neural network architecture. Iteration $k\in\{0,\ldots,K-1\}$ of our unrolled VBA can be conceptually expressed as
\begin{align}
&(\x^{k+1},\Cx^{k+1} ,\z^{k+1},\Ch^{k+1},\bm{\lambda}^{k+1},\newgamma^{k+1}) \nonumber\\
&= \mathcal{A}(\x^{k},\Cx^{k},\z^{k},\Ch^{k},\bm{\lambda}^{k},\newgamma^{k},\newxi^k,\beta^k). \label{eq:Aapp}
\end{align}
The initialization procedure for $(\x^{0},\Cx^{0},\z^{0},\Ch^{0},\bm{\lambda}^{0},\newgamma^{0})$ is detailed in Algorithm~\ref{algo:VBA}. For $k\in\{0,\ldots,K-1\}$, the expressions of $(\x^{k+1},\Cx^{k+1},\z^{k+1},\Ch^{k+1},\newgamma^{k+1},\bm{\lambda}^{k+1})$ as a function of the input arguments of $\mathcal{A}(\cdot)$ are given respectively by {\eqref{Cx_hat}-\eqref{mx_hat}}, \eqref{Cz}-\eqref{mz}, \eqref{lambda}, and \eqref{e:dbk}-\eqref{gamma}. Furthermore, $(\newxi^k,\beta^k)_{0 \leq k \leq K-1}$ are now learned, instead of being constant and preset by the user. This leads to the \emph{unfoldedVBA} architecture depicted in Fig.~\ref{fig:unfoldedVBA}, which can be summarized into the composition of $K$ layers $\mathcal{L}_{K-1}\circ \cdots \circ \mathcal{L}_{0}$. Each layer $\mathcal{L}_{k}$ with $k \in \{0,\ldots,K-1\}$ is made of three main blocks, that are two neural networks, namely $\textrm{NN}_{\sigma}^k$ and $\textrm{NN}_{\xi}^k$, and the core VBA block $\mathcal{A}(\cdot)$. There remains to specify our strategy for building the two inner networks, with the aim to learn $(\newxi^k,\beta^k)_{0 \leq k \leq K-1}$.


\subsection{Learning hyperparameter $\xi$}
\label{cnn_xi}
\begin{figure}[h]
\centering
\includegraphics[width = 8cm]{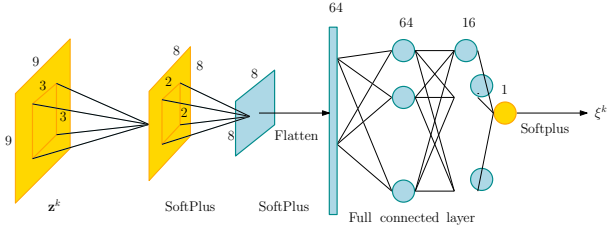}  
\caption{\footnotesize Neural network architecture $\textrm{NN}_{\xi}^{k}$ for estimating $\xi^{k}$, for $k\in\{0,...,K-1\}$.}
\label{fig:CNN}
\end{figure}

For every $k\in\{0,...,K-1\}$, neural network {$\textrm{NN}_{\xi}^{k}$} takes as input the current kernel estimate $\h^k = \textbf{T}\z^{k} + \textbf{t}$ and delivers $\xi^k$ as an output. The architecture of the neural network is shown in Fig.~\ref{fig:CNN}. Note that the Softplus function, defined as
\begin{equation}
(\forall x \in \mathbb{R}) \quad \textrm{Softplus}(x) = \textrm{ln}(1+\exp{(x)}),
\end{equation}
is used as a last layer, in order to enforce the {strict} positivity of the output hyperparameter $\xi^{k}$.

\subsection{Learning noise {parameter} $\beta$}
\label{nn_sigma}

When the noise {parameter} $\beta$ is unknown, it might be useful to include a procedure to learn it automatically, again in a supervised fashion. In this case, we propose to introduce simple nonlinear mappings such that, for every $k\in\{0,\dots,K-1\}$,
\begin{align}
{{\sigma}^{k}} &=\textrm{NN}_{\sigma}^{k}(\y),\nonumber\\
&= \textrm{Softplus}(\rho_{k})\widehat{\sigma}(\y)+\textrm{Softplus}(\tau_{k}),
\end{align}
and $\beta^k = ({{\sigma}^{k}})^{-2}$. Hereabove, $\y$ is the observed degraded image, from which we deduce the wavelet-based variance estimator (also used in \cite{Bertocchi2019}), 
\begin{align}
\widehat{\sigma}(\y) = \frac{\textrm{median}(|\mathbf{W}_{H} \y|)}{0.6745},
\end{align}
where $|\mathbf{W}_{H} \y|$ gathers the absolute value of the diagonal coefficients of the first level Haar wavelet decomposition of the degraded image $\y$. Moreover, $(\rho_k,\tau_k)_{0 \leq k \leq K-1}$ are two {scalar} parameters to be learned during the training phase.

%

\subsection{Complete architecture}

We now present our complete blind deconvolution architecture for grayscale images and color images in Fig.~\ref{fig:pipeline_both}. First, let us notice that VBA and its unrolled variant is designed for grayscale images. We thus generalized the architecture from Fig.~\ref{fig:pipeline_both}(top), to process color images. To this end, we first transform the input RGB image to its YUV representation, which takes human perception into consideration. The network {$\textrm{NN}_{\sigma}^{k}$} is first applied to the luminance part $\y_{Y}$ of the image. After applying the \emph{unfoldedVBA} network (see Fig.~\ref{fig:unfoldedVBA}), we obtain $\z_{\text{VBA}}$ and $\x_{\text{VBA}}$ as outputs. The latter is a restored version of the luminance channel. The remaining (U,V) color channels are simply obtained by median filtering of $(\y_{U},\y_{V})$. Both architectures in Fig.~\ref{fig:pipeline_both} additionally {involve} post-processing layers. More precisely, we first include a linear layer so as to encode the linear transformation \eqref{h}, and then deduce the estimated blur kernel $\widehat{\h}$. Second, we also allow a post-processing layer $\mathcal{L}_{\text{pp}}$ acting on the image, so as to reduce possible residual artifacts, finally yielding $\widehat{\x}$. In the case of color images, the post-processing is applied on the RGB representation to avoid chromatic artifacts.


\subsection{Training procedure}
\label{ssec:train}
\begin{figure}[h]
\centering
\begin{tabular}{c}
\includegraphics[width = 7.5cm]{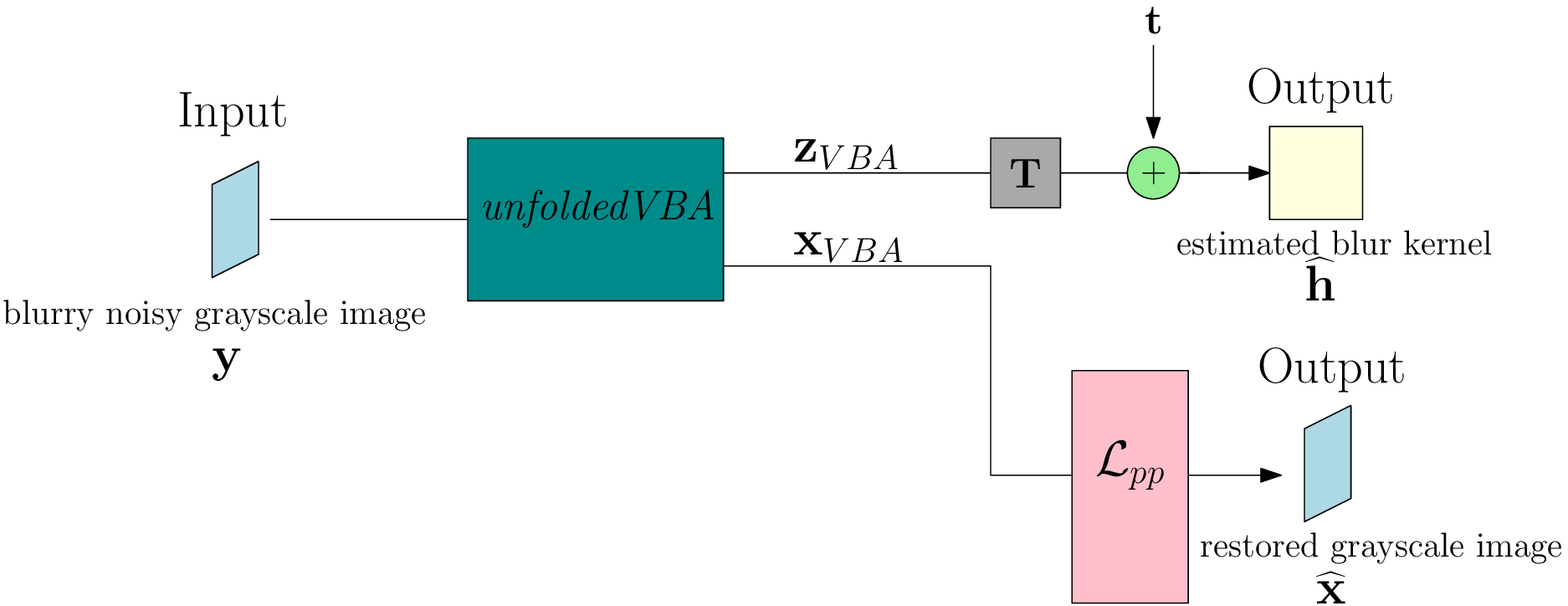}  
\\
\includegraphics[width = 8cm]{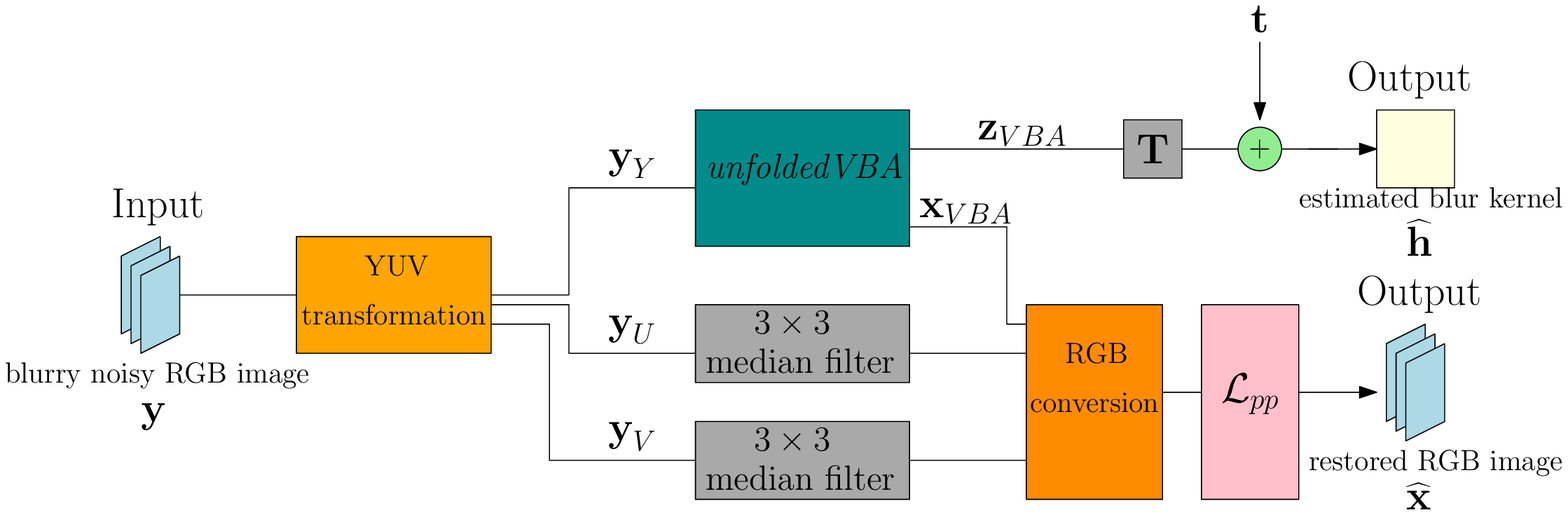}
\end{tabular}
\caption{\footnotesize Proposed  blind image restoration pipeline for grayscale (top) and color (bottom) images.}
\label{fig:pipeline_both}
\end{figure}

The training of both proposed architectures from Fig.~\ref{fig:pipeline_both} {requires} to define a loss function, measuring the discrepancy between the output $(\widehat{\x},\widehat{\h})$ and the ground truth $(\overline{\x},\overline{\h})$, that we denote hereafter by {$\ell(\widehat{\x},\widehat{\h},\overline{\x},\overline{\h})$}. In the blind deconvolution application, one can for instance consider a loss function related to the error reconstruction on the kernel, or to the image quality, or a combination of both. Two training {procedures} will be distinguished and discussed in our experimental section, namely:\\
\textbf{Greedy training} The parameters of the unfolded VBA are learned in a greedy fashion so as to minimize the kernel reconstruction error at each layer. Then, the post-processing network is learned in a second step, so as to maximize an image quality metric such as the SSIM \cite{Wang2004}.\\
\textbf{End-to-end training} The parameters of the complete architecture are learned end-to-end so as to maximize the image quality metric.

Whatever the chosen training procedure, it is necessary to make use of a {back-propagation} step, that is to differentiate the loss function with respect to all the parameters of the network. Most operations involved in Fig.~\ref{fig:pipeline_both} can be differentiated efficiently using standard auto-{differentiation} tools. However, we {observed} in our experiments that it {is} beneficial (and sometimes even necessary) for a stable training phase to avoid using such tools for differentiating the VBA layer $\mathcal{A}(\cdot)$ involved in Fig.~\ref{fig:unfoldedVBA}. In practice, we used the explicit expressions for the partial derivatives of it. Note that we followed the approach from \cite{Benjamin2020} to obtain the expression of the derivatives for the CG solver. 



\section{Numerical experiments}
\label{sec:experiments}
\subsection{Problem formulation and settings}

\subsubsection{Problem overview}
We focus on the resolution of the blind image deconvolution problem, where $\widetilde{\x}\in\mathbb{R}^{N}$ is an original image, either grayscale or colored one. 
We come back to the model presented in Section \ref{se:obsemod},
where the linear operator {$\widetilde{\textbf{H}} \in\mathbb{R}^{N\times N}$} models the application of a blur kernel {$\widetilde{\h}\in\mathbb{R}^{M}$} to the image. The noise $\n$ is assumed to be an additive white Gaussian noise with zero mean and standard deviation $\sigma$. In the case of color images, we assume that the same kernel, and the same noise level, is applied to each of the three RGB channels. 

\subsubsection{VBA settings}
Let us now specify our practical choices, for the implementation of the VBA step. In all our experiments, we seek for kernels whose entries satisfy two equality constraints, namely a sum to one constraint, and an axial symmetry along the main diagonal axis. This can be easily translated into the affine constraint \eqref{h}. In such case, the degree of freedom of the kernel model is equal to {$P = \frac{(\sqrt{M}+1)\sqrt{M}}{2}-1$}. Regarding the choice for the prior, we set $\textbf{A}\in\mathbb{R}^{(2M+1)\times M}$  as the matrix that computes the horizontal and vertical differences between pixels, augmented with an additional first row corresponding to an averaging operation, which takes the form $[1,\ldots,1]/M$. This choice allows to promote smooth variations in the kernels, while satisfying the required full column rank assumption on $\textbf{A}$. A constant vector with entries equal to $\frac{1}{M}$ is set for the prior mean {$\textbf{m}$}. Matrices $(\textbf{D}_j)_{1 \leq j \leq J}$ and parameter $\kappa$, involved in \eqref{eq:priorx}, are set in such a way that the chosen prior on the image yields an isotropic total-variation regularization (see our comment in Sec.~\ref{eq:secprior}). We must now specify the initialization for VBA iterates/layers. Our initial guess $\x^{0}$ for the image is the degraded image. The associated covariance matrix $\Cx^{0}$ is initialized using the identity matrix. {The blur is initialized with a uniform kernel with size $5 \times 5$, from which we deduce the corresponding $\z^0$, and the covariance matrix $\Ch^{0}$ is set to a multiple of identity matrix. The hyperparameters $(\alpha,\eta)$ involved in the prior law on parameter $\gamma$are set to zero in practice which is equivalent to impose a non-informative Jeffrey improper prior. Finally, the conjugate gradient solver used for the update of the image is run over 10 iterations which {appears} sufficient to reach practical stability. The solver is initialized with the degraded image.

%
%

\subsubsection{Datasets}
Let us now introduce the two datasets we {employ} to train and test our network, and compare it to state-of-the-art techniques. In both cases, the training set is made of 100 images from the COCO training set. The validation set contains 40 images from the BSD500 validation set. The test set consists of 30 images from the Flickr30 test set. Each image is center-cropped using a window of size $N = 256\times 256$. Each original image $\widetilde{\x}$ is associated to a degraded version of it, $\y$, {built} from Model \eqref{eq:model1}. Various blur kernels and noise {levels} are used, as detailed hereafter.\\
\textbf{Dataset 1:} All involved images are converted in grayscale. Each image of the database is blurred with 10 randomly generated Gaussian blurs, and then corrupted by additive noise. Thus in total, we have 1000 ($=100\times 10$) training images, 400 ($= 40\times 10$) validation images, and 300 ($=30\times 10$) test images for Dataset 1. The Gaussian blurs are of size $9\times 9$. Two of them are isotropic with standard deviation randomly generated following a uniform law within $[0.2,0.4]$. Eight of them are anisotropic with orientation either $\pi/4$ or $3 \pi/4$ (with equal probability) and vertical/horizontal widths (i.e., standard deviations of the 2D Gaussian shape) uniformly drawn within $[0.15,0.4]$. On this dataset, the noise standard deviation is set to $\sigma = 0.01$, and assumed to be known (so that blocks $(\text{NN}^k_\sigma)_{1 \leq k \leq K}$ of our architecture are overlooked).\\
\textbf{Dataset 2:} All the images are then colored ones. We degraded each of them with 15 different blurs, namely 10 Gaussian blurs (simulated using the same procedure as above), two uniform blurs with width $5 \times 5$ and $7 \times 7$ pixels, and 3 out-of-focus blurs. For the latter, the vertical and horizontal widths are set randomly within $[0.2,0.5]$, and the orientation is either $\pi/4$ or $3 \pi/4$ (with equal probability). Furthermore, for each blurred image, zero-mean Gaussian noise is added, with standard deviation 
 $\sigma$ randomly chosen, with uniform distribution over $[0.005,0.05]$. The noise level is not assumed to be known and is estimated using the proposed $\text{NN}_\sigma$ architecture. In total, we have 1500 ($=100\times 15$) training images, 600 ($=40\times 15$) validation images, and 450 ($=30\times 15$) test images, on this dataset.

Examples of blurs involved in \emph{Dataset 2} are depicted in Fig.~\ref{fig:blur}. 

\begin{figure}[h]
\centering
\begin{tabular}{c@{}c@{}c@{}c@{}c@{}c@{}}
\includegraphics[height = 1cm]{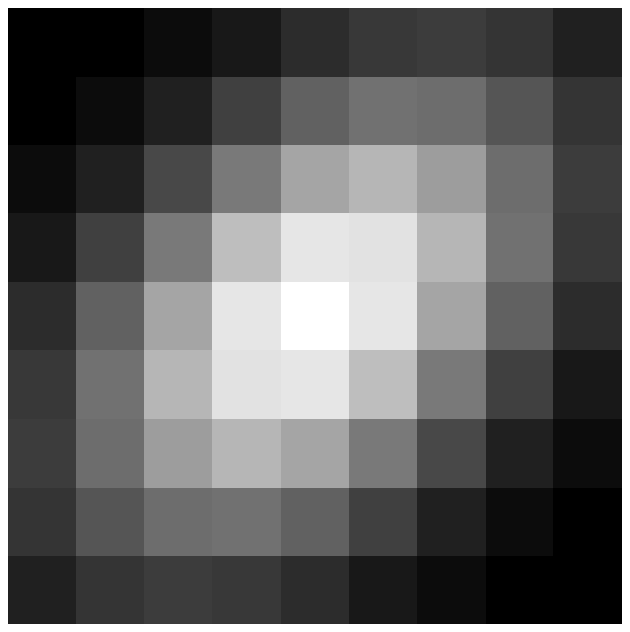}& 
\includegraphics[height = 1cm]{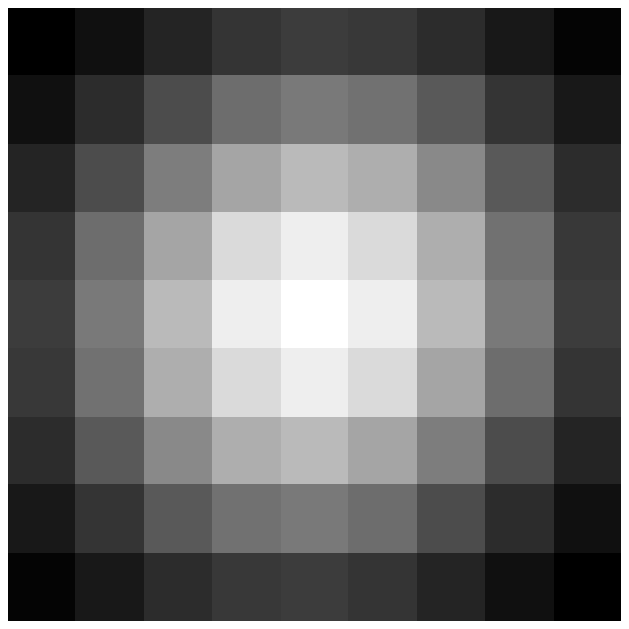}&
\includegraphics[height = 1cm]{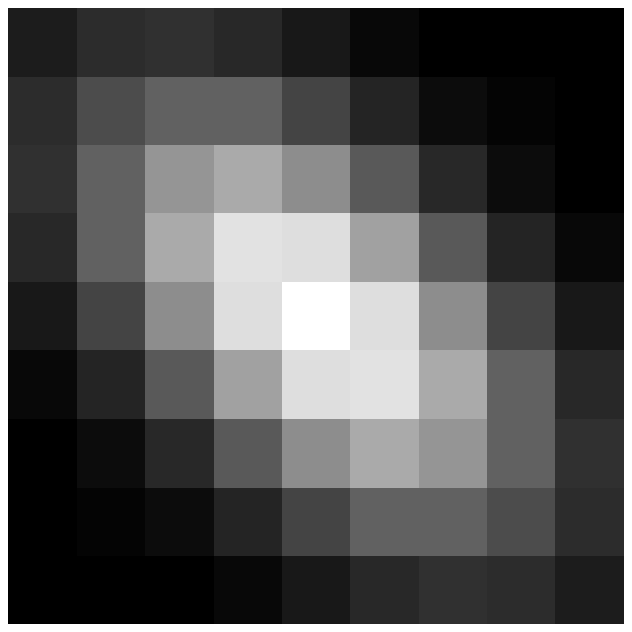}&
\includegraphics[height = 1cm]{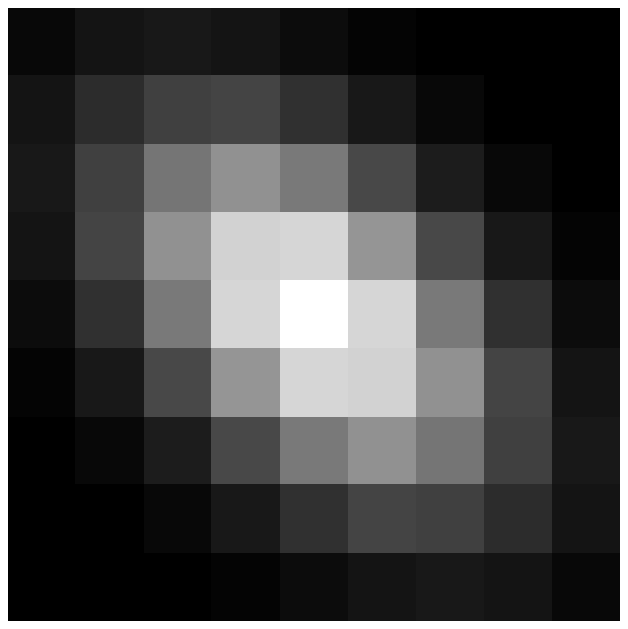}&
\includegraphics[height = 1cm]{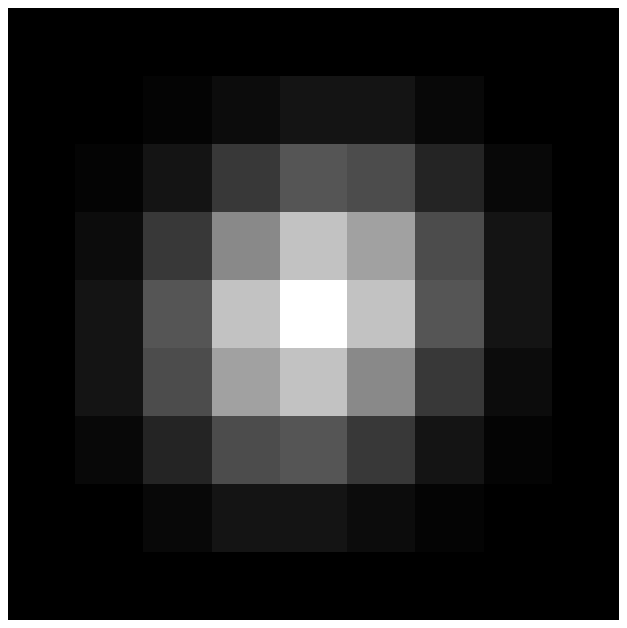} &
\includegraphics[height = 1cm]{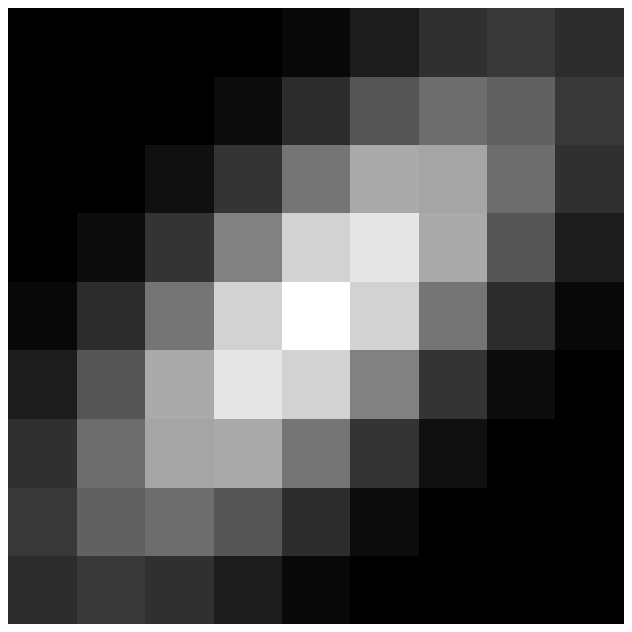}\\
\includegraphics[height = 1cm]{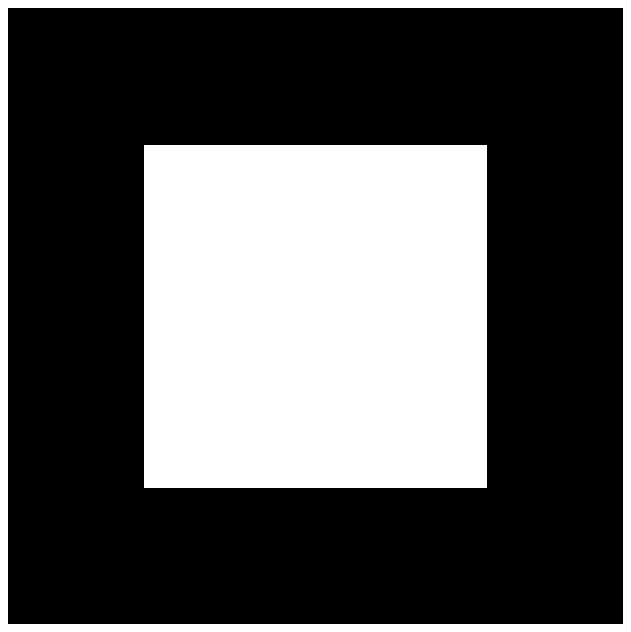}& 
\includegraphics[height = 1cm]{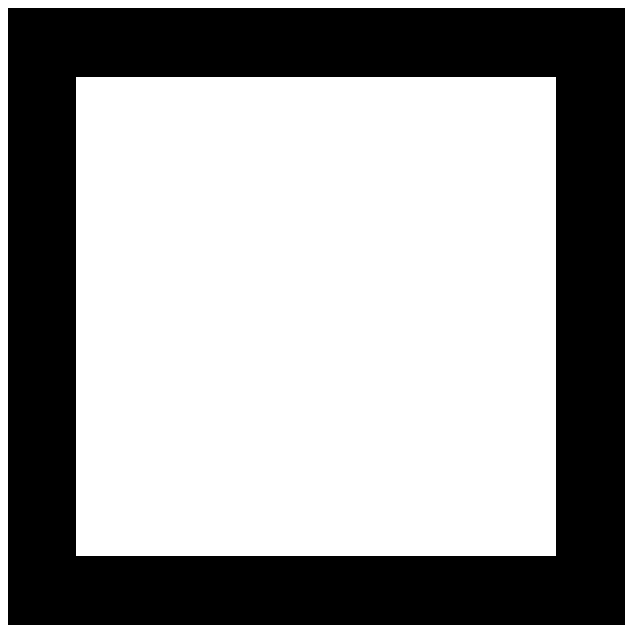}&
\includegraphics[height = 1cm]{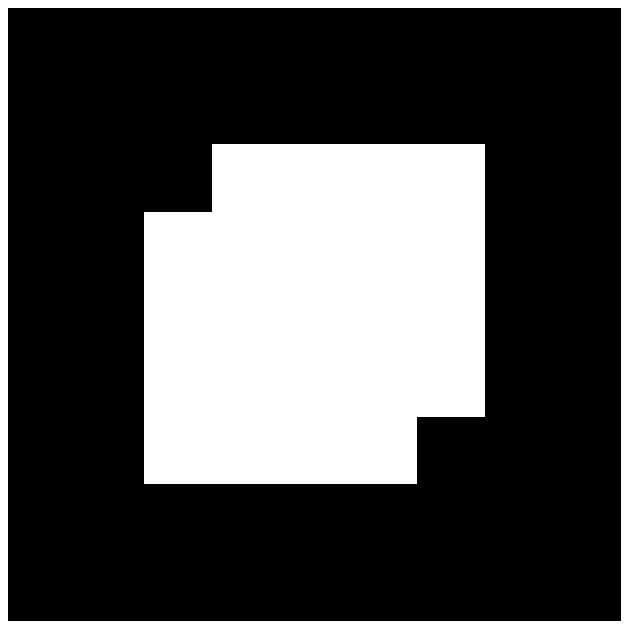}&
\includegraphics[height = 1cm]{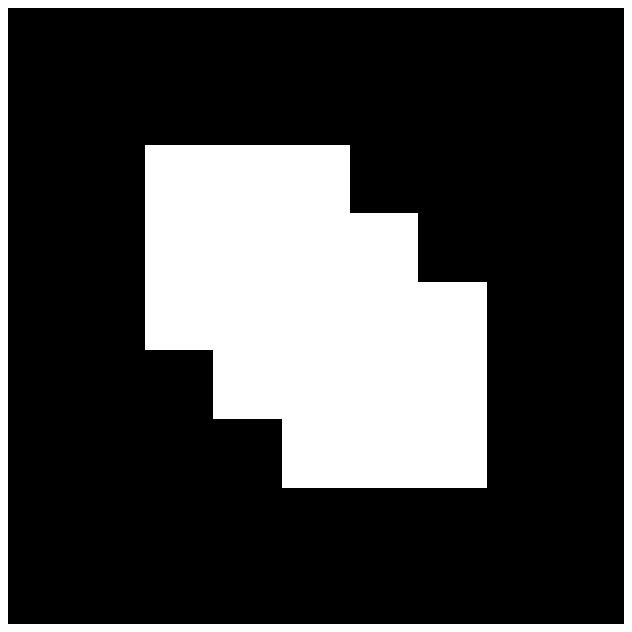}&
\includegraphics[height = 1cm]{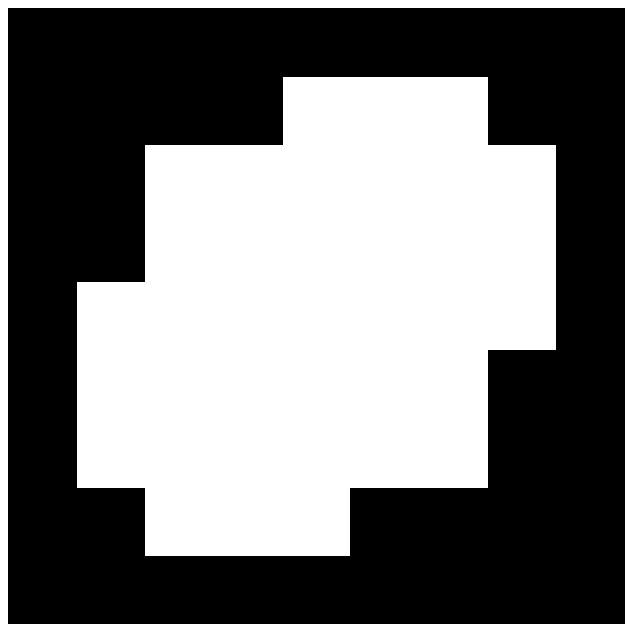} &
\includegraphics[height = 1cm]{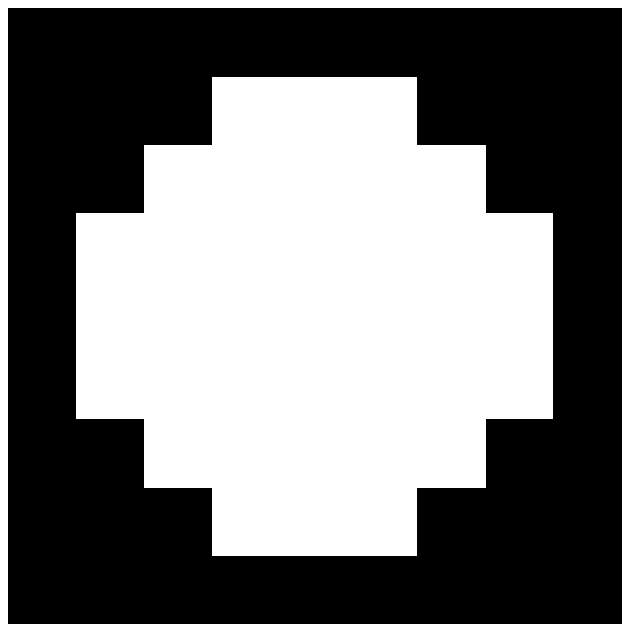}
\end{tabular}
\caption{\footnotesize Examples of blur kernels used to construct \emph{Dataset 2}.}
\label{fig:blur}
\end{figure}
%

\subsubsection{Training specifications}
We present results obtained by adopting the two training strategies described in Section~\ref{ssec:train}. For the \emph{greedy training}, we make use of the mean squared error on the estimated kernel, as a loss function for the \emph{unfoldedVBA} layers, defined as {$\ell(\widehat{\x},\widehat{\h},\widetilde{\x},\widetilde{\h}) = \|\widetilde{\h}-\widehat{\h}\|^{2}$}. The SSIM loss (\cite{Wang2004}), between the output image $\widehat{\x}$ and the ground truth $\widetilde{\x}$ is used to train the post-processing layer $\mathcal{L}_{\rm pp}$. For the \emph{end-to-end training}, we use again SSIM between $\widehat{\x}$ and $\widetilde{\x}$. We use warm initialization for \emph{end-to-end training}, that is we initialize with the weights learnt during the \emph{greedy training} phase, associated with a weight decay procedure. ADAM optimizer, with mini-batch size equal to 10, is used for all the training procedures. {Its parameters such as learning rate (lr), weight decay (wd) and epochs number are finetuned, so as to obtain stable performance on each validation set.} The number of layers $K$ (i.e., number of VBA iterations) is set during the \emph{greedy training}, and kept the same for the \emph{end-to-end training}. In practice, we increase $K$ as long as a significant decrease in the averaged MSE over the training set was observed. 
We specify in Table \ref{tab:train} all the retained settings. {The train/validation/test phase are conducted with a code implemented in Pytorch (version 1.7.0) under Python (version 3.6.10) environment, and {run} on an Nvidia DGX wokstation using one Tesla V100 SXM2 GPU (1290 MHz frequency,  32GB of RAM).} Our code is made available at \url{https://github.com/yunshihuang/unfoldedVBA}.


\begin{table}
\renewcommand{\arraystretch}{1.2}
\scriptsize
\centering
\begin{tabular}{|c|m{3.1cm}|m{3.3cm}|}
\cline{2-3}
 \multicolumn{1}{c|}{} & Dataset 1 & Dataset 2\\
\hline
\parbox[t]{2mm}{\multirow{8}{*}{\rotatebox[origin=c]{90}{Greedy training}}}
&  \multicolumn{2}{c|}{\textit{UnfoldedVBA}} \\
\cline{2-3}
& $K = 6$, epoch $= 10$ & $K=21$, epoch $= 10$\\
& lr  $= 5 \times 10^{-3}$ & lr $= 5 \times 10^{-3}$ (for $\mathcal{L}_{0}$), lr $= 10^{-3}$ (for other layers) \\
\cline{2-3}
&  \multicolumn{2}{c|}{\textit{Post-processing $\mathcal{L}_{pp}$ }}\\
\cline{2-3}
& U-net \cite{Olaf2015} & Residual network~\cite[Fig.4]{Bertocchi2019}\\
& epoch $= 30$, lr $= 10^{-3}$& epoch $= 200$, lr $= 10^{-3}$\\
\hline
\parbox[t]{6mm}{\multirow{2}{*}{\rotatebox[origin=c]{90}{ \parbox[t]{1.3cm}{End-to-end\\training}}}} & $K = 6$ & $K=21$ \\
& epoch $=6$ & epoch $=6$\\
& lr  $= 5  \times 10^{-5}$ & lr  $= 5  \times 10^{-5}$ \\
&  wd $=10^{-4}$ & wd $=10^{-4}$\\
\hline
\end{tabular}
\vspace{0.1cm}
\caption{\footnotesize Settings for the training phases in our experiments}
\label{tab:train}
\end{table}

\subsubsection{Comparison to other methods}

The proposed method is compared to several blind deconvolution approaches available in the literature:\\
\textbf{Optimization-based methods}: We first evaluate the VBA described in Section~\ref{sec:proposed}, in the favorable situation where the noise level $\sigma$ is assumed to be known, and parameter $\xi$ is finetuned empirically (see more details hereafter). {VBA is run until reaching practical convergence, i.e. when the relative squared distance between two consecutive image iterates gets lower than $10^{-5}$.} We also test two optimization-based approaches for blind deconvolution. The first one is called \emph{deconv2D}. It makes use of the proximal alternating algorithm from {\cite{Bolte2010}}, to minimize a least-squares data fidelity term combined with various priors, namely total variation and positivity constraint on the image, sum-to-one and quadratic constraint on the kernel. This method is implemented in Matlab, and inherits some of the software accelerations discussed in \cite{Abboud2019} for blind video deconvolution. The second competitor in this category is the \emph{blinddeconv} approach \footnote{Matlab code: \url{https://dilipkay.wordpress.com/blind-deconvolution/}}  from \cite{Krishnan2011}. For the sake of fair comparisons, for both {datasets}, we {finetune} the hyperparameters of these three methods on 40\% of the training set and apply an average of the found values on the test set. Moreover, following the use of} these three methods, we perform a non-blind deconvolution step BM3D-DEB \footnote{Matlab code: \url{https://webpages.tuni.fi/foi/GCF-BM3D/index.html#ref_software}} \cite{Lebrun2012}, which uses their respective estimated blur kernel to restore the image.
\\
\textbf{Deep learning methods:} We perform comparisons with three recent deep learning architectures for blind deconvolution. SelfDeblur \footnote{Python/Pytorch code: \url{https://github.com/csdwren/SelfDeblur}} \cite{Ren2020} is an unsupervised approach able to jointly perform the image restoration and kernel estimation tasks.  DBSRCNN \footnote{Python/Pytorch code: \url{https://github.com/Fatma-ALbluwi/DBSRCNN}} \cite{Albluwi2018} and DeblurGAN \footnote{Python/Pytorch (training) and Matlab C-mex (testing) codes: \url{https://github.com/KupynOrest/DeblurGAN}} \cite{Kupyn2018} are two supervised deep learning techniques. In {contrast} with SelfDeblur, they both only provide the estimated image, but do not estimate the kernel. Both these methods have been retrained on each of our datasets, using the same settings as in their initial implementation. Moreover, we adapted DBSRCNN to color images using the same pipeline as for our method, that is applying DBSRCNN on the luminance channel while simply nonlinearly filtering the chrominance ones. 

\subsubsection{Evaluation metrics}
All the methods are evaluated in terms of their performance on the blur kernel estimation (when available) and on the image restoration. Different metrics are used. For the blur kernels, we evaluate (\textit{i}) the MSE, (\textit{ii}) the so-called $\mathcal{H}_{\infty}$ error defined as the $\ell_{\infty}$ norm of the difference between the 2D discrete Fourier coefficients (with suitable padding) of the estimated and of the true kernel, and (\textit{iii}) the mean absolute error (MAE) defined as the $\ell_1$ norm of the difference between $\widetilde{\h}$ and $\widehat{\h}$. For evaluating the image quality, we compute (\textit{i}) the SSIM, (\textit{ii}) the PSNR (Peak-Signal-to-Noise Ratio), and (\textit{iii}) the PieAPP value \cite{Prashnani2018}, between the estimated image $\widehat{\x}$ and the ground truth $\widetilde{\x}$. 

\subsection{Experimental results}

\subsubsection{Dataset 1}

\begin{figure}[H]
\centering
\begin{tabular}{@{}c@{}c@{}}
\includegraphics[width = 4.2cm]{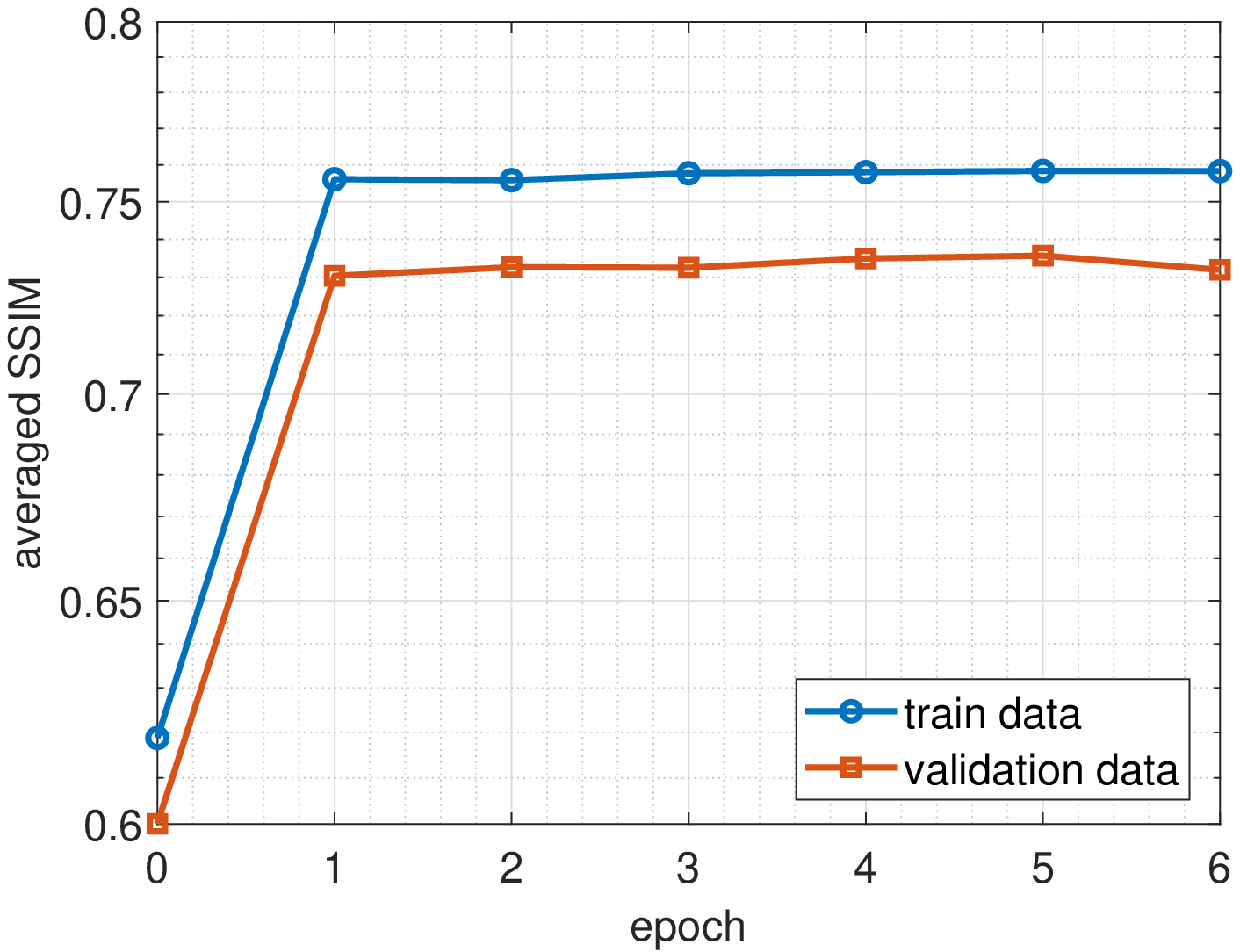}  
&
\includegraphics[width = 4.2cm]{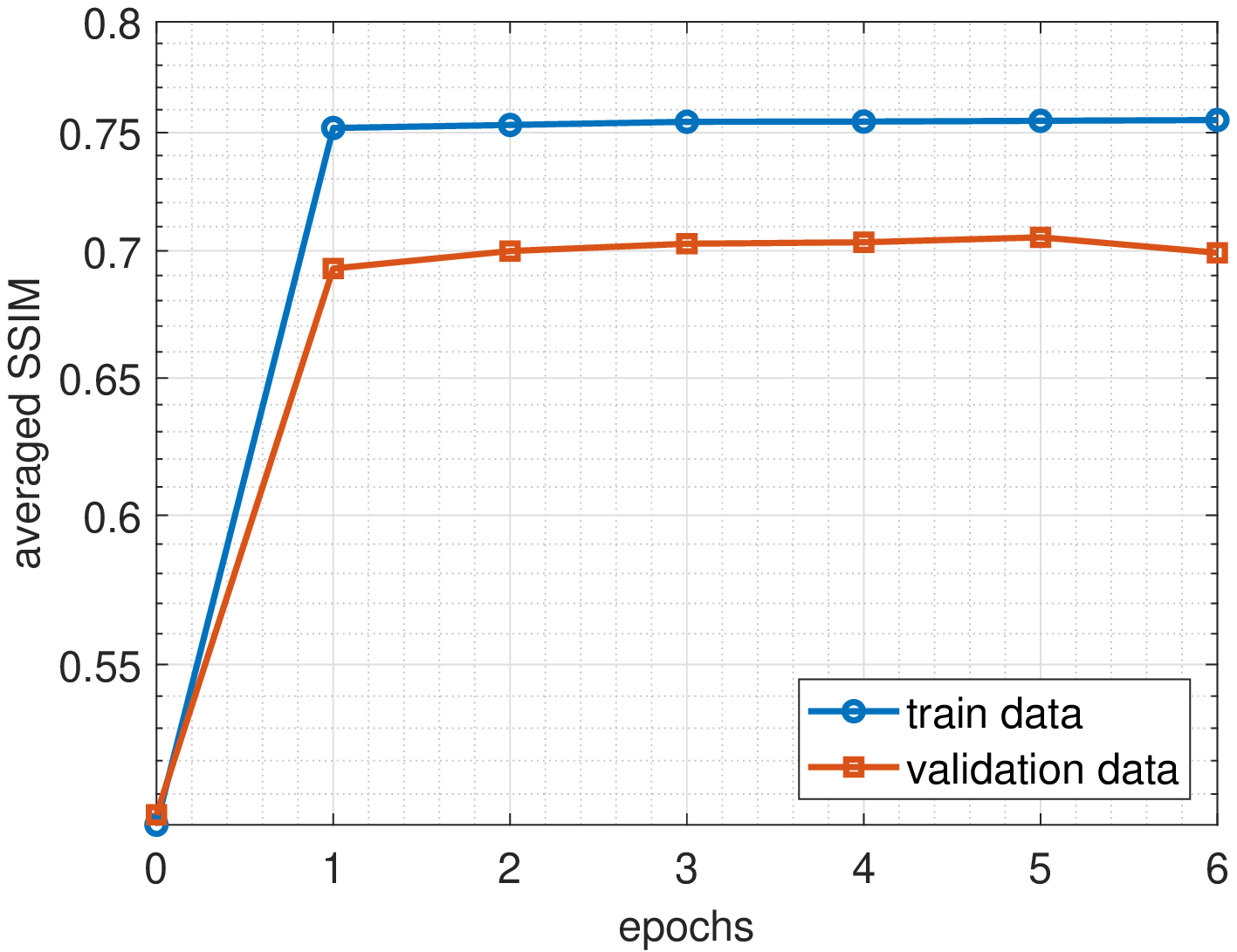}  
\end{tabular}
\caption{\footnotesize Evolution of SSIM loss along epochs of \emph{end-to-end training} phase, averaged either on training or on validation sets of \emph{Dataset 1} (left) and \emph{Dataset 2} (right). }
\label{fig:SSIM_loss}
\end{figure}

\begin{figure*}[h]
\footnotesize
\begin{tabular}{c@{}c@{}c@{}c@{}c@{}}
  \centering
\includegraphics[height=3cm, trim = {1cm 1cm 1cm 1cm}, clip]{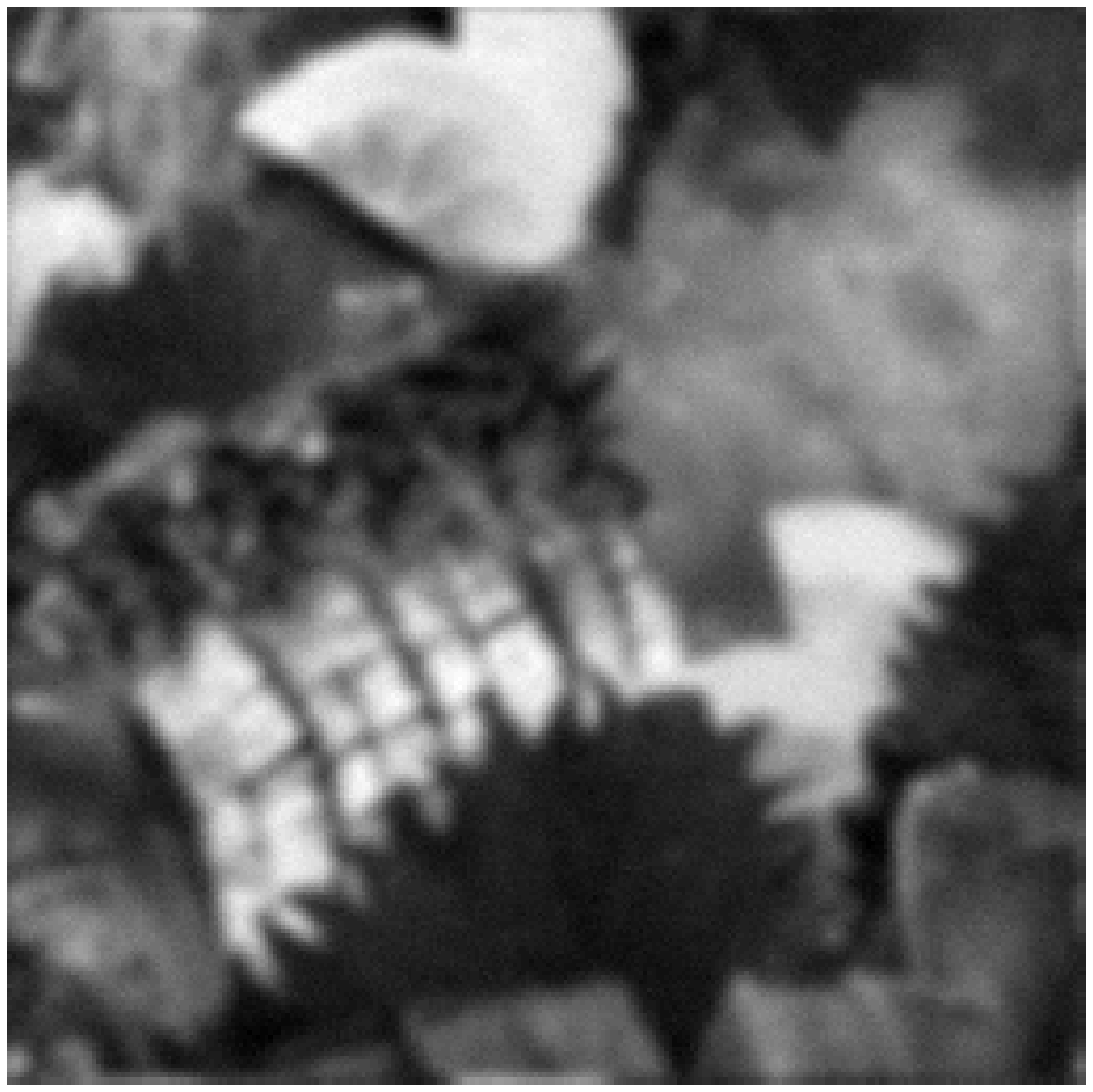} &\hspace{-0.7cm}
\begin{overpic}
[height=3cm, trim = {1cm 1cm 1cm 1cm}, clip]{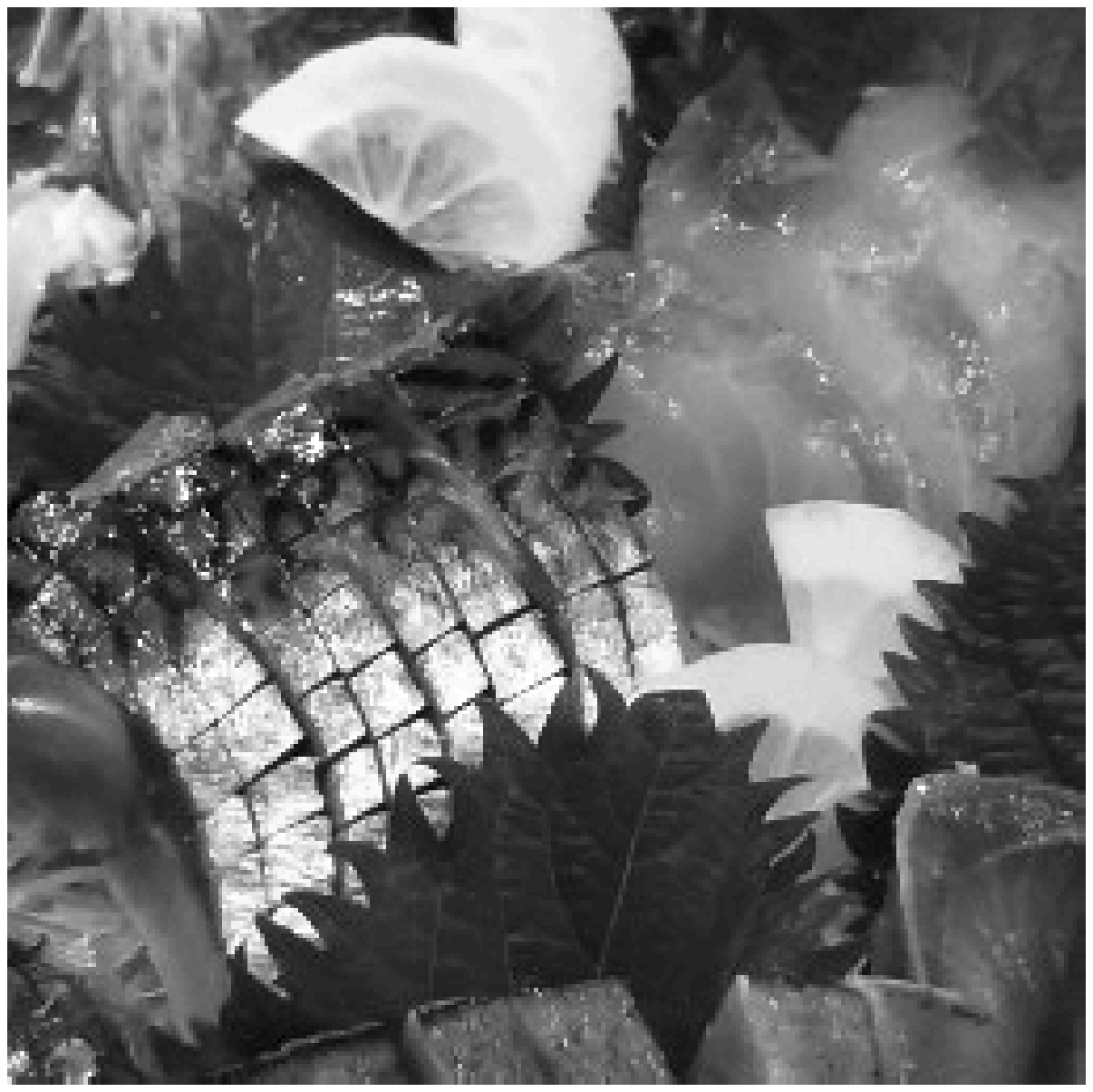}
\put(22,6)
{\includegraphics[height=1cm,trim={4cm 1.5cm 3.4cm 1cm},clip]{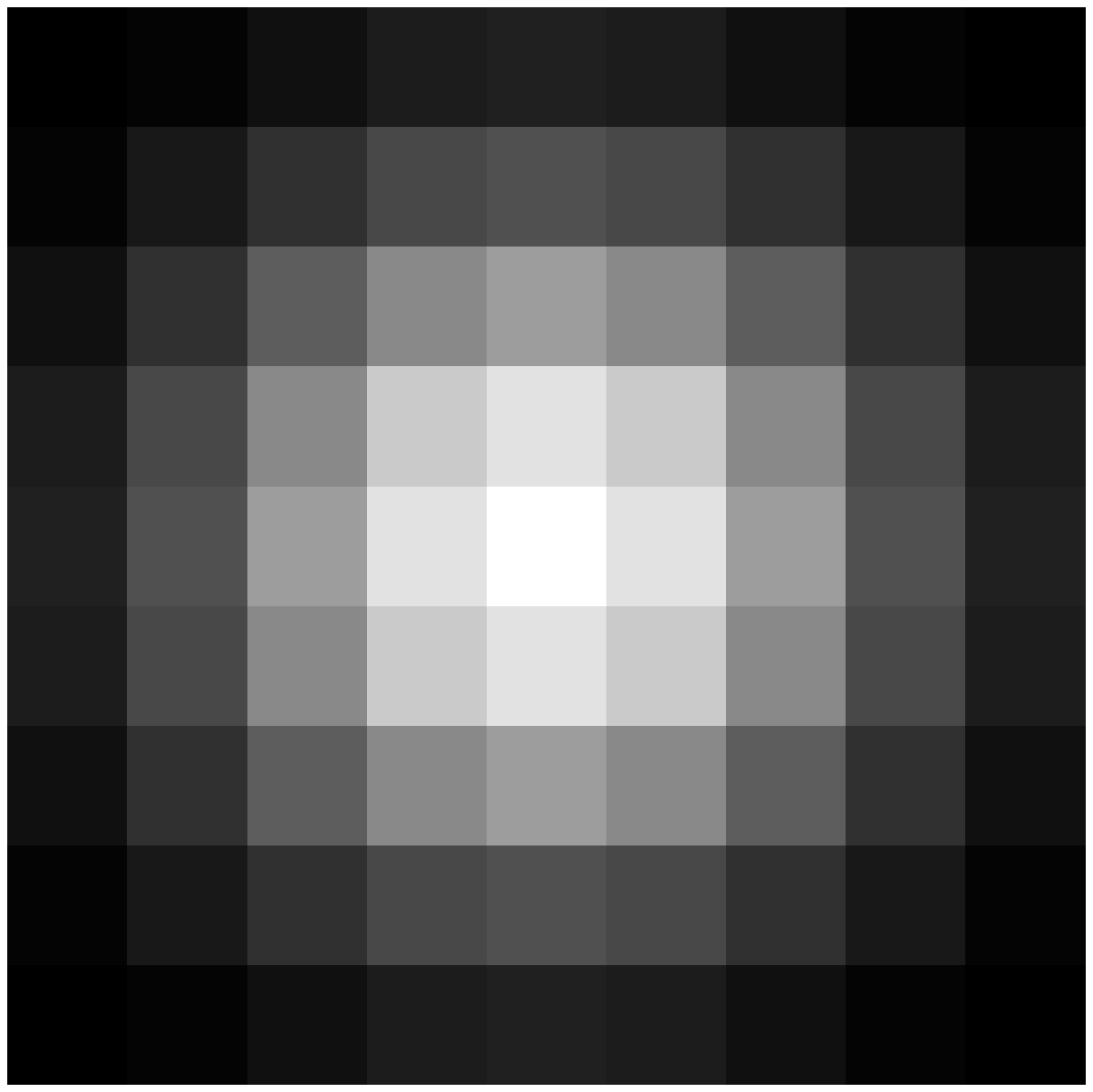}}
\end{overpic}&\hspace{-0.7cm}

\begin{overpic}
[height=3cm, trim = {1cm 1cm 1cm 1cm}, clip]{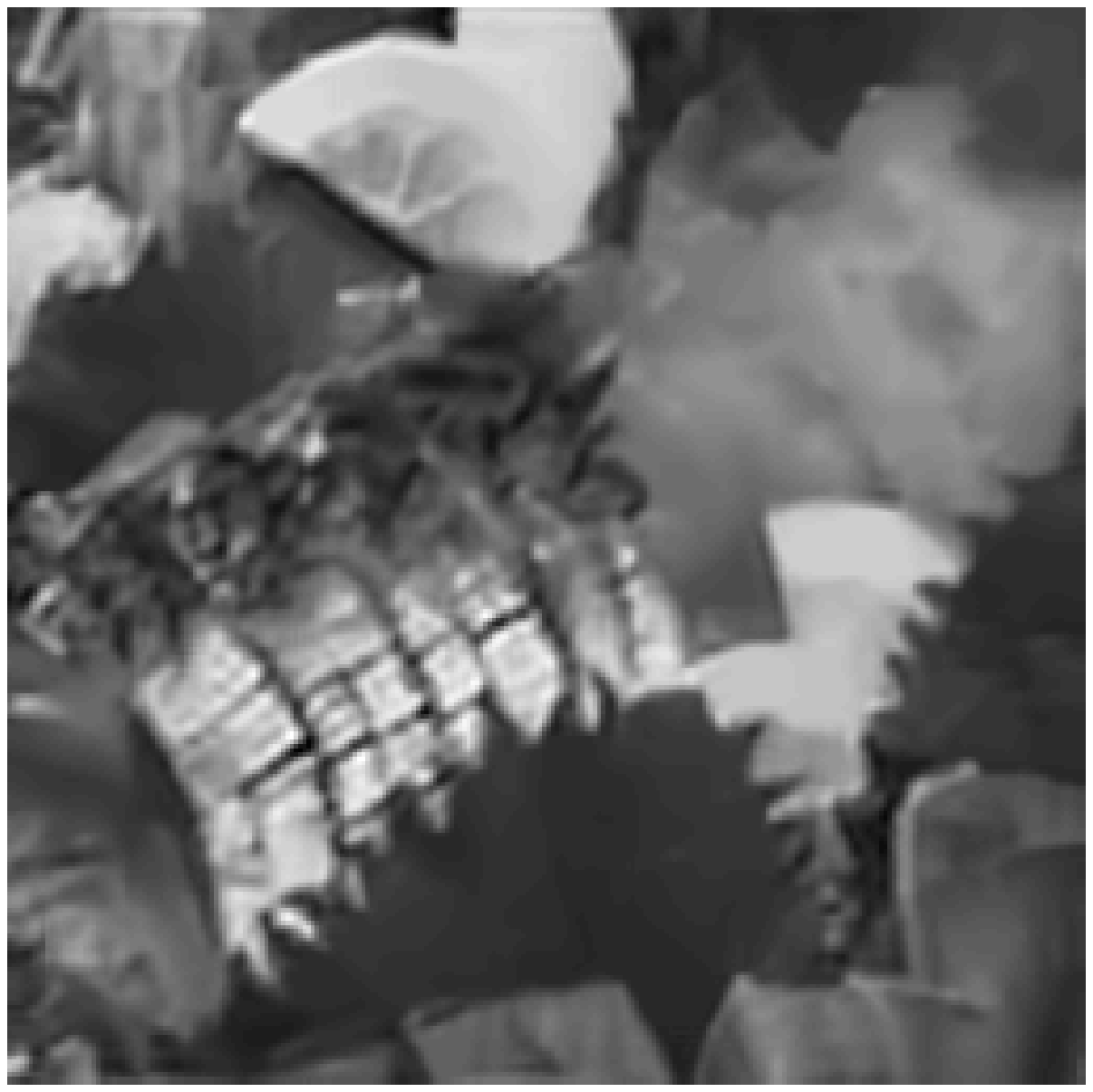}
\put(22,6)
{\includegraphics[height=1cm,trim={4cm 1.5cm 3.4cm 1cm},clip]{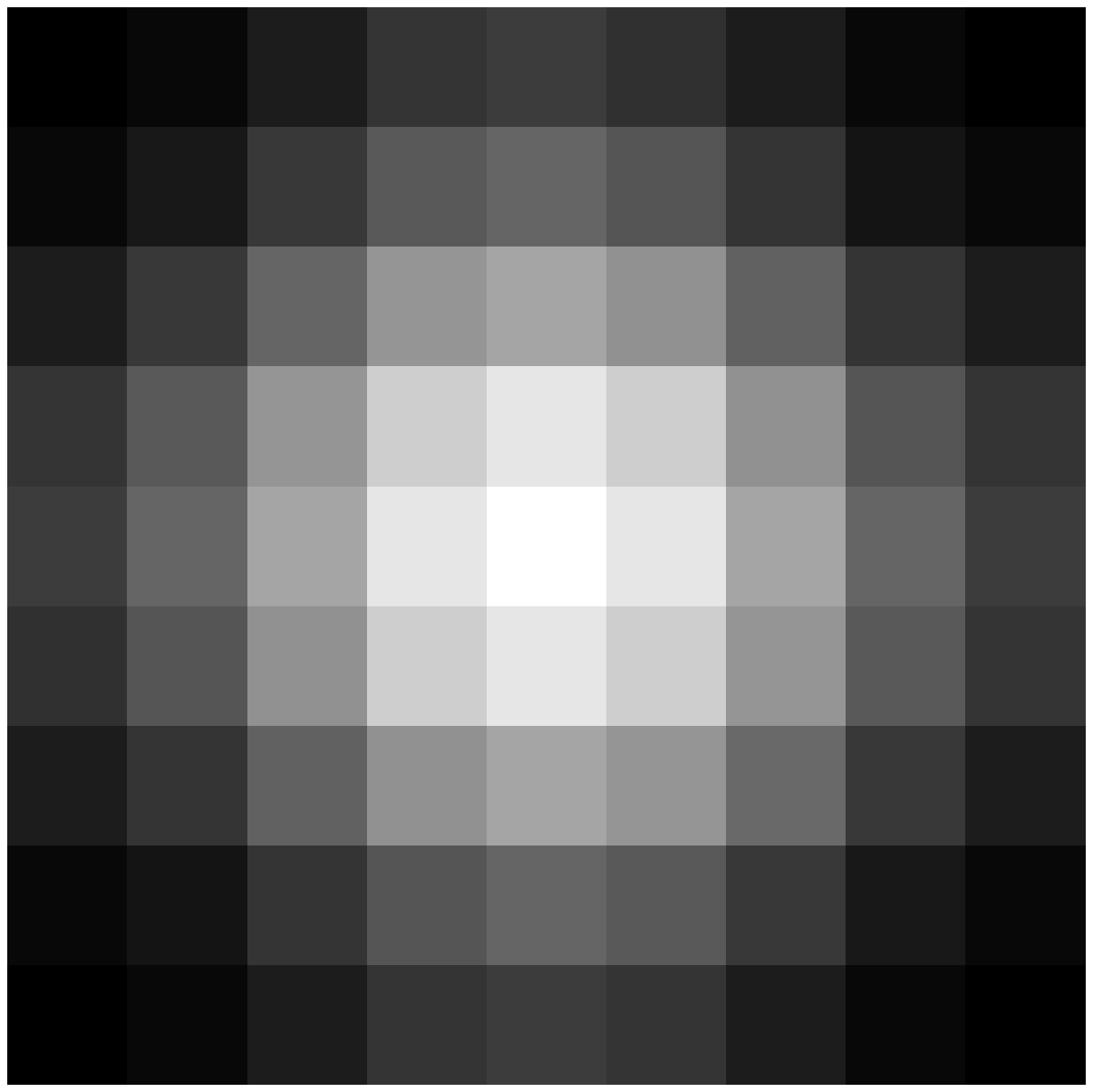}}
\end{overpic}&\hspace{-0.7cm}

\begin{overpic}
[height=3cm, trim = {1cm 1cm 1cm 1cm}, clip]{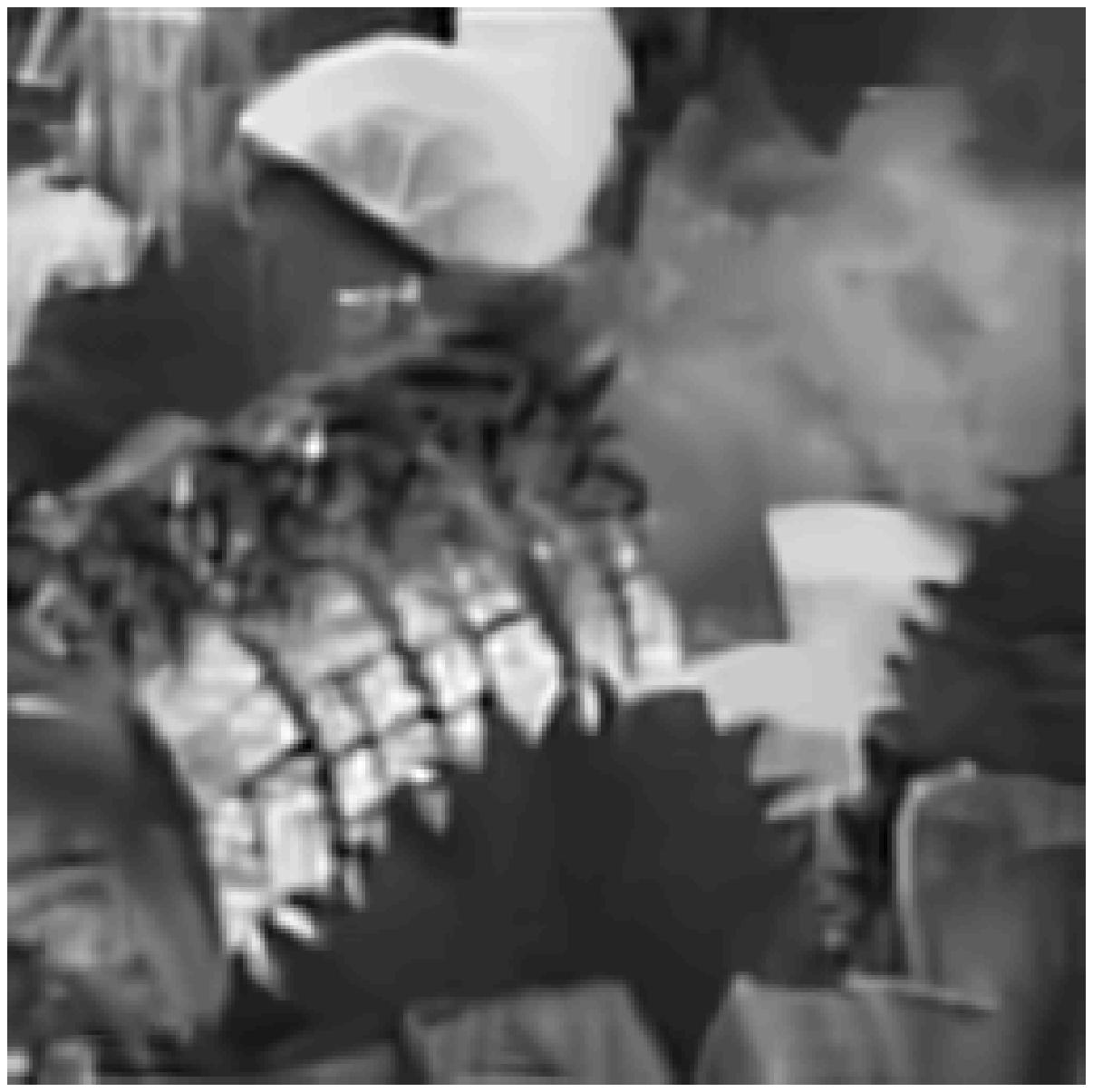}
\put(22,6)
{\includegraphics[height=1cm,trim={4cm 1.5cm 3.4cm 1cm},clip]{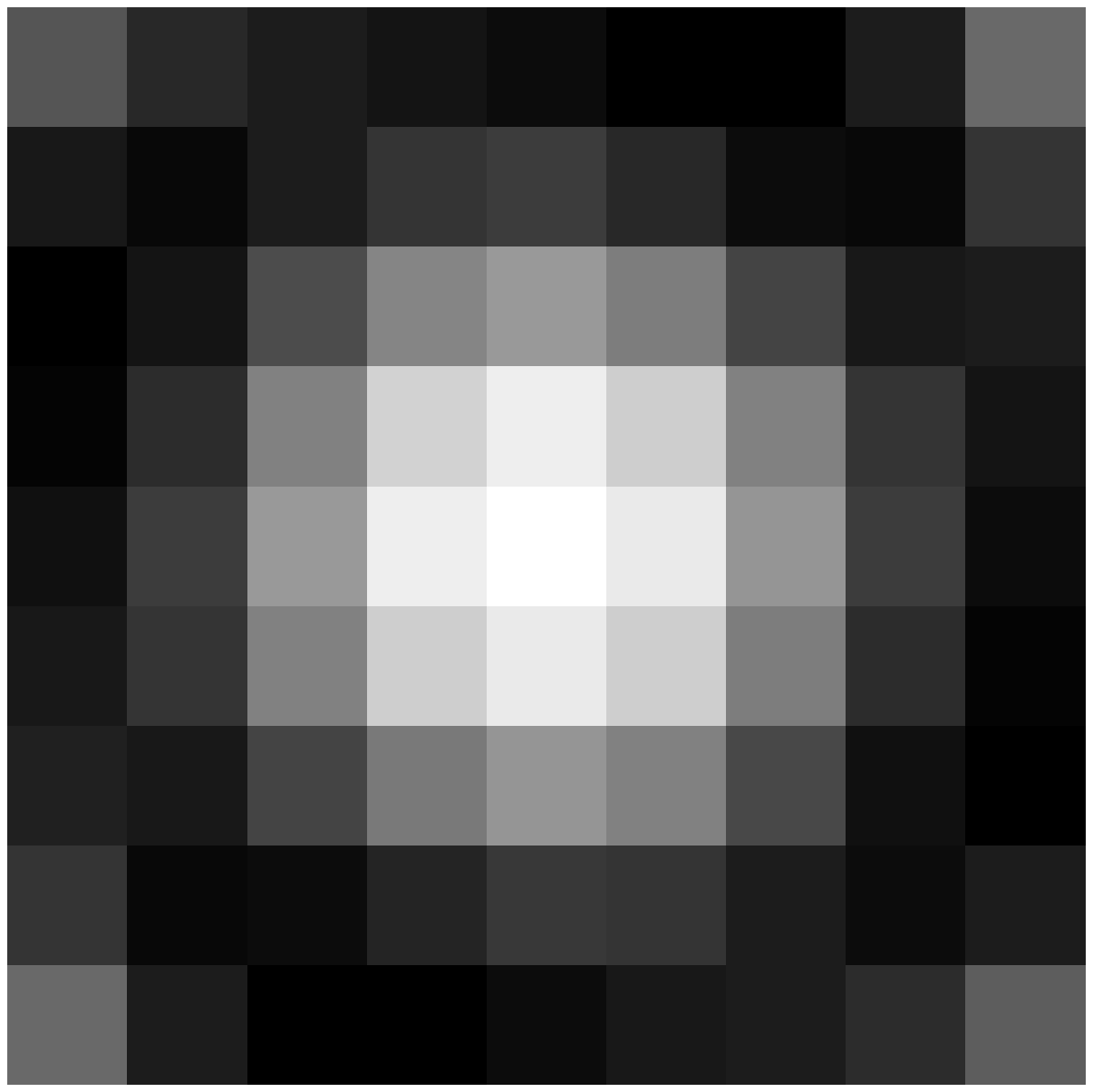}}
\end{overpic}&\hspace{-0.7cm}

\begin{overpic}
[height=3cm, trim = {1cm 1cm 1cm 1cm}, clip]{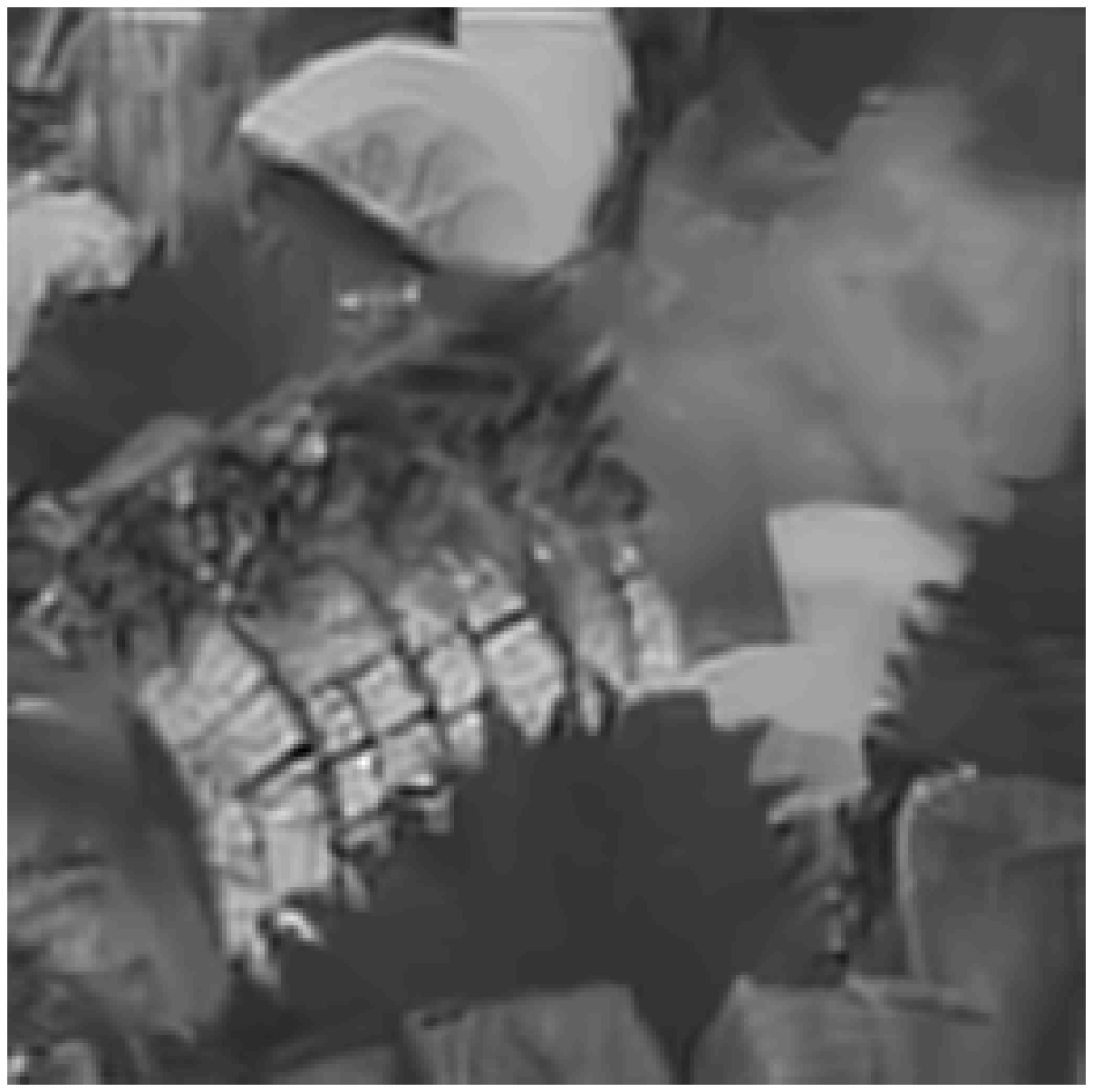}
\put(22,6)
{\includegraphics[height=1cm,trim={4cm 1.5cm 3.4cm 1cm},clip]{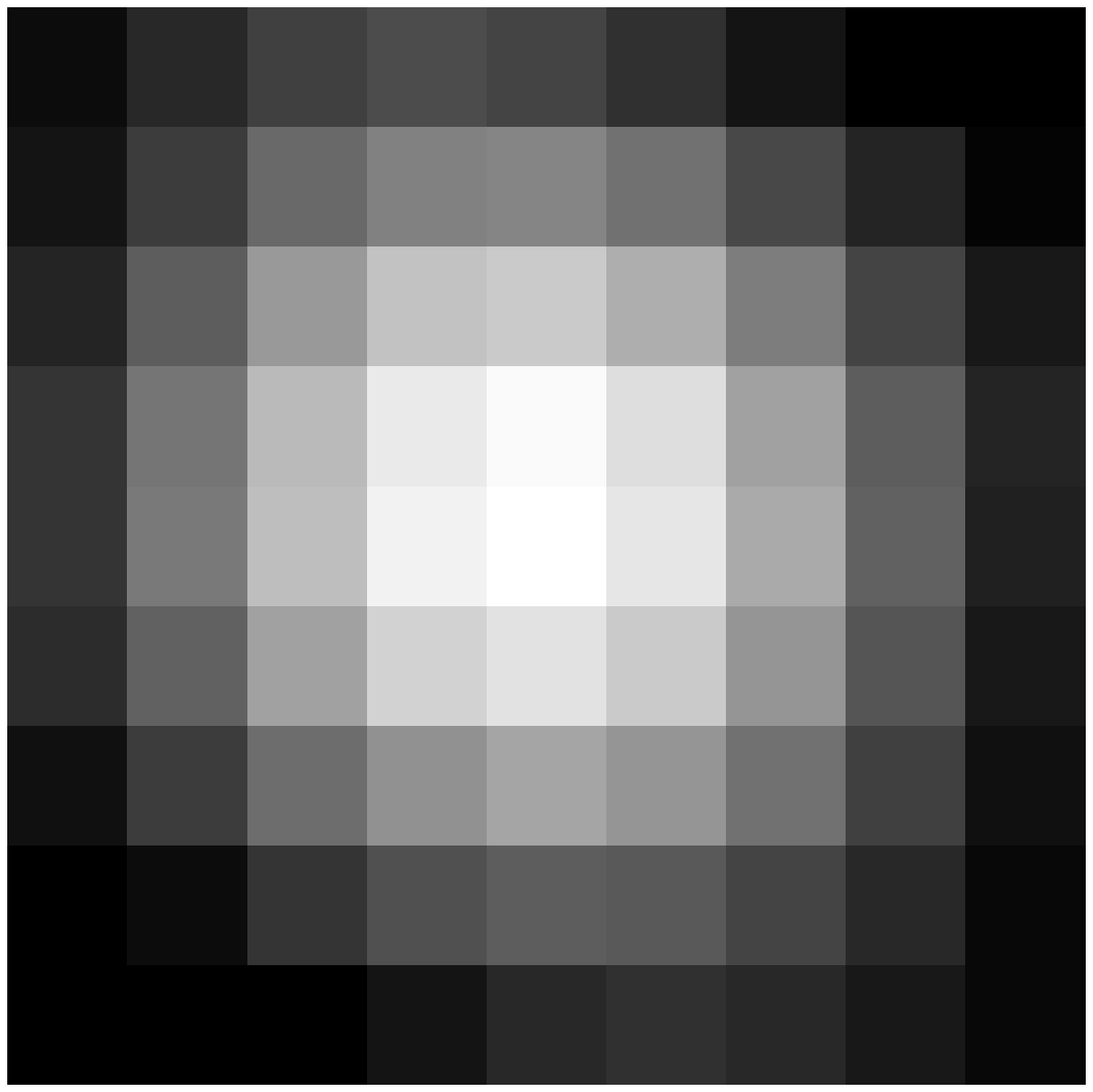}}
\end{overpic}\\

\textbf{Degraded} & \textbf{Original} & \textbf{VBA} & \textbf{deconv2D} & \textbf{blinddeconv}\\
& & MSE = 0.0010& MSE = 0.0016& MSE = $5.3788\times10^{-4}$   \\  
& PieAPP = 4.7468& PieAPP = 1.6914&PieAPP = 2.7365&PieAPP = 1.7290\\

\begin{overpic}
[height=3cm, trim = {1cm 1cm 1cm 1cm}, clip]{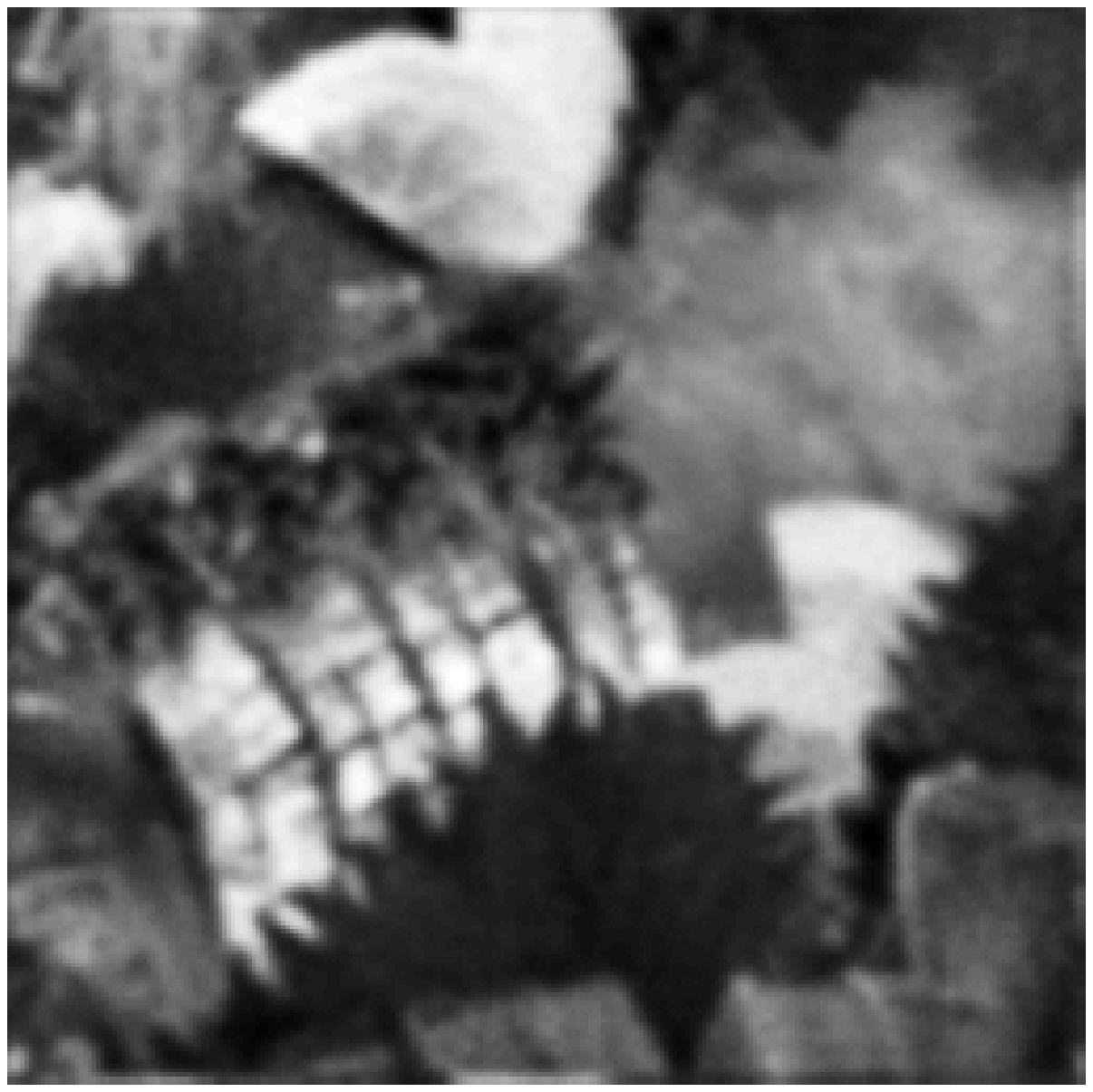}
\put(22,6)
{\includegraphics[height=1cm,trim={4cm 1.5cm 3.4cm 1cm},clip]{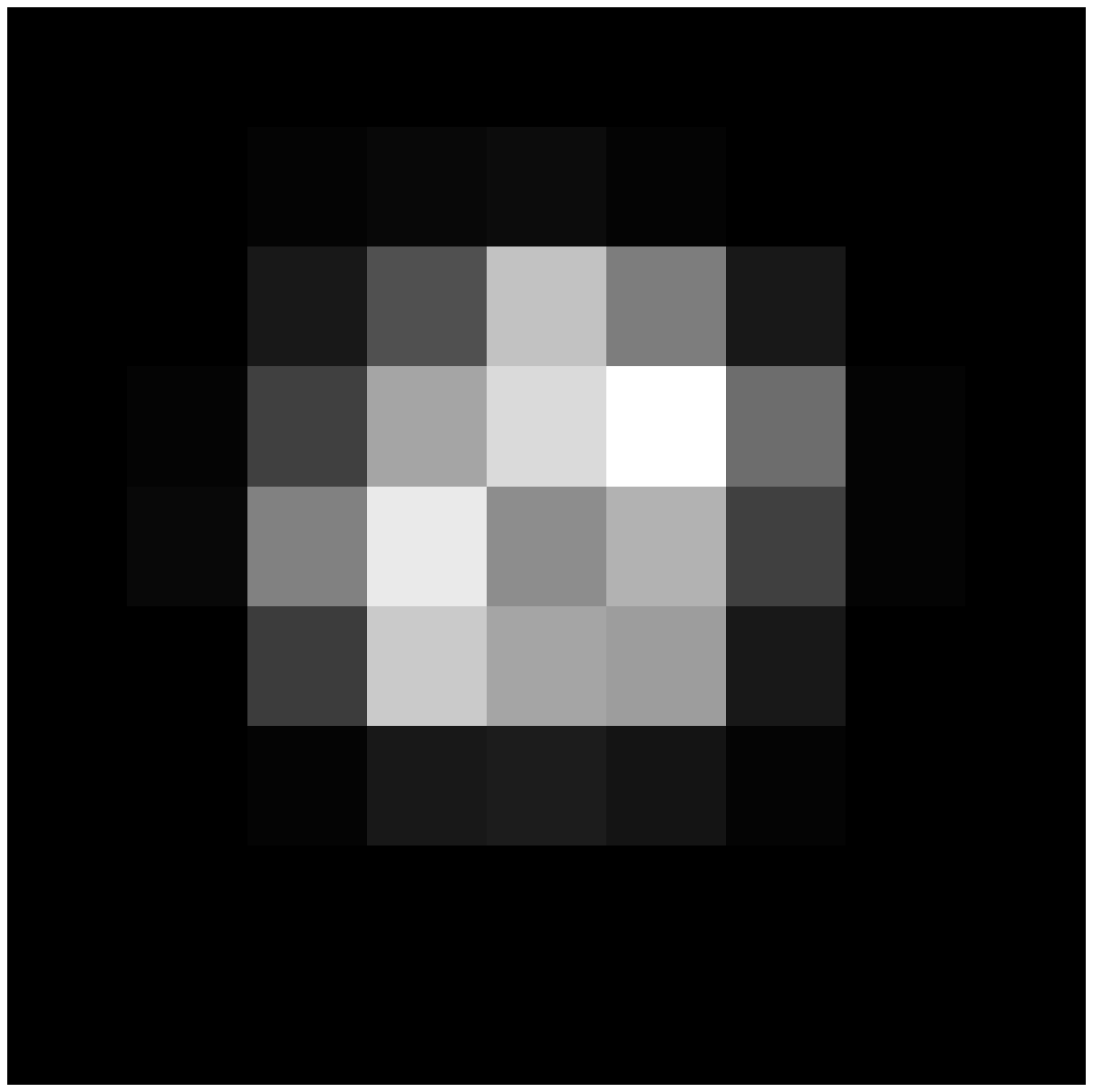}}
\end{overpic}&\hspace{-0.7cm}

\includegraphics[height=3cm, trim = {1cm 1cm 1cm 1cm}, clip]{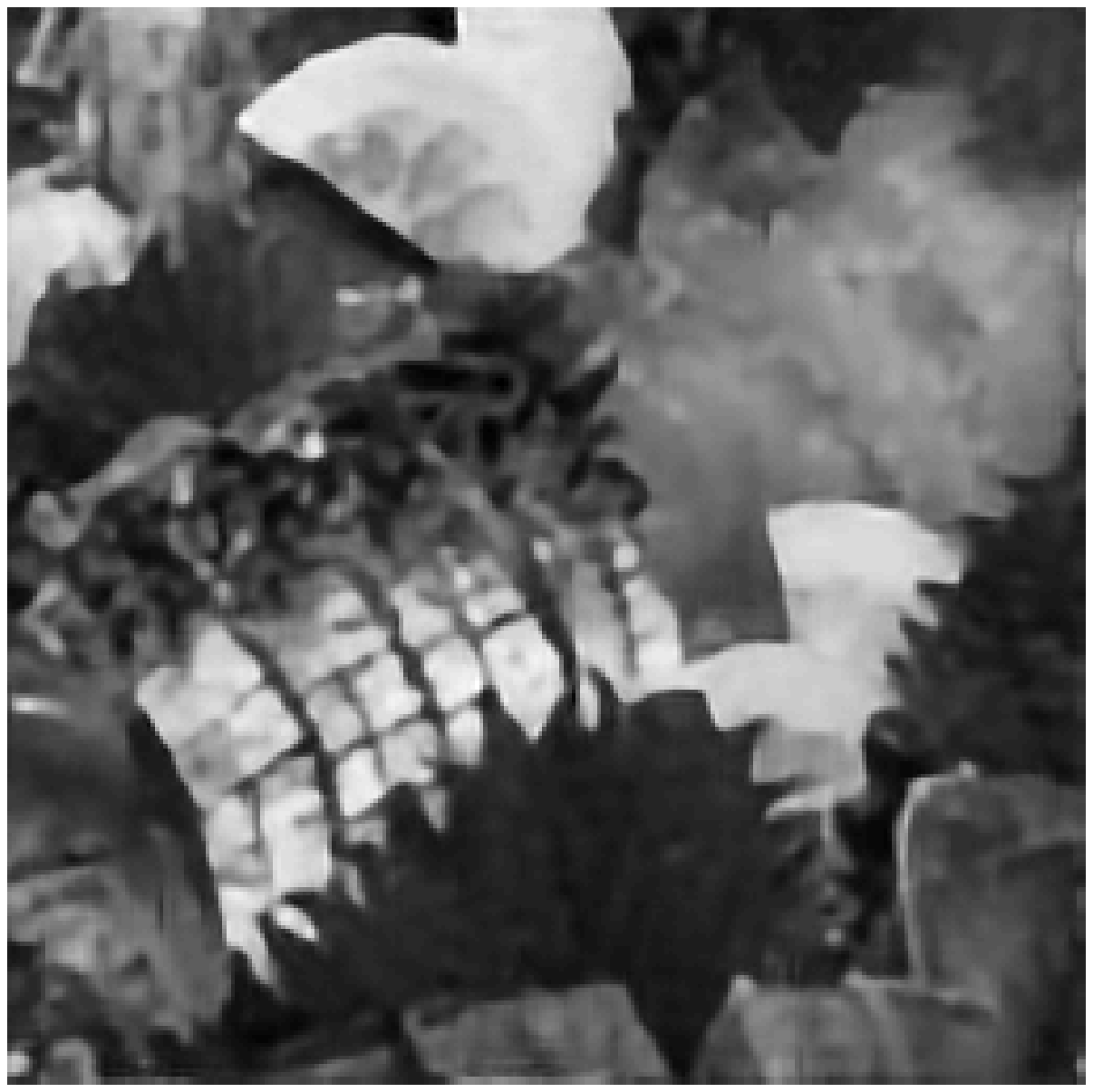}
&\hspace{-0.7cm}

\includegraphics[height=3cm, trim = {1cm 1cm 1cm 1cm}, clip]{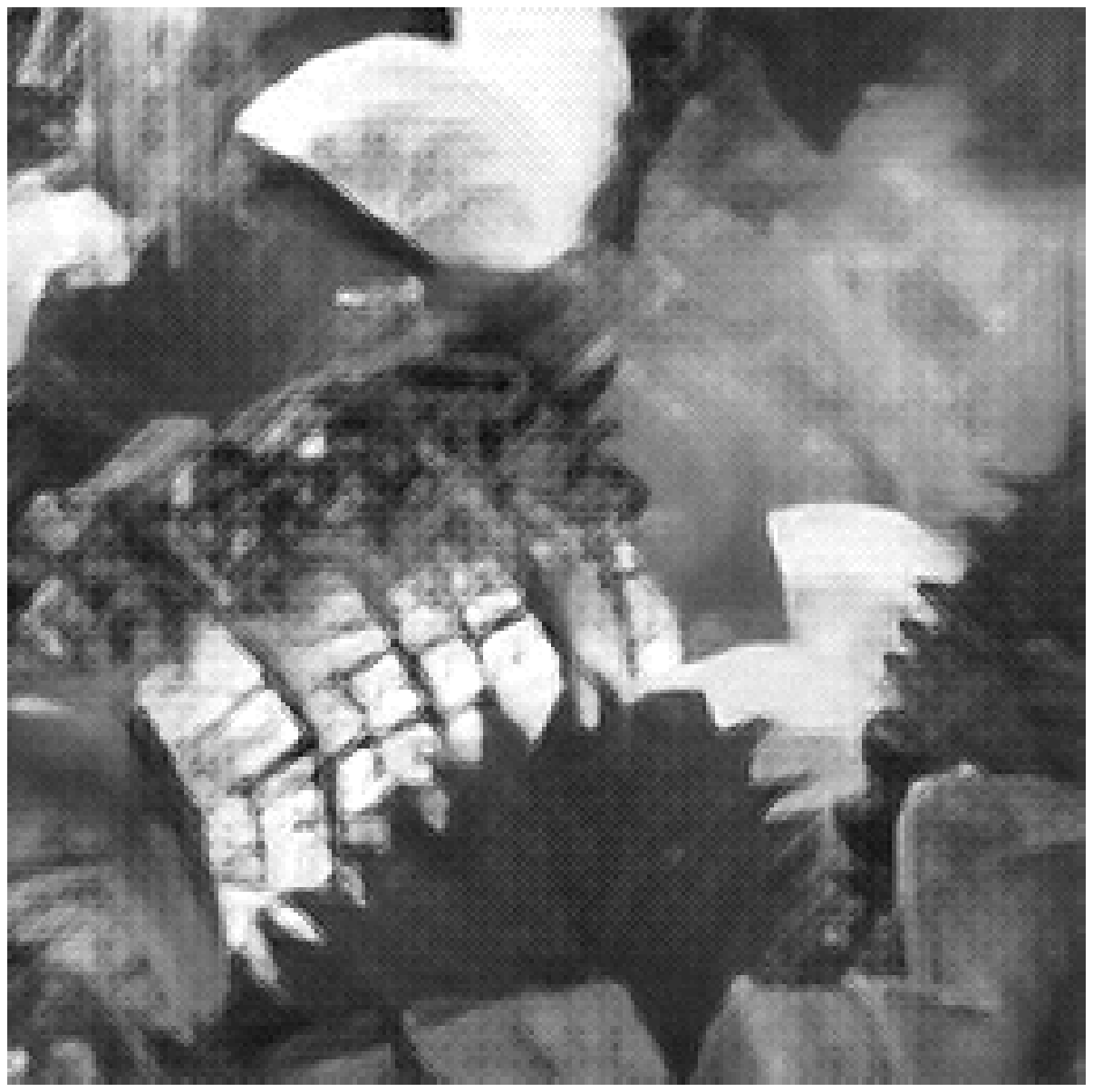}
&\hspace{-0.7cm}

\begin{overpic}
[height=3cm, trim = {1cm 1cm 1cm 1cm}, clip]{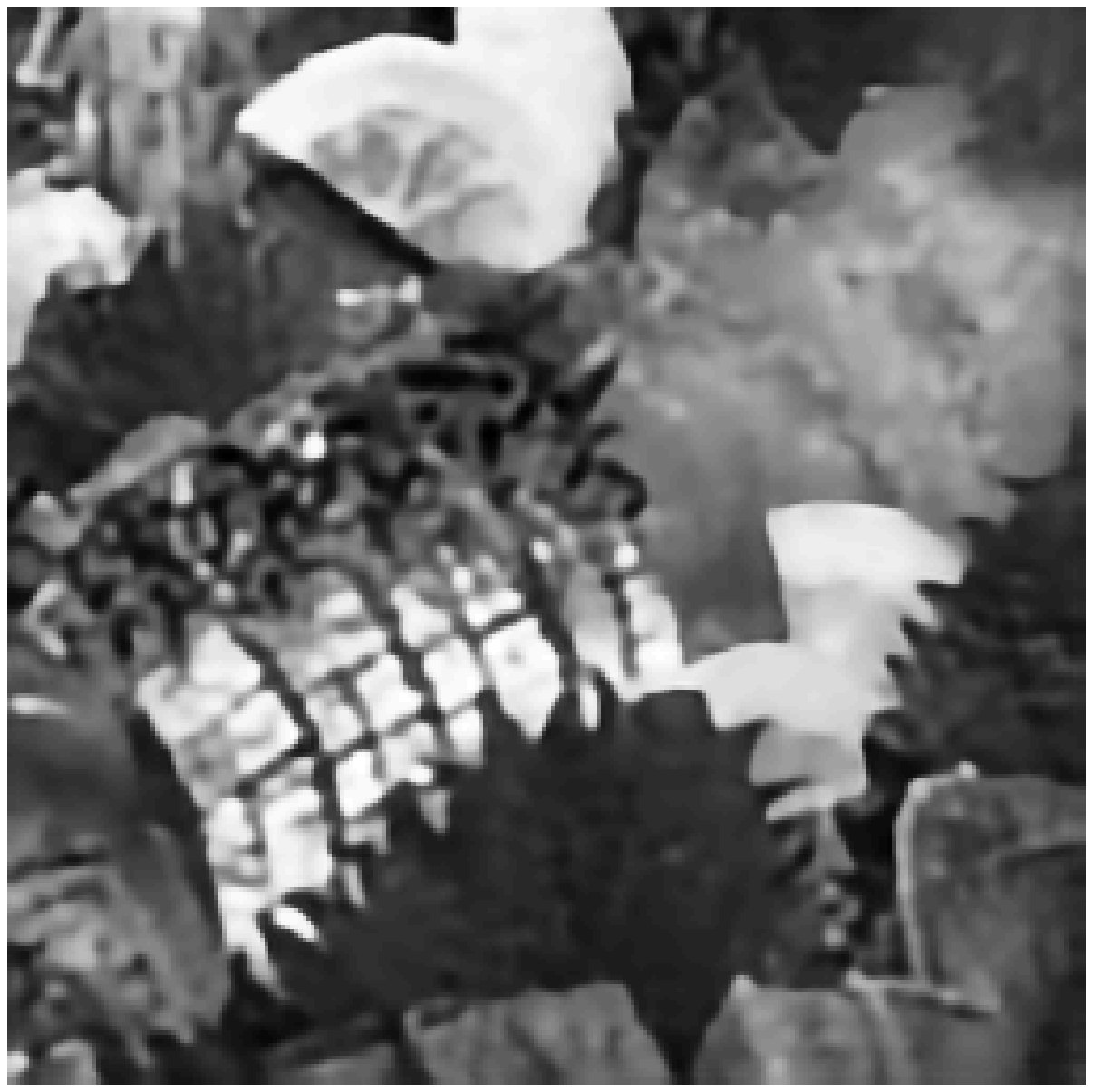}
\put(22,6)
{\includegraphics[height=1cm,trim={4cm 1.5cm 3.4cm 1cm},clip]{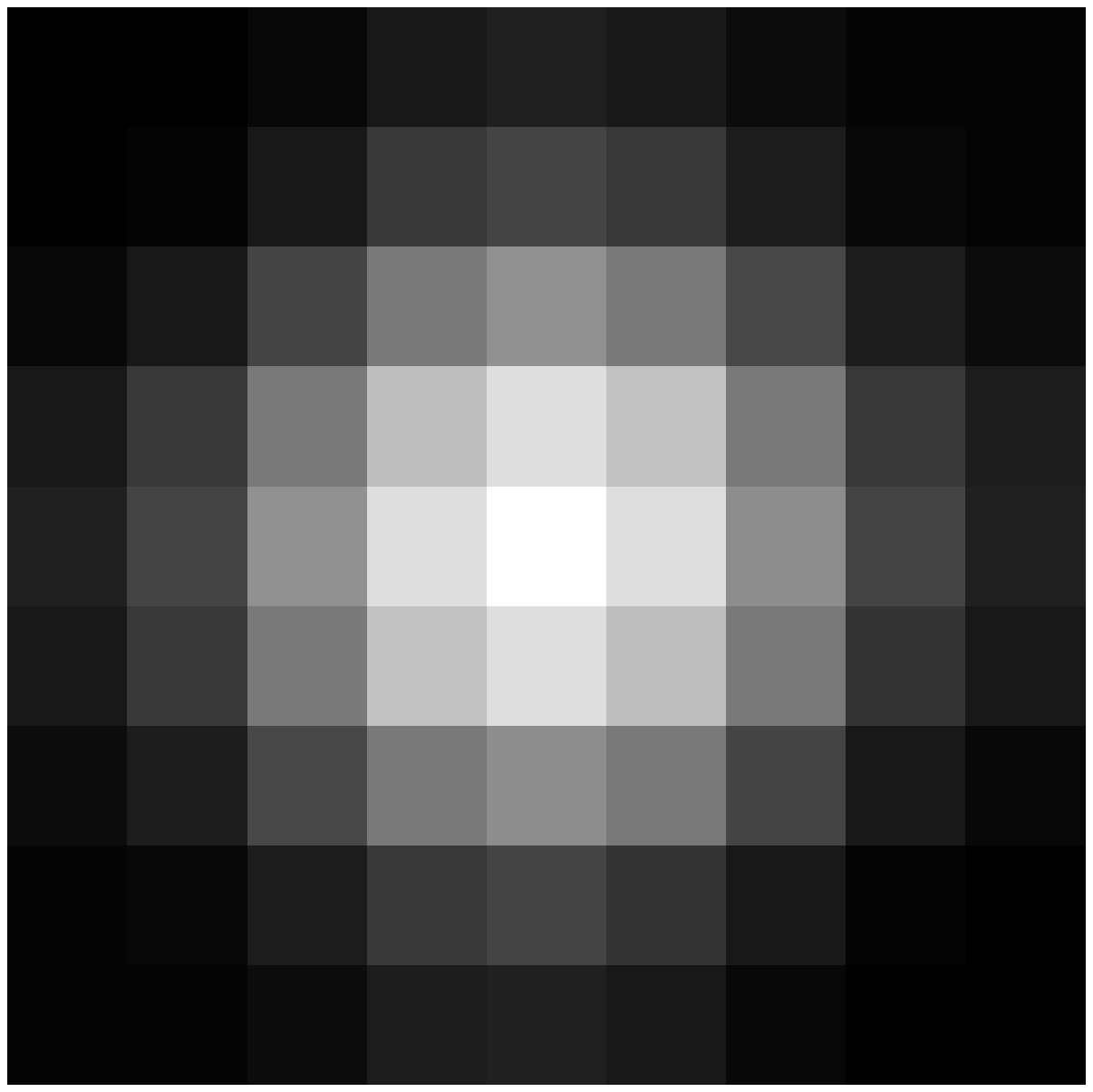}}
\end{overpic}&\hspace{-0.7cm}

\begin{overpic}
[height=3cm, trim = {1cm 1cm 1cm 1cm}, clip]{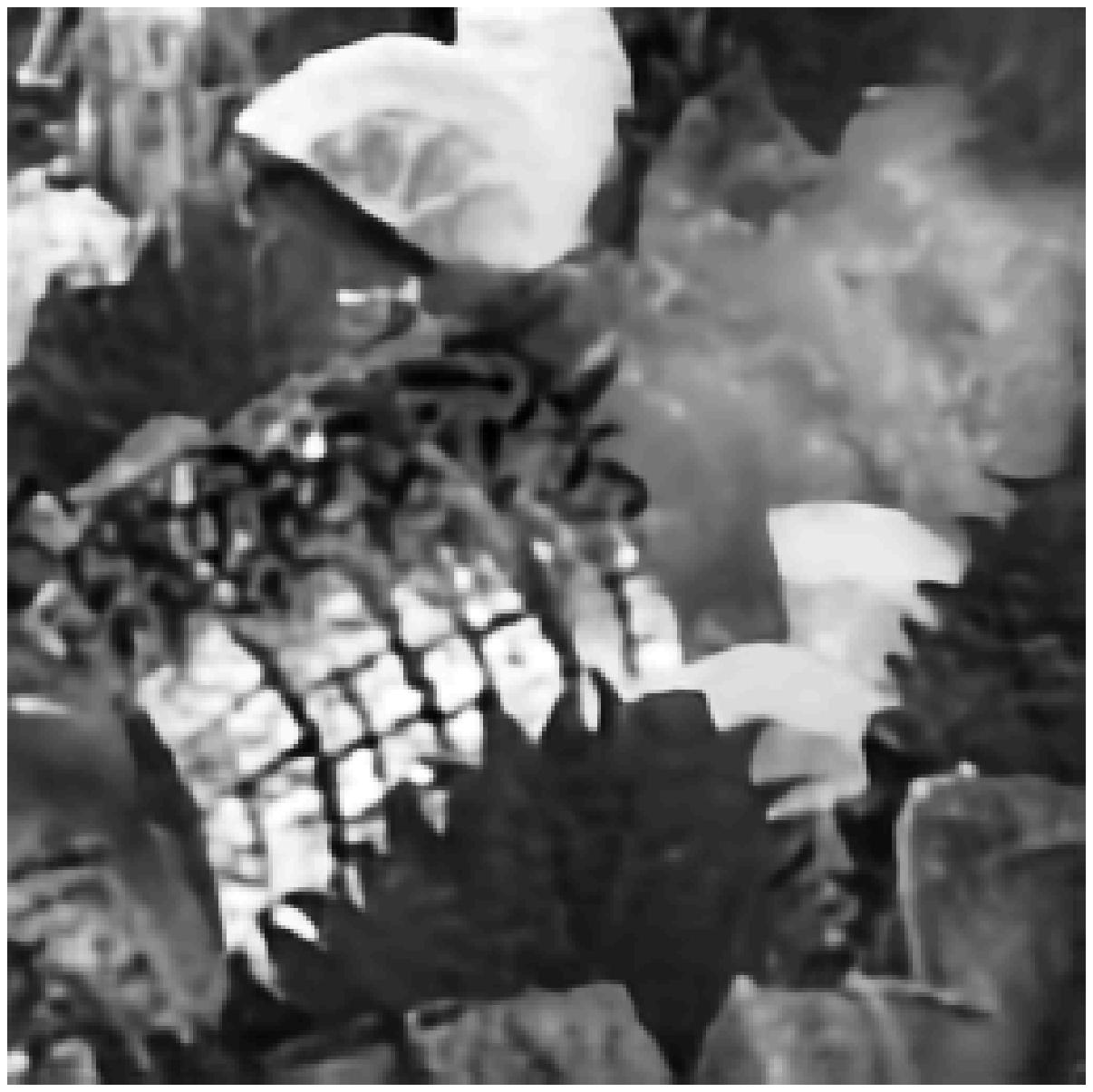}
\put(22,6)
{\includegraphics[height=1cm,trim={4cm 1.5cm 3.4cm 1cm},clip]{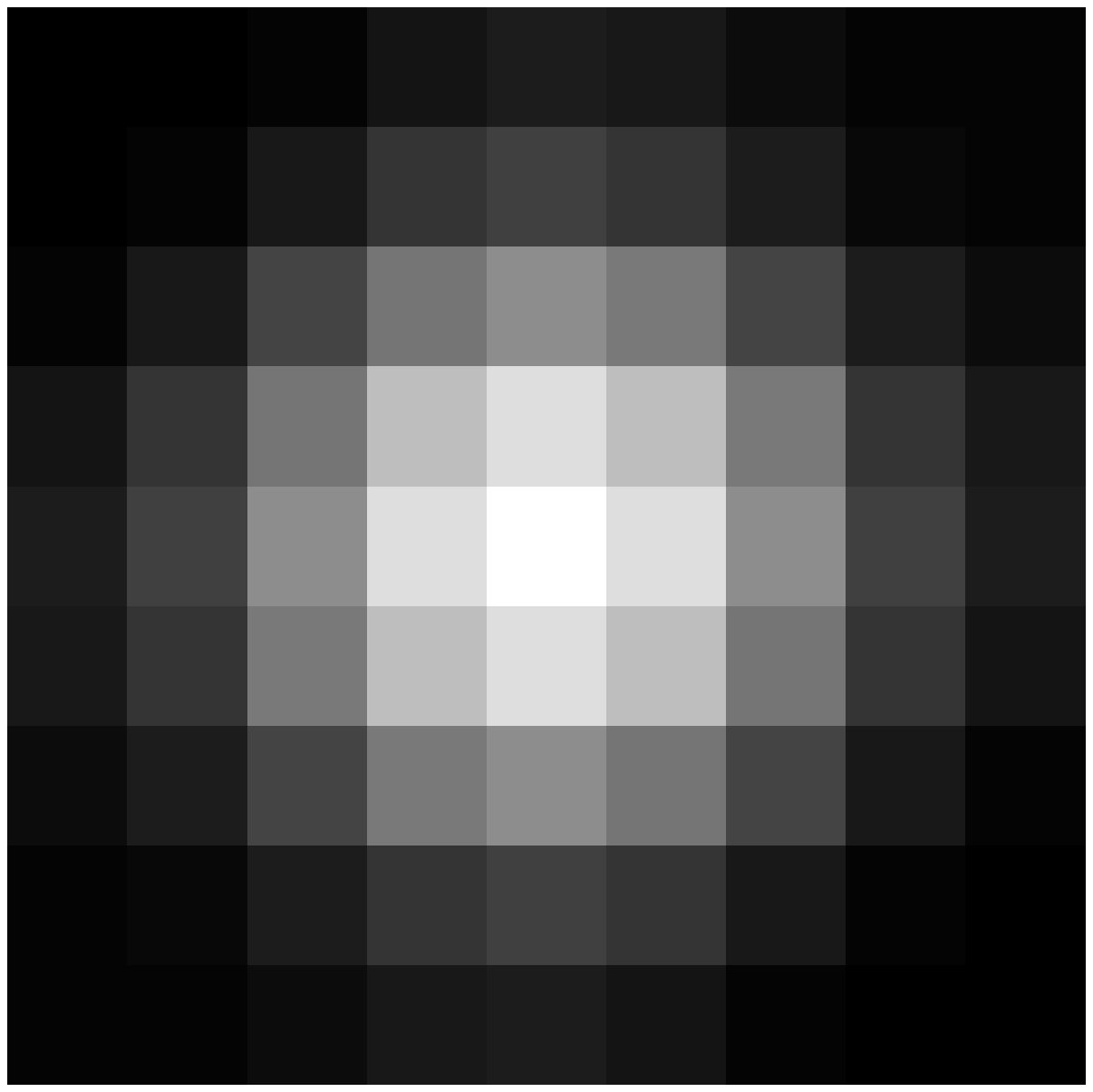}}
\end{overpic}\\

\textbf{SelfDeblur} & \textbf{DBSRCNN} & \textbf{DeblurGAN} & \textbf{proposed (greedy)} & \textbf{proposed (end-to-end)} \\
MSE = 6.2415&&&MSE = \textbf{1.2555}$\mathbf{\times10^{-4}}$&MSE =\textbf{1.1817}$\mathbf{\times10^{-4}}$ \\
PieAPP = 4.1130&PieAPP = 1.8130&PieAPP = 1.9762&PieAPP = 1.2950&PieAPP = 1.2088\\

\includegraphics[height=3cm, trim = {1cm 1cm 1cm 1cm}, clip]{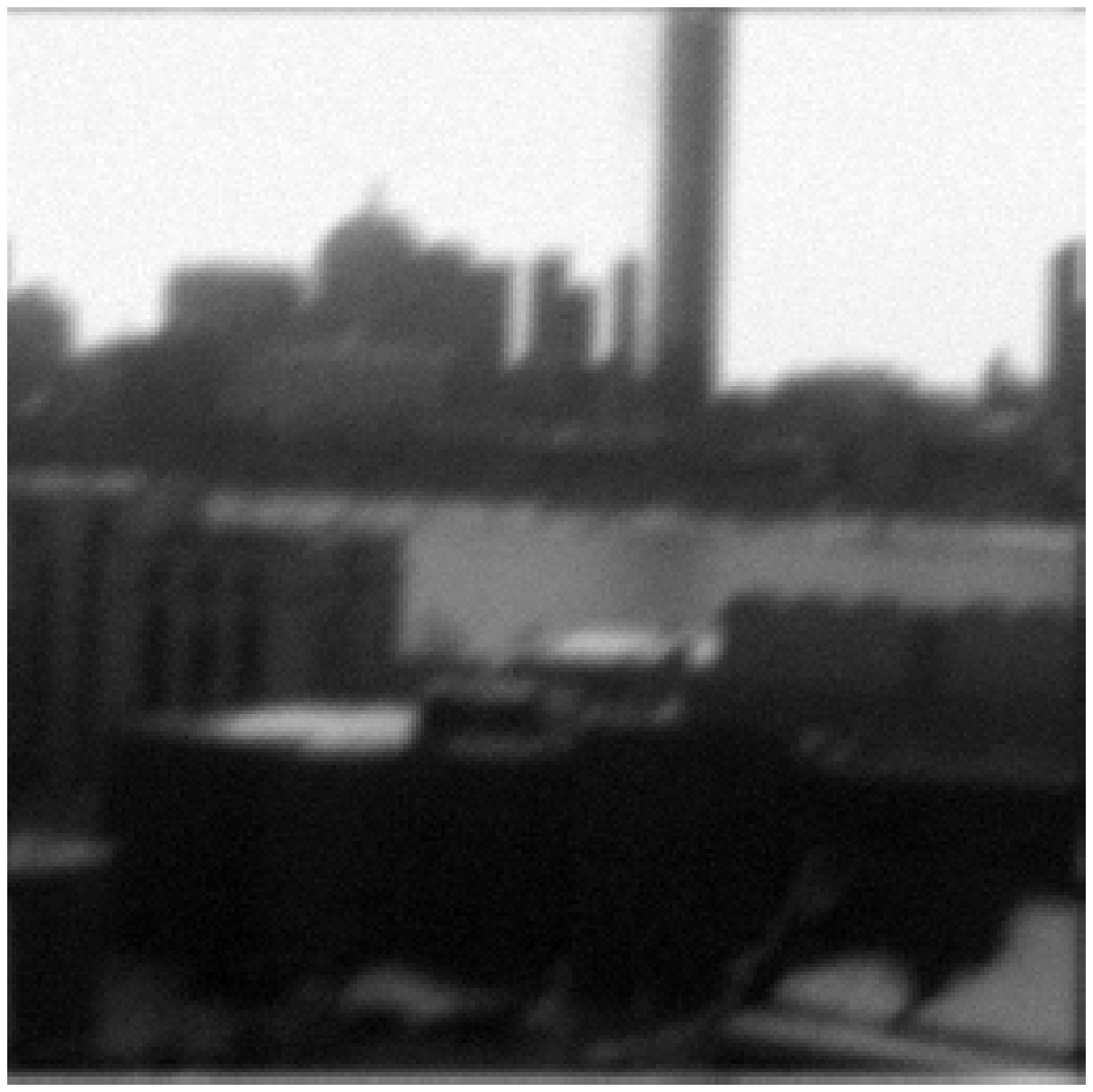} &\hspace{-0.7cm}

\begin{overpic}
[height=3cm, trim = {1cm 1cm 1cm 1cm}, clip]{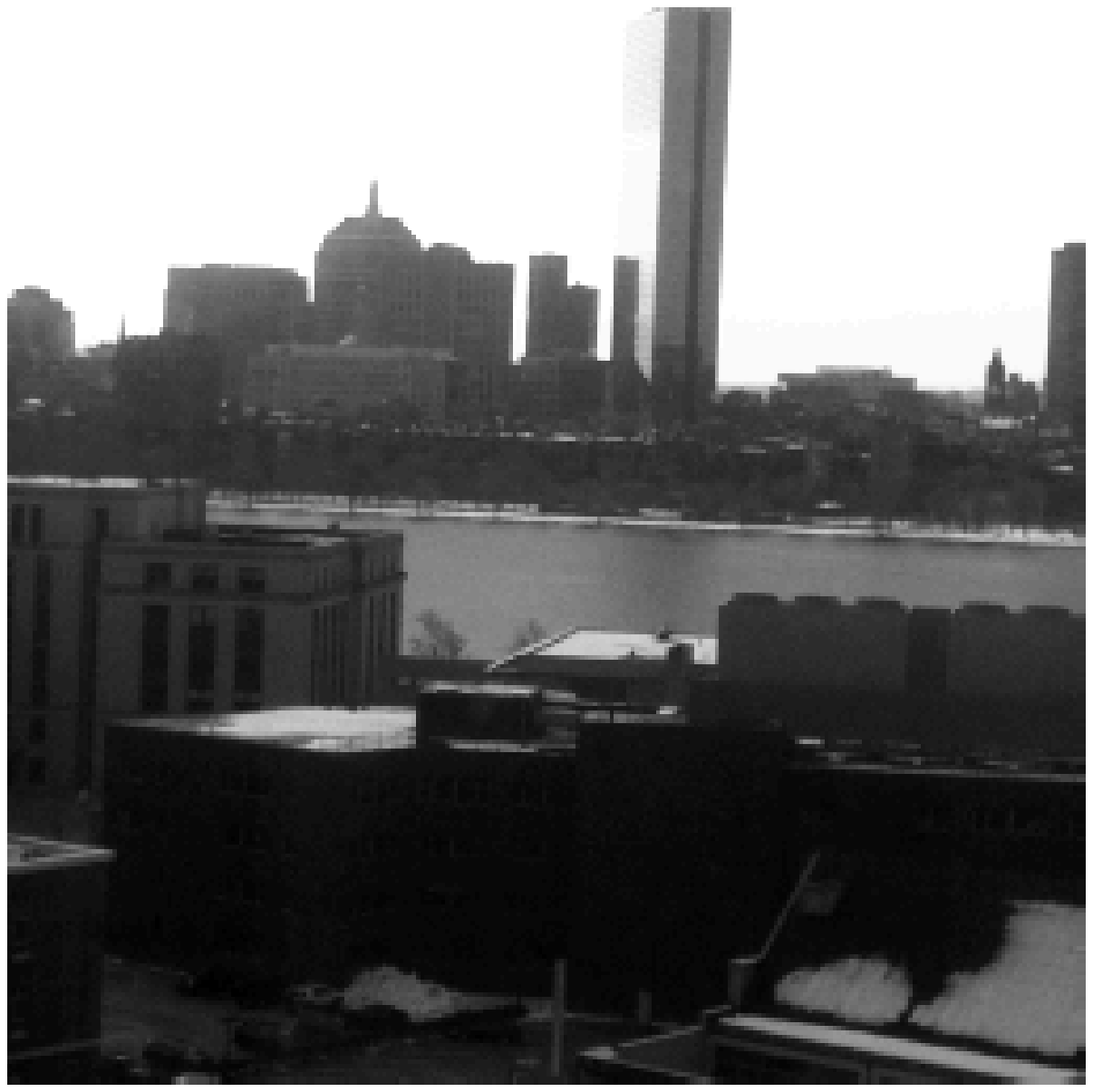}
\put(22,6)
{\includegraphics[height=1cm,trim={4cm 1.5cm 3.4cm 1cm},clip]{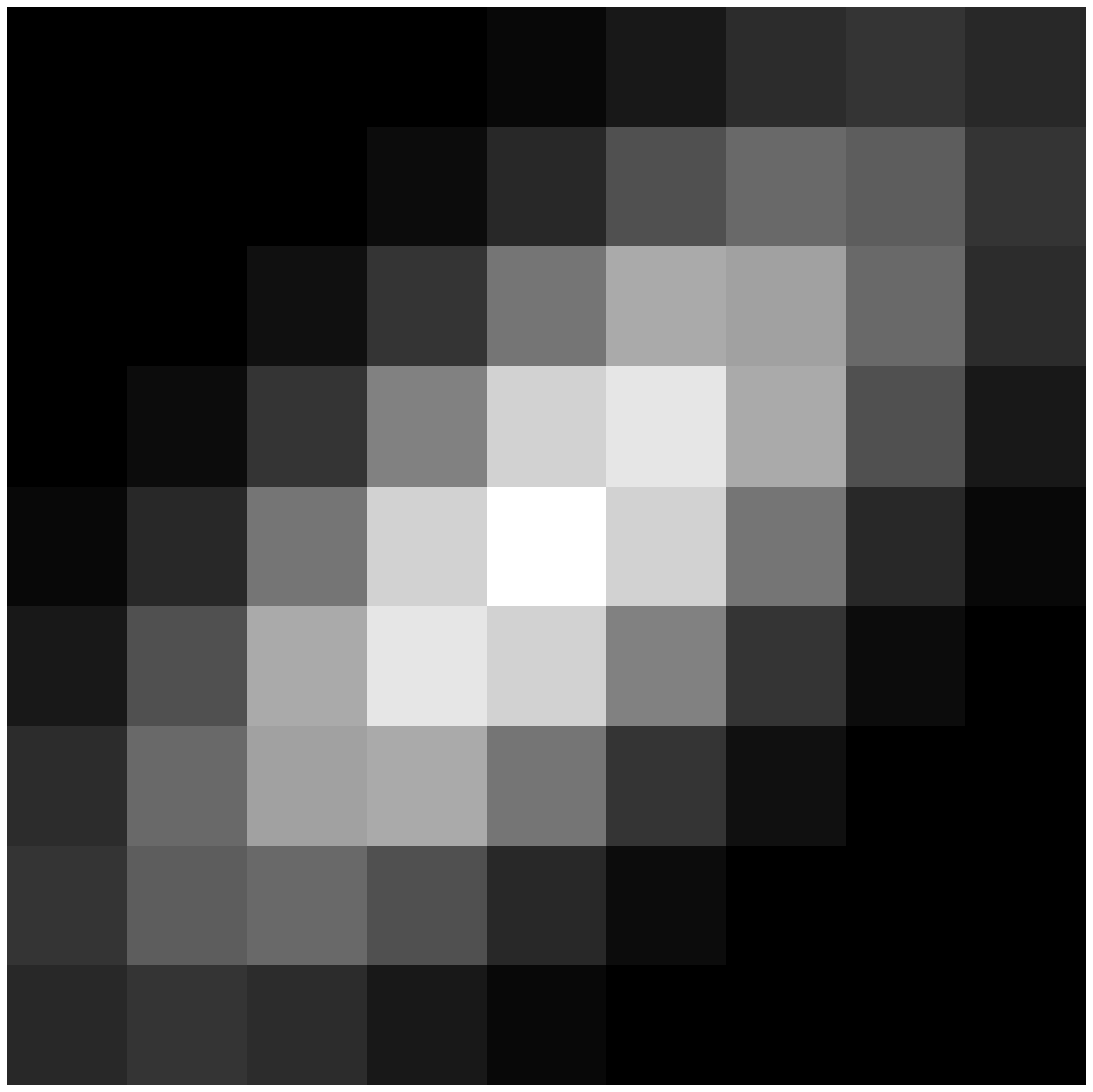}}
\end{overpic}&\hspace{-0.7cm}

\begin{overpic}
[height=3cm, trim = {1cm 1cm 1cm 1cm}, clip]{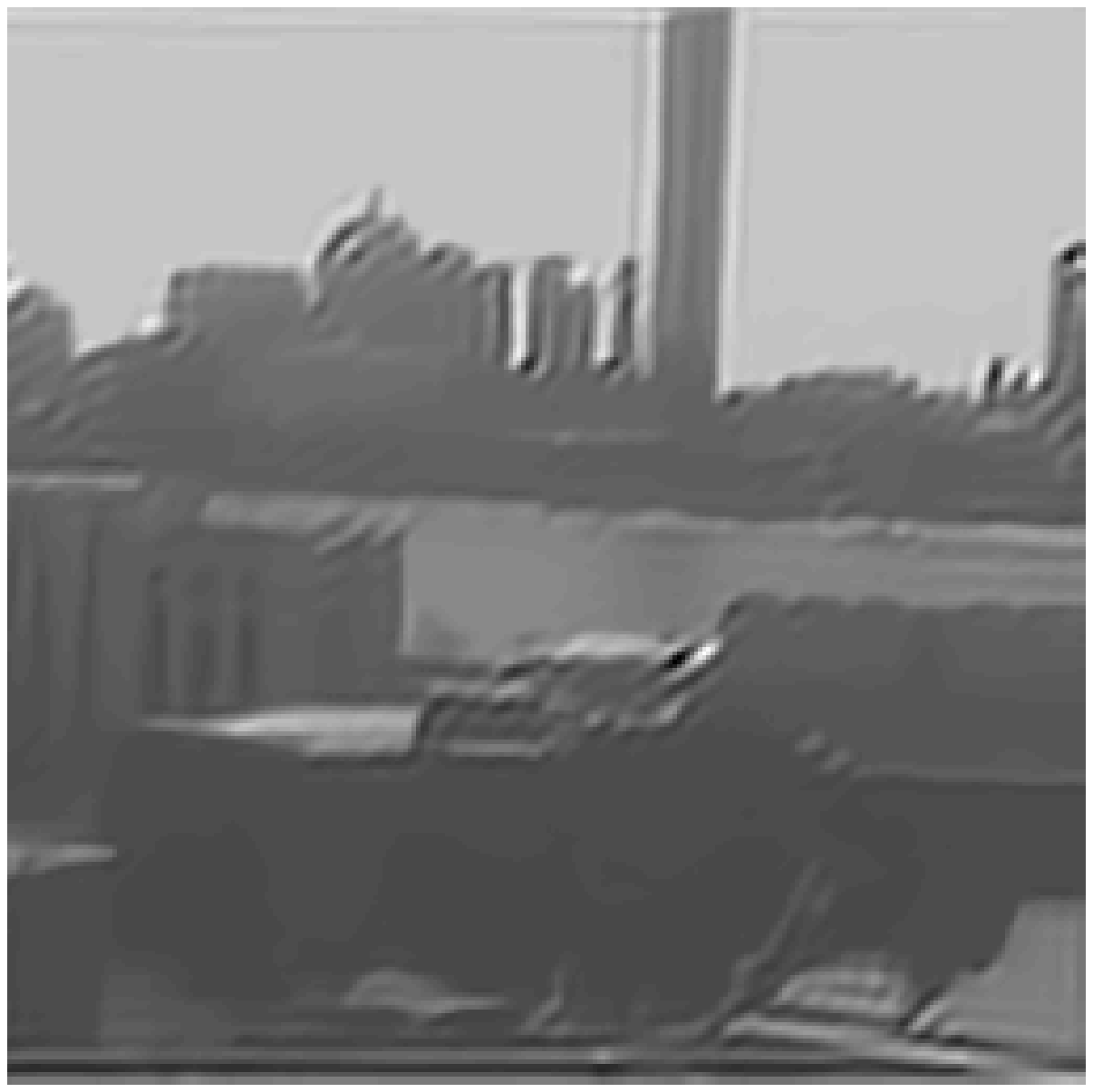}
\put(22,6)
{\includegraphics[height=1cm,trim={4cm 1.5cm 3.4cm 1cm},clip]{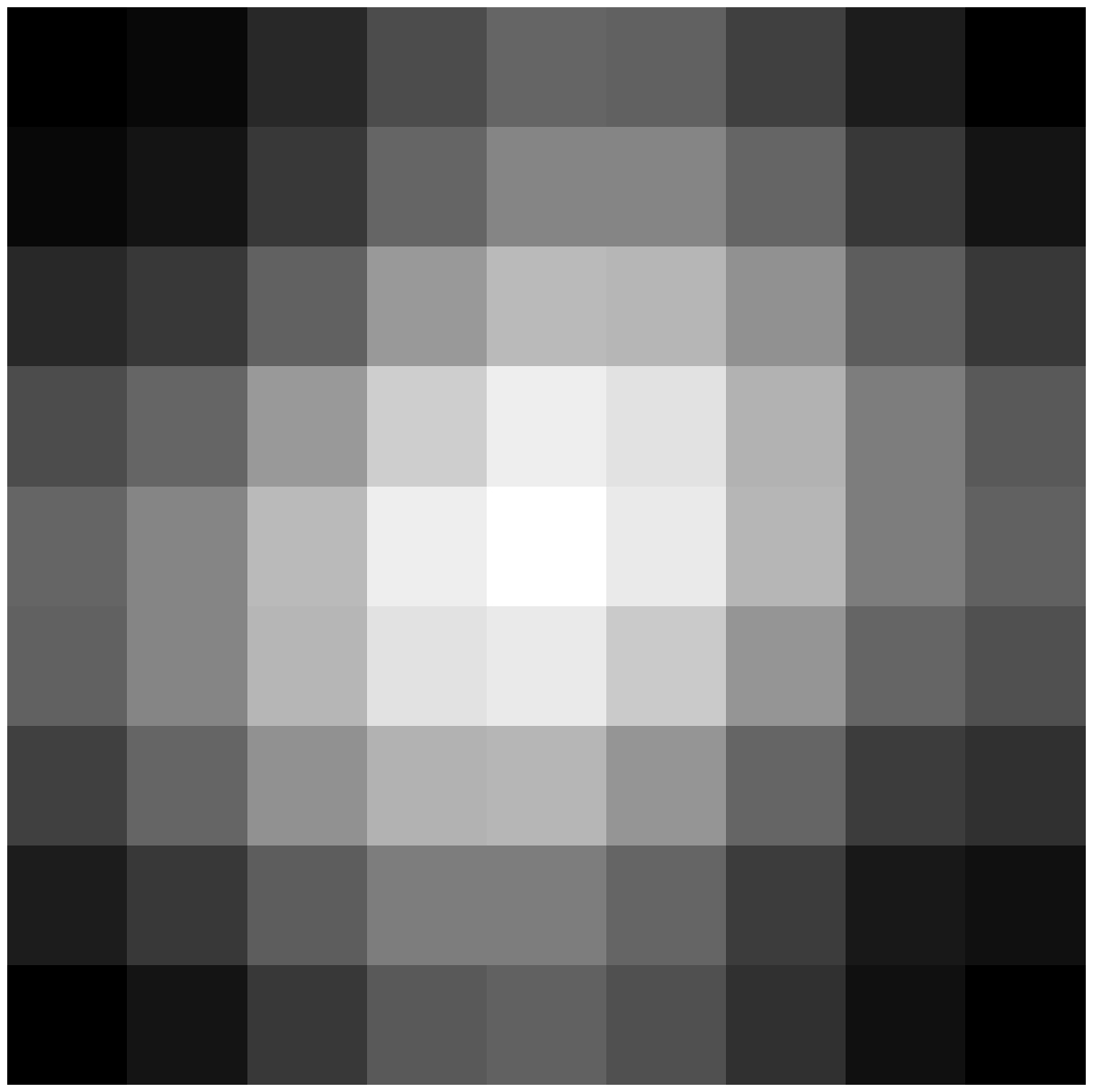}}
\end{overpic}&\hspace{-0.7cm}

\begin{overpic}
[height=3cm, trim = {1cm 1cm 1cm 1cm}, clip]{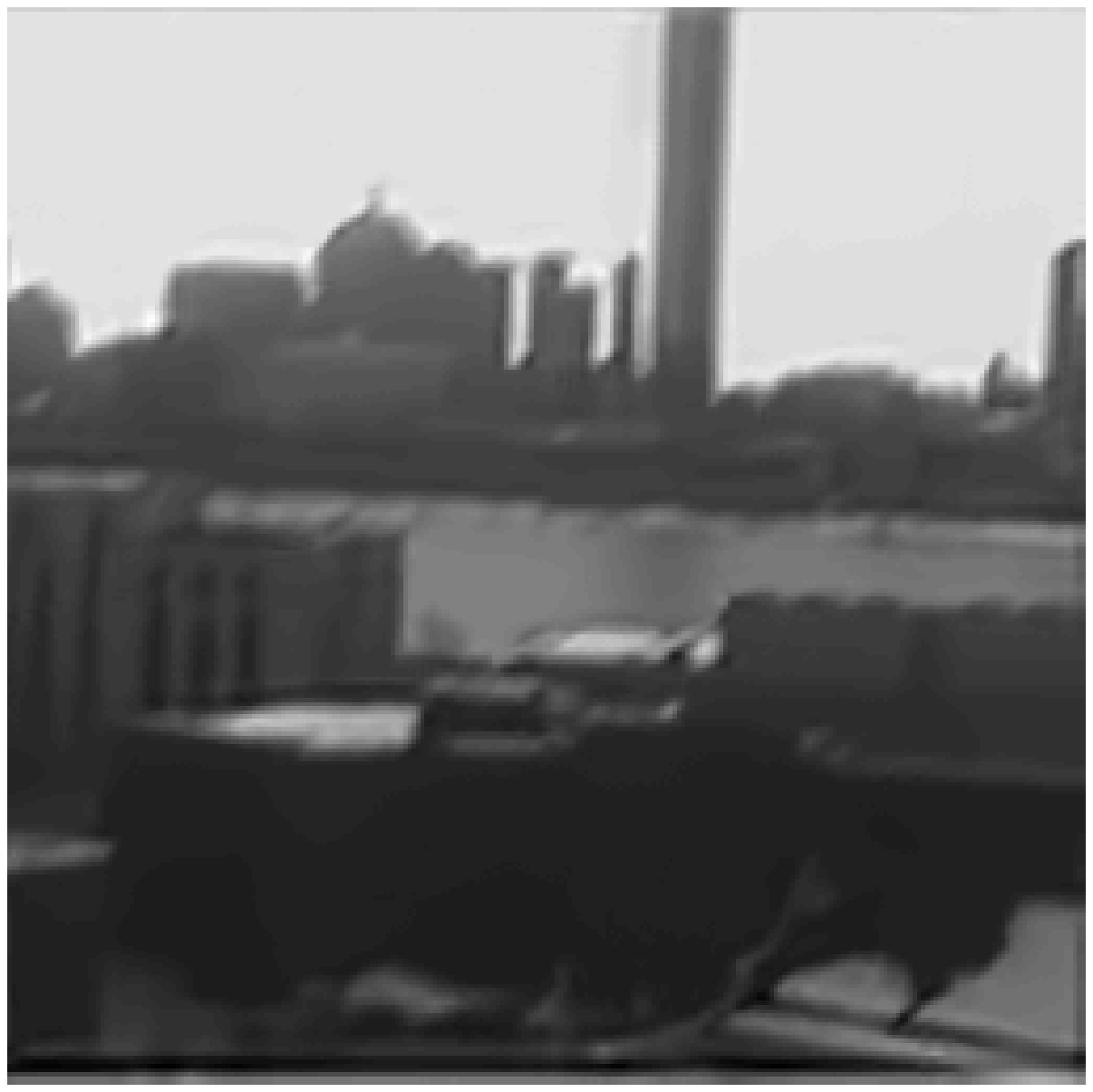}
\put(22,6)
{\includegraphics[height=1cm,trim={4cm 1.5cm 3.4cm 1cm},clip]{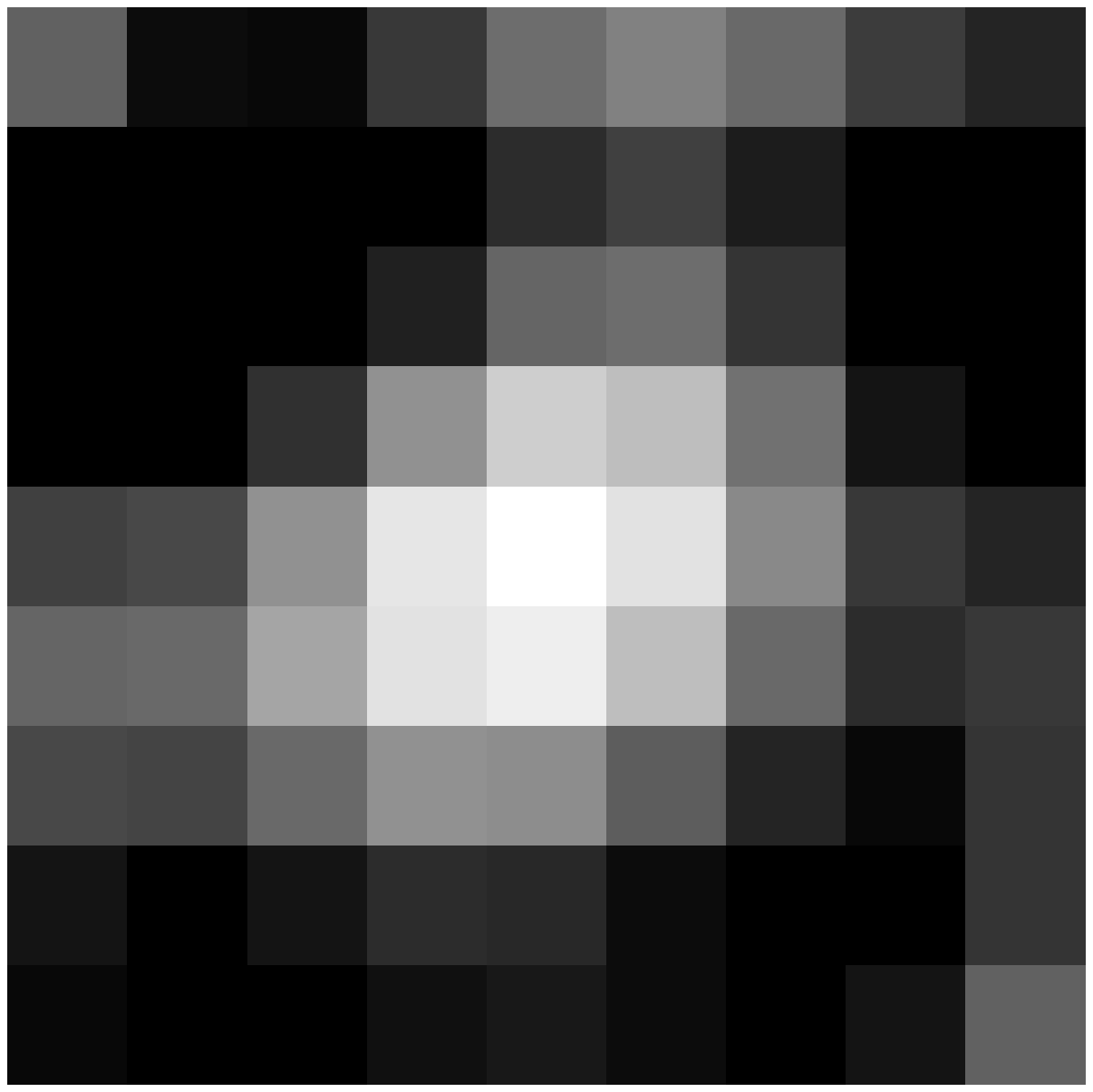}}
\end{overpic}&\hspace{-0.7cm}

\begin{overpic}
[height=3cm, trim = {1cm 1cm 1cm 1cm}, clip]{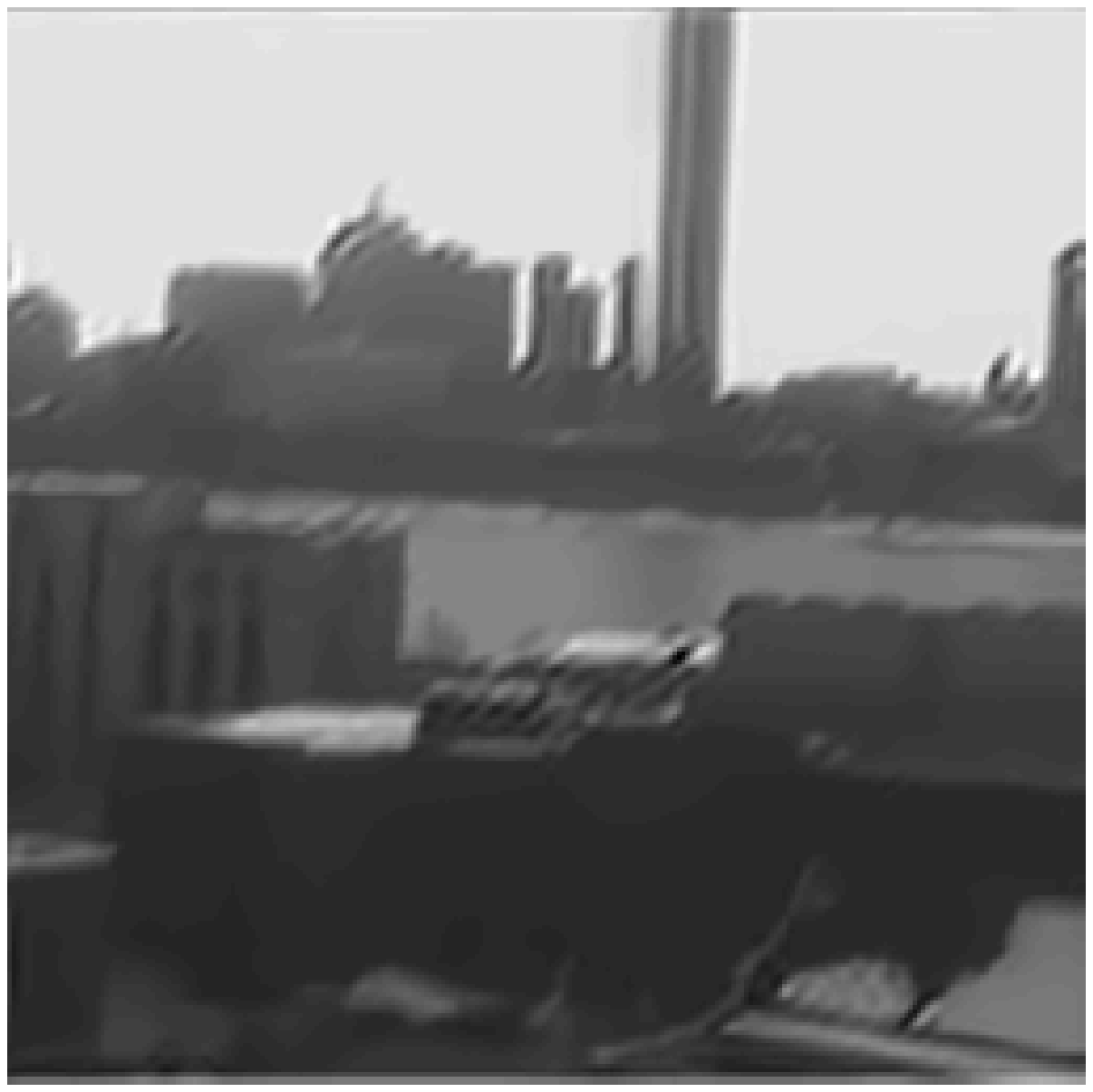}
\put(22,6)
{\includegraphics[height=1cm,trim={4cm 1.5cm 3.4cm 1cm},clip]{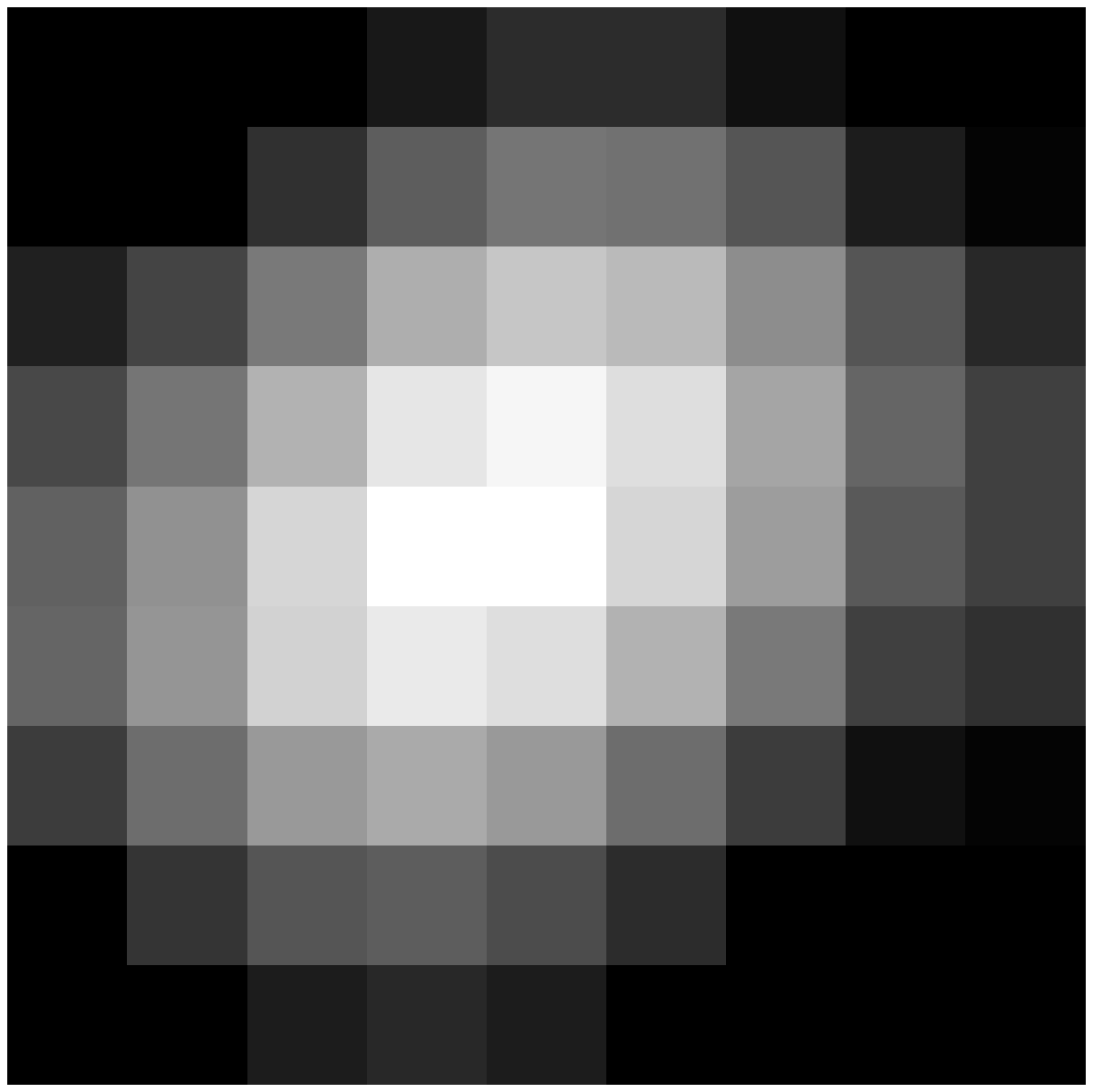}}
\end{overpic}\\

\textbf{Degraded} & \textbf{Original} & \textbf{VBA} & \textbf{deconv2D} & \textbf{blinddeconv}\\
& & MSE = 0.0054&MSE = 0.0058&MSE = 0.0040\\  
PieAPP = 3.4024&&PieAPP = 1.6356&PieAPP = 1.9079&PieAPP = 1.8397\\

\begin{overpic}
[height=3cm, trim = {1cm 1cm 1cm 1cm}, clip]{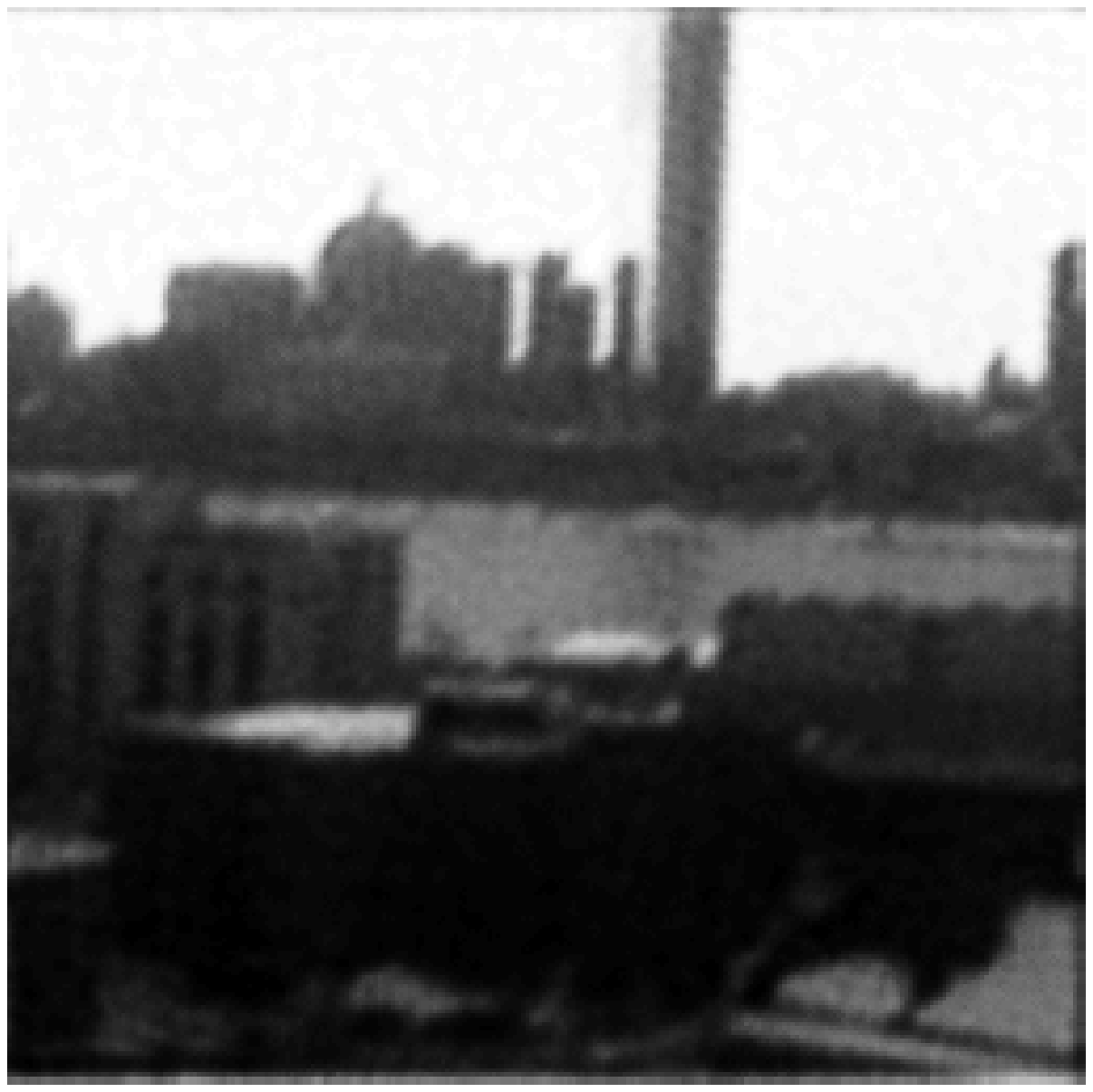}
\put(22,6)
{\includegraphics[height=1cm,trim={4cm 1.5cm 3.4cm 1cm},clip]{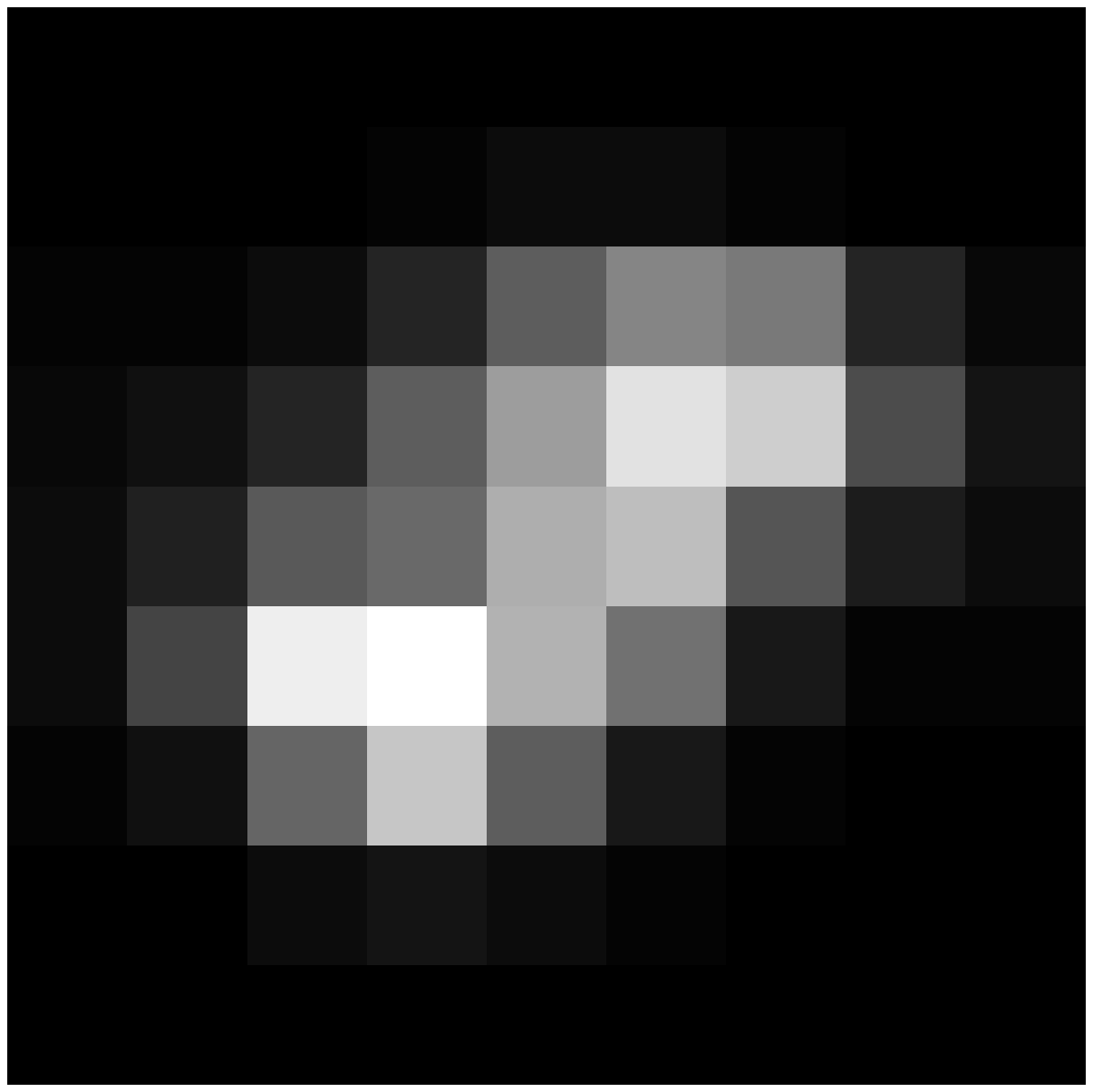}}
\end{overpic}&\hspace{-0.7cm}

\includegraphics[height=3cm, trim = {1cm 1cm 1cm 1cm}, clip]{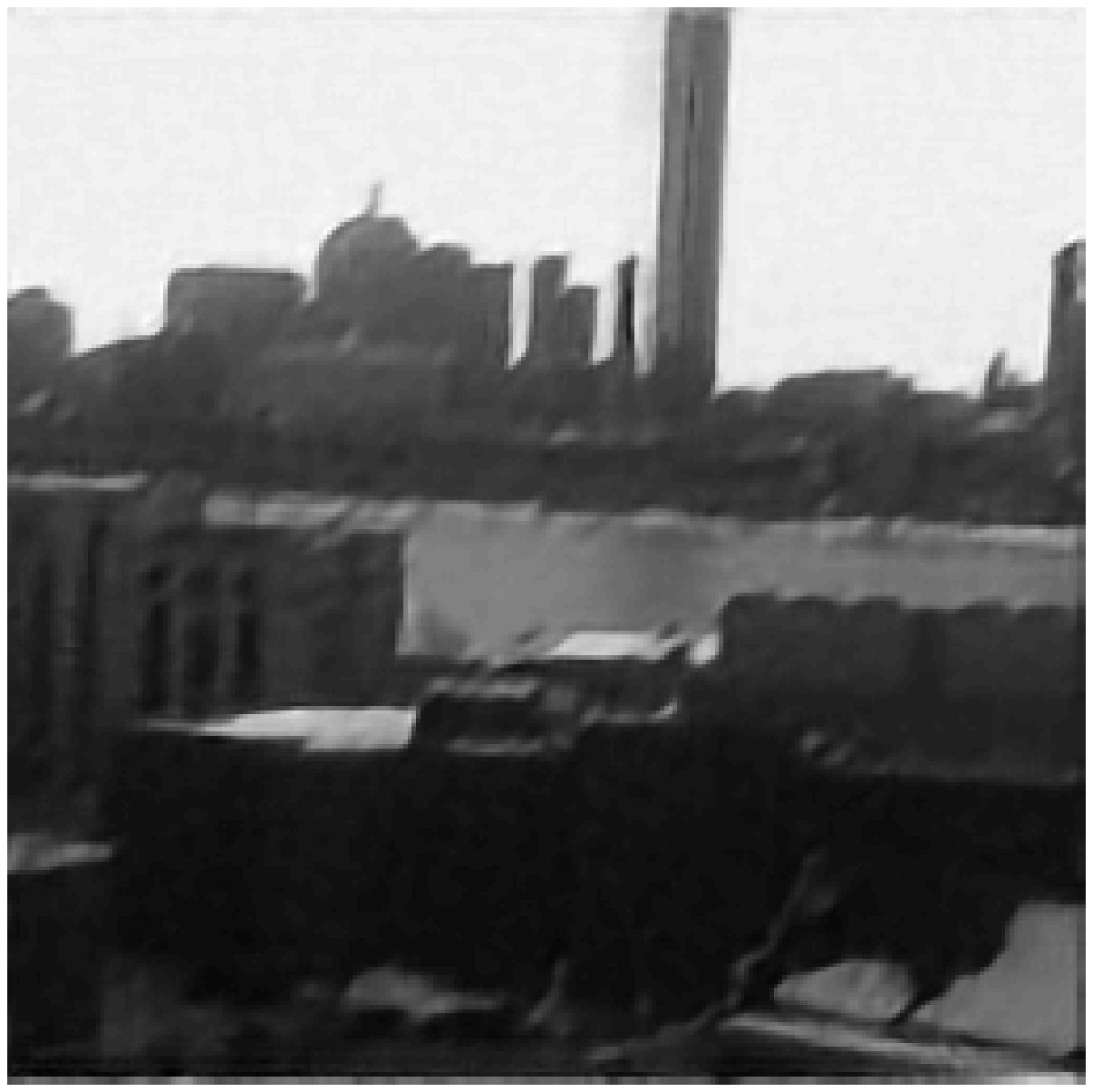}
&\hspace{-0.7cm}

\includegraphics[height=3cm, trim = {1cm 1cm 1cm 1cm}, clip]{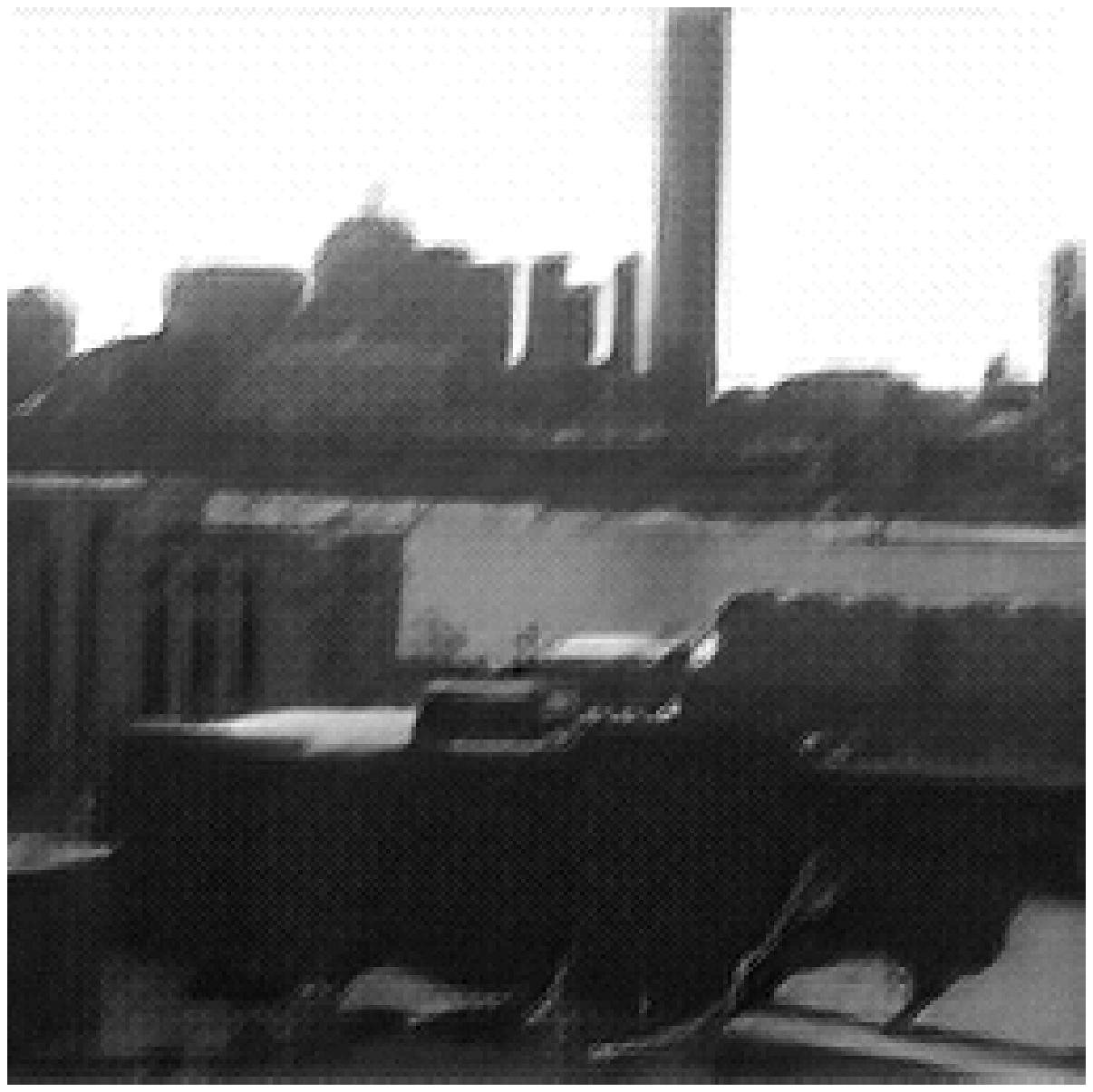}
&\hspace{-0.7cm}

\begin{overpic}
[height=3cm, trim = {1cm 1cm 1cm 1cm}, clip]{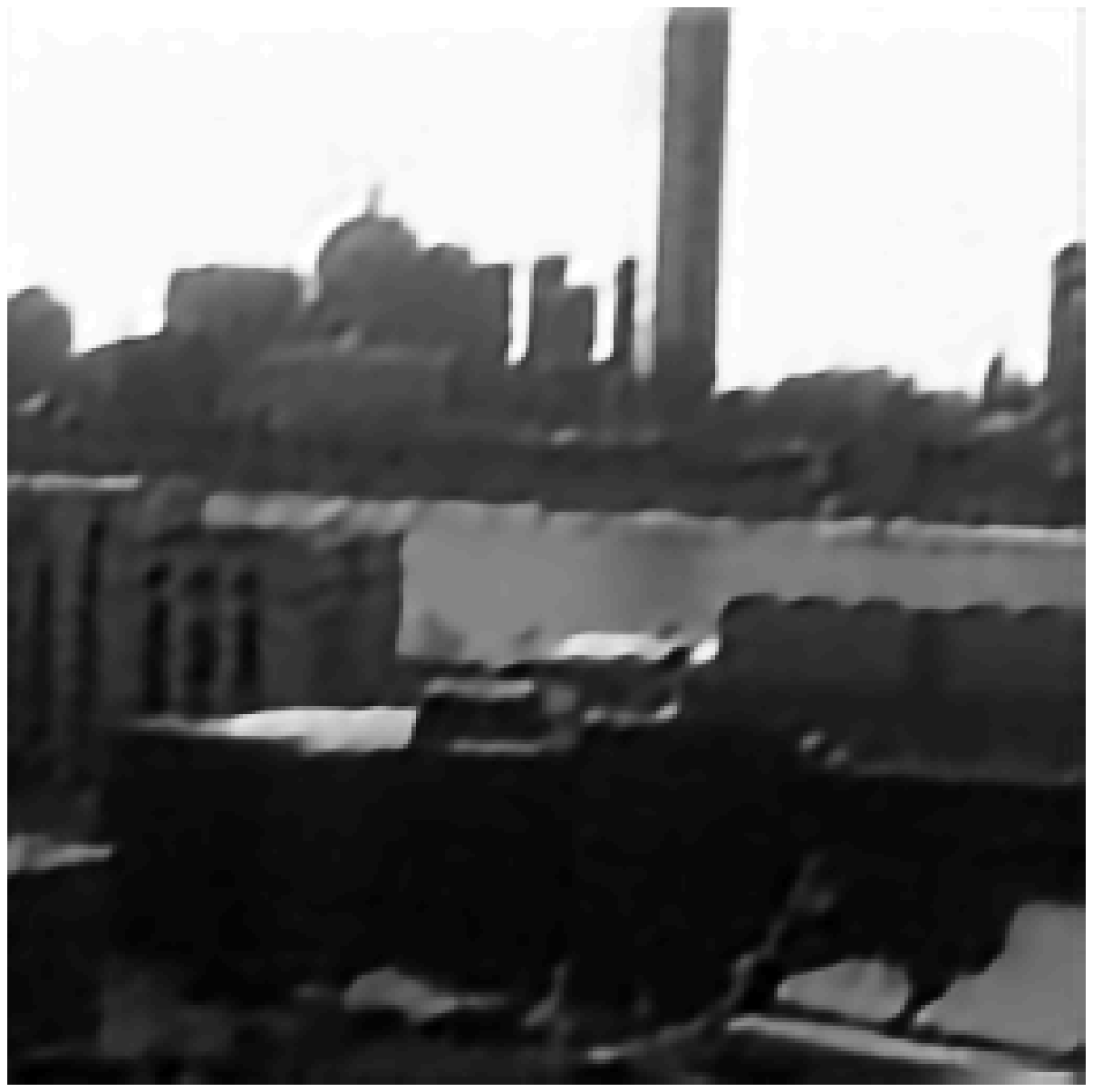}
\put(22,6)
{\includegraphics[height=1cm,trim={4cm 1.5cm 3.4cm 1cm},clip]{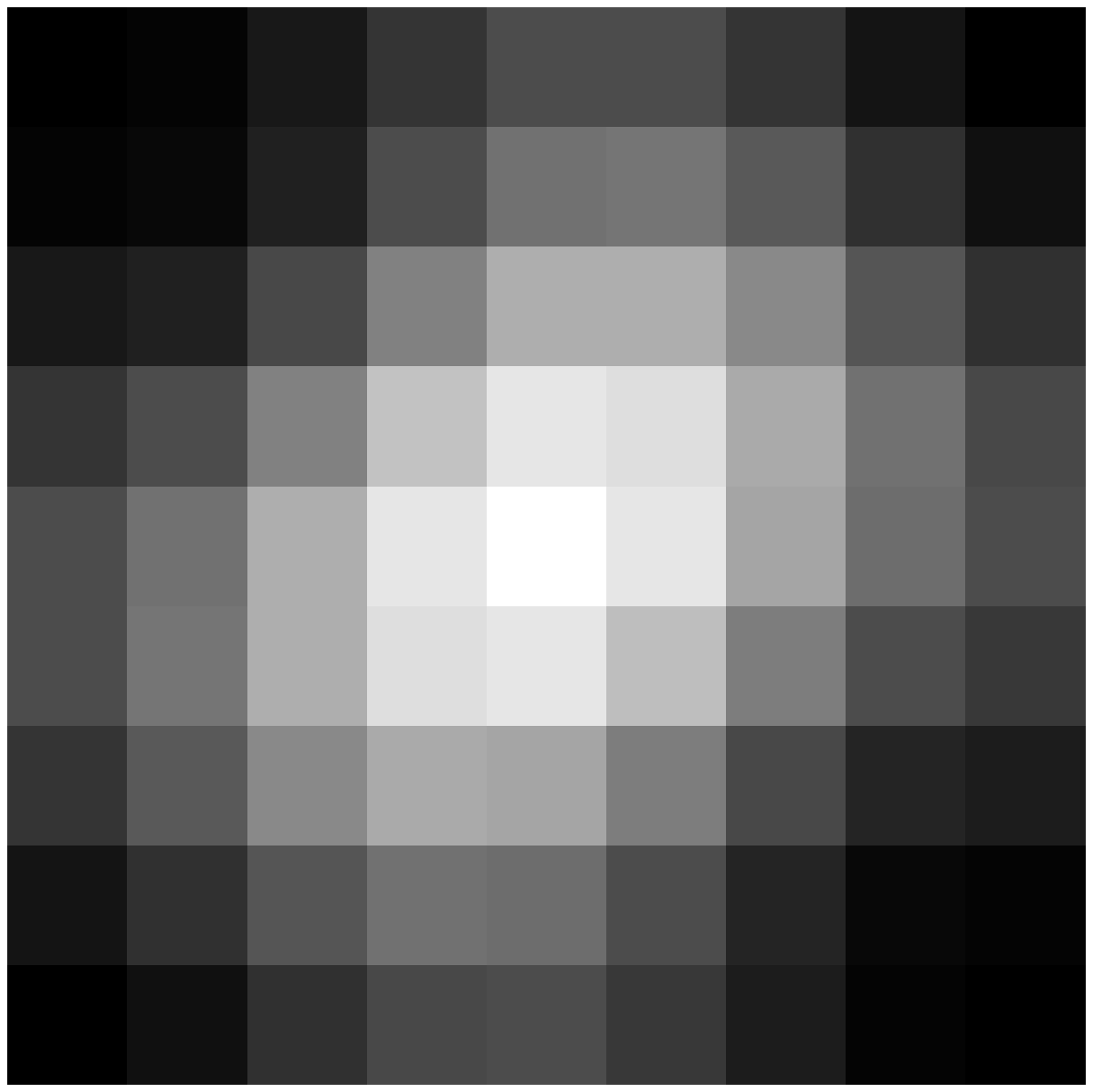}}
\end{overpic}&\hspace{-0.7cm}

\begin{overpic}
[height=3cm, trim = {1cm 1cm 1cm 1cm}, clip]{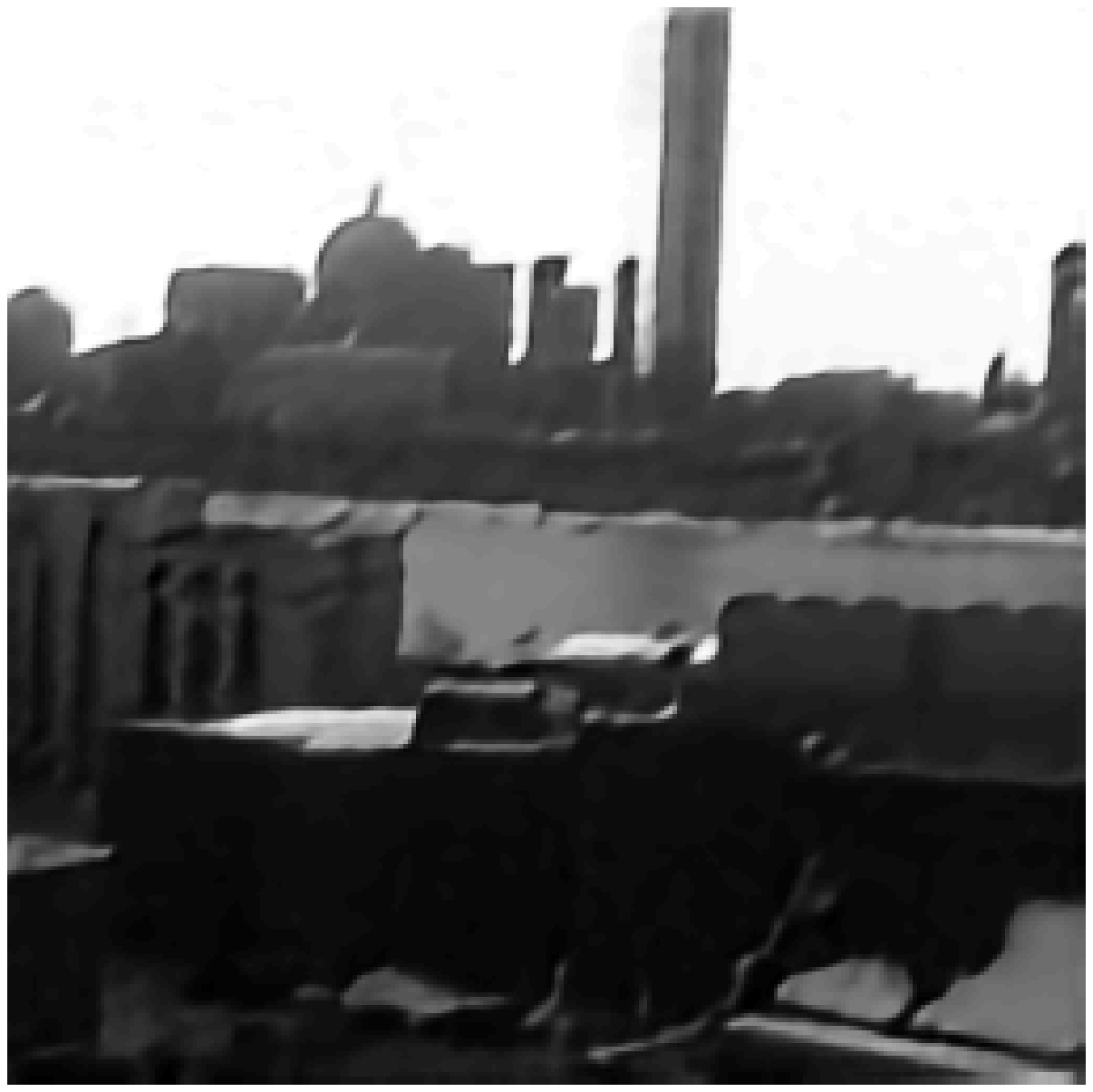}
\put(22,6)
{\includegraphics[height=1cm,trim={4cm 1.5cm 3.4cm 1cm},clip]{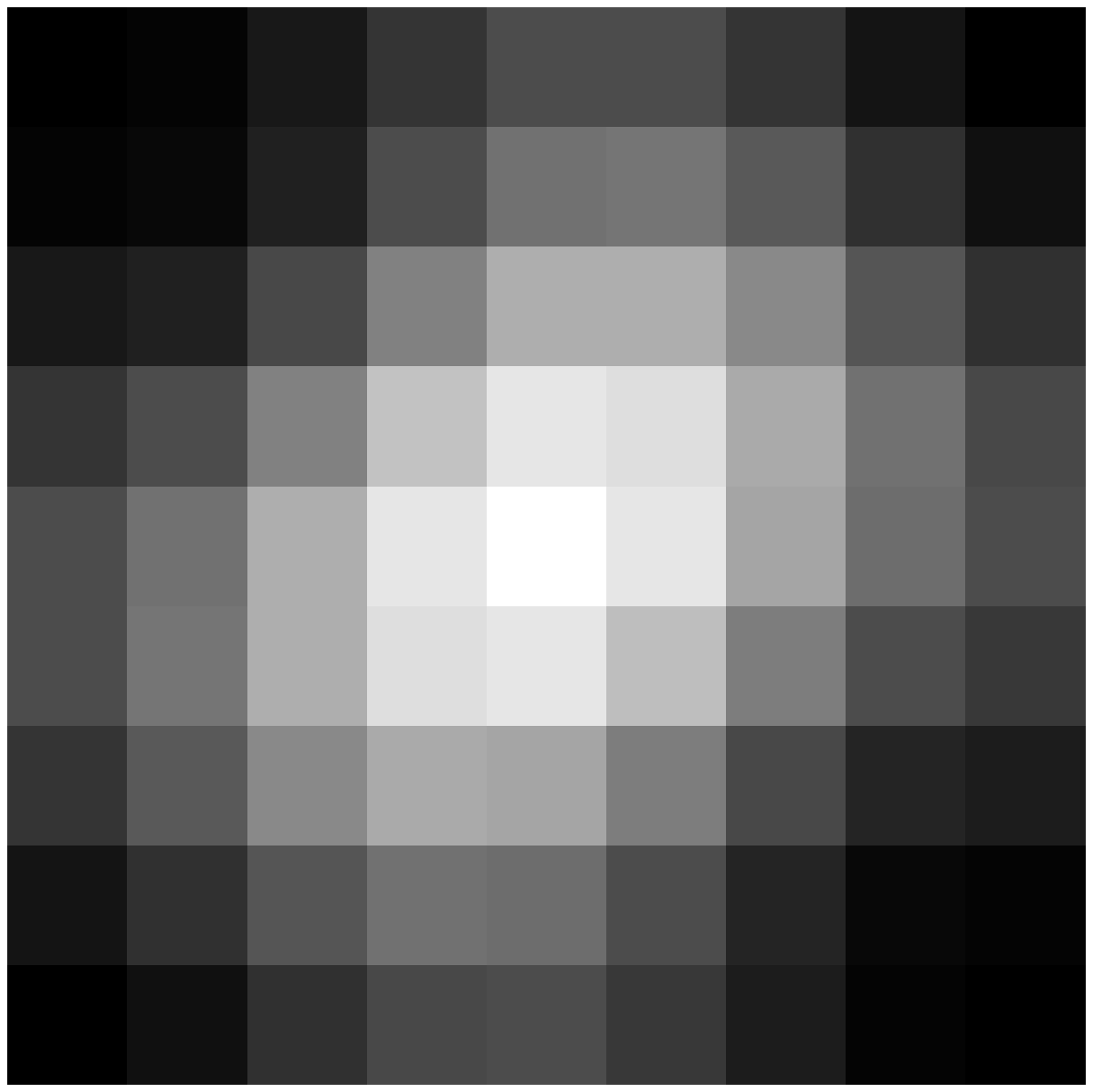}}
\end{overpic}\\

\textbf{SelfDeblur} & \textbf{DBSRCNN} & \textbf{DeblurGAN} & \textbf{proposed (greedy)} & \textbf{proposed (end-to-end)} \\
MSE = 7.0703&&&MSE = 0.0035&MSE = \textbf{0.0034}  \\
PieAPP = 2.5108&PieAPP = 1.4002 &PieAPP = 1.5206&PieAPP = 1.3922 &PieAPP =\textbf{1.2468} \\

\end{tabular}
\caption{\footnotesize Ground-truth image/blur, degraded image, restored images (with PieAPP index) and estimated blurs (with MSE score) when available, for various methods, on two examples in the test set of \emph{Dataset 1}.}
\label{fig:Dataset1_case3}
\end{figure*}

{In Tables~\ref{table:result_blur1} and \ref{table:result_image1}, we report the results of kernel estimation and image restoration, computed on the test set, using the different methods}. As could be expected, the greedy approach tends to give more weight to the kernel quality than the end-to-end training. Our two training approaches yield great performance, when compared to all the other tested approaches. One can notice that the VBA with finetuned value for $\xi$ performs quite well, showing the validity of our Bayesian formulation. The proposed unrolled VBA technique allows us to avoid a manual tuning of this parameter, and further increases the resulting quality. This is a direct outcome of the supervised training procedure aiming at maximizing quality scores, and also to the introduction of a post-processing step on the images. DBSRCNN has a good performance in terms of image restoration in this dataset. However, it is not capable of estimating the blur kernel, which might be useful for various applications. We display two examples of results in Fig~\ref{fig:Dataset1_case3}, extracted from our test set. One can notice, by visual inspection of these results, the high quality of the restored images. No artifacts can be observed, which is confirmed by a low average value of the PieAPP index on the test set. Moreover, the kernels are generally estimated quite accurately, as shown by the low MSE score and the good retrieval of their general structure. In the few cases when the unfolded VBA algorithm fails to give a perfect recovery of the blur kernel as in Fig.~\ref{fig:Dataset1_case3}(bottom), the estimation is still accurate enough to yield a good recovery of the image whatever \emph{greedy training} or \emph{end-to-end training} is used. One can also notice that our method tends to provide better contrasted images, compared to its closest competitor in the image restoration task that is DBSRCNN. We display in Fig.~\ref{fig:SSIM_loss}(left) the evolution of the SSIM loss during the end-to-end training of the proposed architecture, showing the increase  of the loss, then its stabilization, for both training and validation set, thus confirming an appropriate setting of ADAM optimizer parameters. Finally, Table~\ref{table:time1}(left) displays the average test time for each methods, that is the computational time required to restore one example of the dataset, once the method is finetuned/trained. {We displayed CPU time for a fair comparison between methods, for codes ran on a Dell workstation equipped with an Xeon(R) W-2135 processor (3.7 GHz clock frequency and 12 GB of RAM).} GPU time is also indicated when available. The fastest method is DBSRCNN, though we must emphasize that, in contrast with all the other methods based on Matlab/Python softwares, DBSRCNN makes use of an optimized C implementation, for its test phase on CPU. DeblurGAN is also very fast, but the resulting quality was quite poor in our experiments. Our method reaches a reasonable computational time on CPU. It becomes quite competitive when making use of GPU implementation, as the unrolled architecture is well suited for that purpose. This allows to drop the test time per image to few seconds, making it advantageous, with the addition benefit of better quality results in average, and of an available kernel estimate.

\begin{table}
\scriptsize
\renewcommand{\arraystretch}{1.2}
\centering
\begin{tabular}{|c||c|c|c|c|}
\hline
Method & MSE & $\mathcal{H}_{\infty}$ error & MAE\\
\hline
\hline
VBA &0.0017 (0.0022)&0.1674 (0.1003)&0.0472 (0.0317)\\
\hline
deconv2D &0.0025 (0.0035)&0.1483 (0.1037)&0.0489 (0.0395)\\
\hline
blinddeconv &0.0013 (0.0011) &0.1553 (0.0660) &0.0417 (0.0203) \\
\hline
SelfDeblur &8.9165 (4.0668) &15.0213 (6.8490)& 3.5314 (0.8998)\\
\hline
Proposed (greedy) &\textbf{0.0008} (0.0012)    &  \textbf{0.1165} (0.0677)   & \textbf{0.0281} (0.0168)  \\
\hline
Proposed (end-to-end) &0.0009 (0.0013)  & 0.1188 (0.0672)  & 0.0289 (0.0170)  \\
\hline
\end{tabular}
\vspace{0.1cm}
\caption{\footnotesize Quantitative assessment of the restored kernels. Mean (standard deviation) values computed over the test sets of \emph{Dataset 1}.}
\label{table:result_blur1}
\end{table}

%

\begin{table}
\scriptsize
\renewcommand{\arraystretch}{1.2}
\centering
\begin{tabular}{|c||c|c|c|c|}
\hline
Method & SSIM & PSNR & PieAPP\\
\hline
\hline
Blurred & 0.6542 (0.1072)  &  22.2254 (2.3779)  & 4.1794 (0.9005)  \\
\hline
VBA & 0.7603 (0.0814)&23.7332 (2.5672) & 1.5109 (0.6184) \\
\hline
deconv2D &0.7668 (0.0912) &24.5970 (2.8656)&1.9289 (0.4959)\\
\hline
blinddeconv&0.7528 (0.0963)&23.9347 (2.4299)&1.9170 (0.6630)\\
\hline
SelfDeblur & 0.6948 (0.1006)&22.2704 (2.1255) &3.3178 (0.7291) \\
\hline
DBSRCNN &0.7780 (0.0895)&\textbf{24.9561} (2.9800)&1.5959 (0.6463)\\
\hline
DeblurGAN &0.6613 (0.0731)&22.4388 (2.4074)&1.8937 (0.7630) \\
\hline
Proposed (greedy) & 0.7945 (0.0890)   &  \textbf{24.7093} (2.9351) & 1.4047 (0.6437) \\
\hline
Proposed (end-to-end) & \textbf{0.7989} (0.0886)  & \textbf{24.6638} (3.0711)  &\textbf{1.1976} (0.5433) \\
\hline
\end{tabular}
\vspace{0.1cm}
\caption{\footnotesize Quantitative assessment of the restored images. Mean (standard deviation) values computed over the test sets of \emph{Dataset 1}.}
\label{table:result_image1}
\end{table}

\begin{table}
\renewcommand{\arraystretch}{1.2}
\centering
\footnotesize
\begin{tabular}{|c||c|c|}
\hline
Method & Dataset 1 & Dataset 2\\
\hline
\hline
VBA & 153s (15s) & 156s (18s)\\ \hline
deconv2D & 16s & 19s \\ \hline
blinddeconv & 19s & 22s \\ \hline
SelfDeblur & 452s (51s) &  455s (55s)\\ \hline
DBSRCNN & 1s & 2s \\ \hline
DeblurGAN & 2s (1s) & 3s (2s)\\ \hline
Proposed & 36s (4s) & 113s (12s)\\
\hline
\end{tabular}
\vspace{0.1cm}
\caption{\footnotesize Average test time per image, using {CPU (resp. GPU)}. }
\label{table:time1}
\end{table}
%

\subsubsection{Dataset 2}
\begin{figure*}[h]
\footnotesize
\begin{tabular}{c@{}c@{}c@{}c@{}c@{}}
  \centering
\includegraphics[height=3cm, trim = {1cm 1cm 1cm 1cm}, clip]{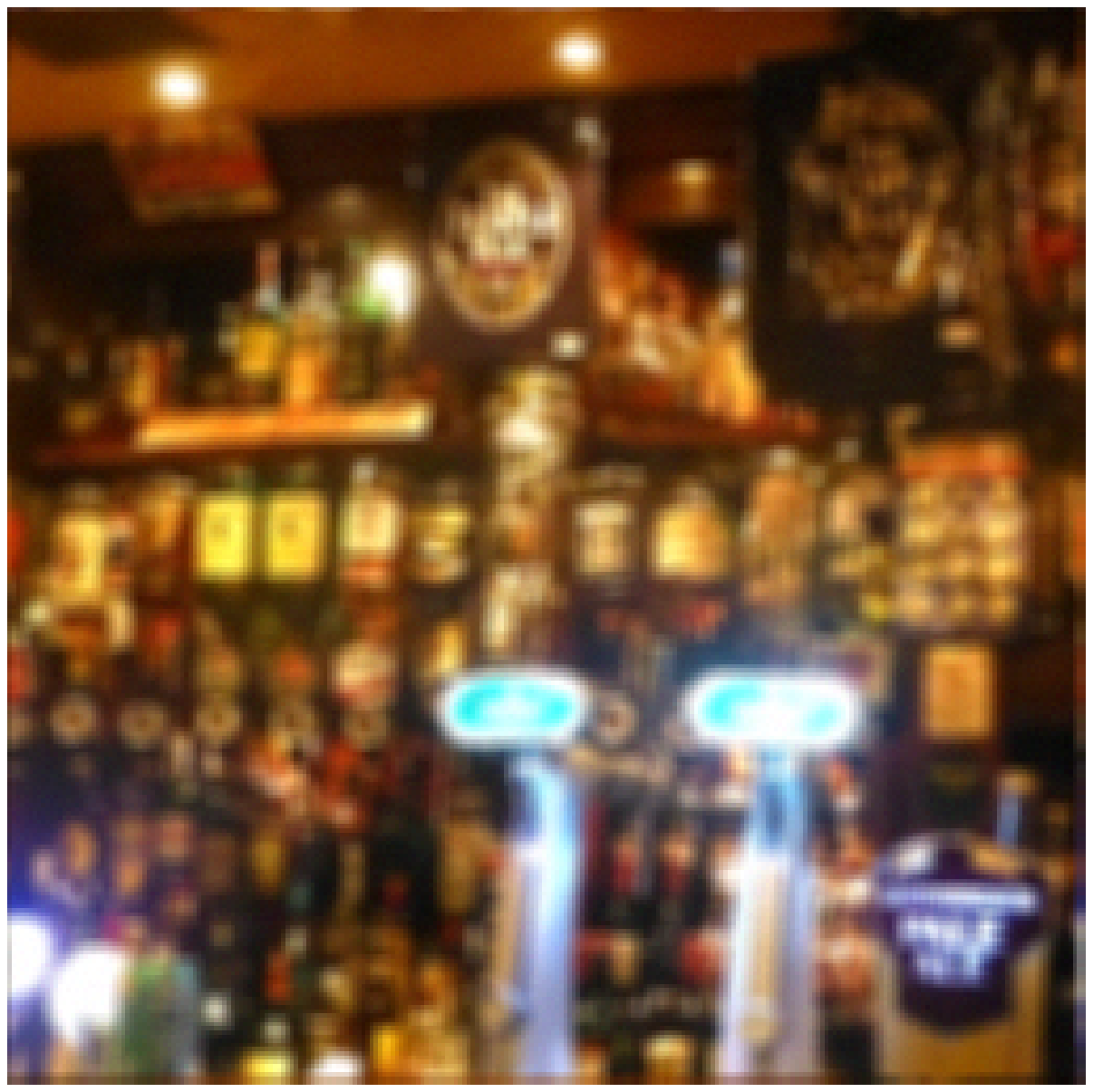} &\hspace{-0.7cm}

\begin{overpic}
[height=3cm, trim = {1cm 1cm 1cm 1cm}, clip]{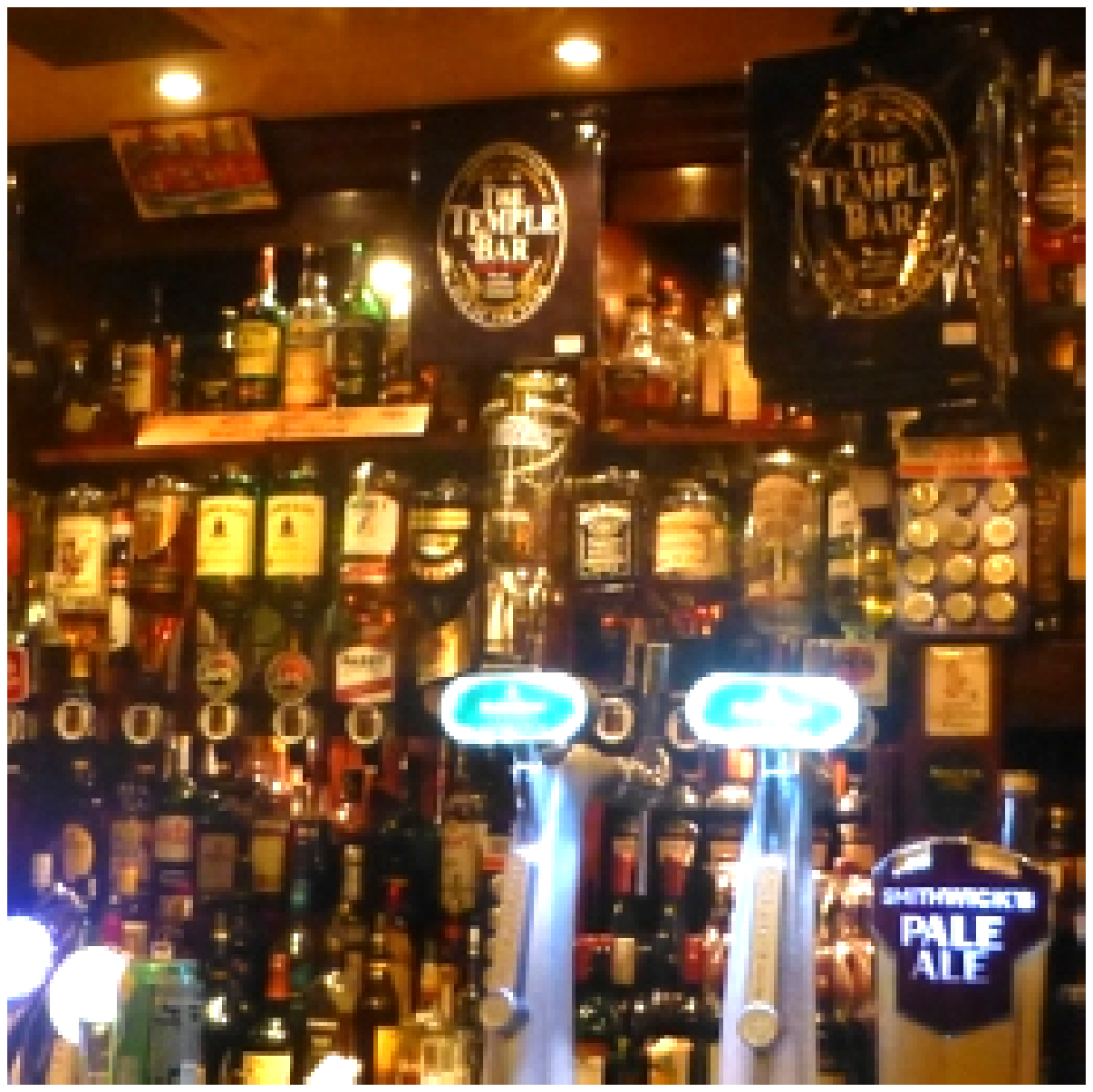}
\put(22,6)
{\includegraphics[height=1cm,trim={4cm 1.5cm 3.4cm 1cm},clip]{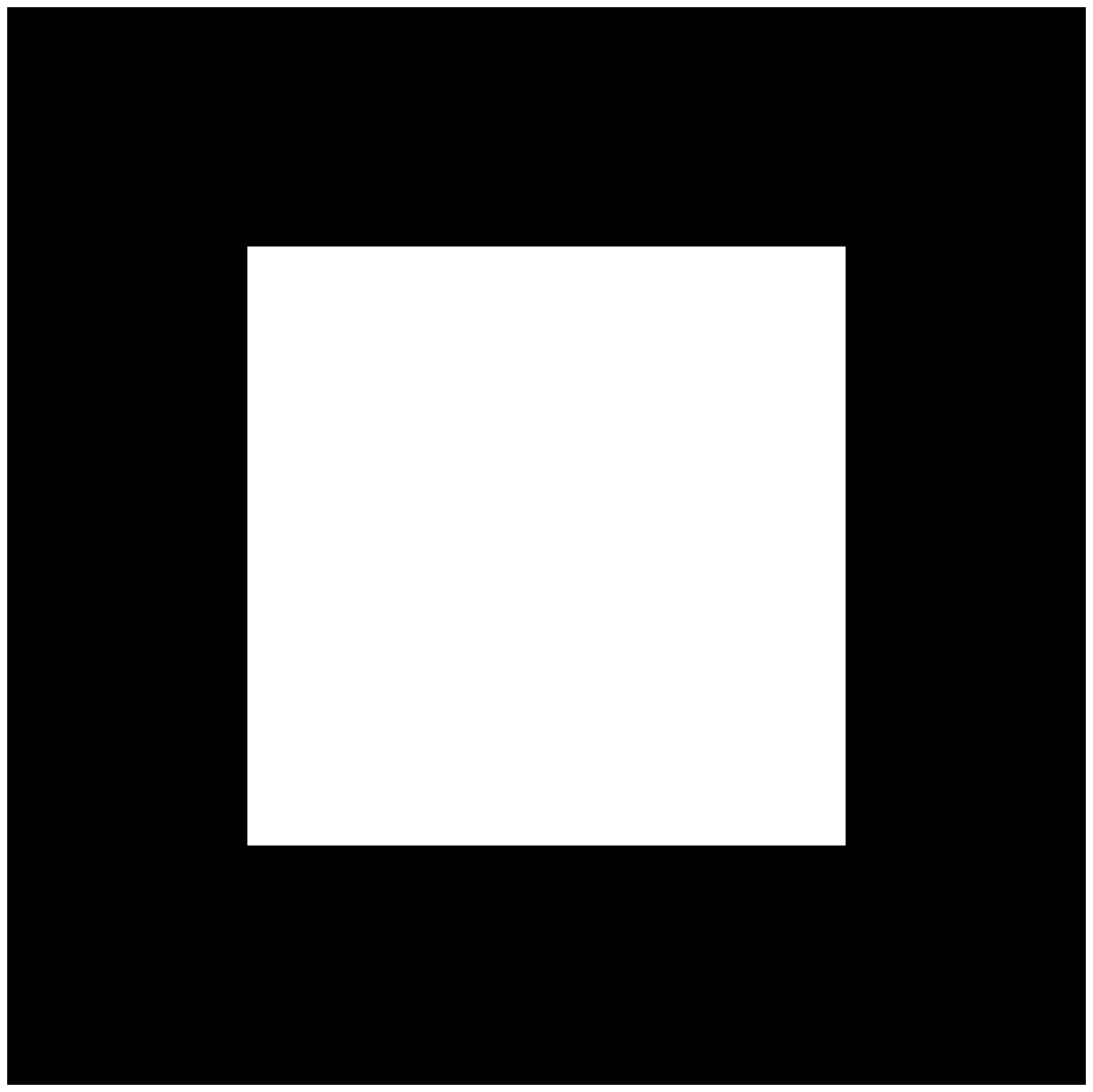}}
\end{overpic}&\hspace{-0.7cm}

\begin{overpic}
[height=3cm, trim = {1cm 1cm 1cm 1cm}, clip]{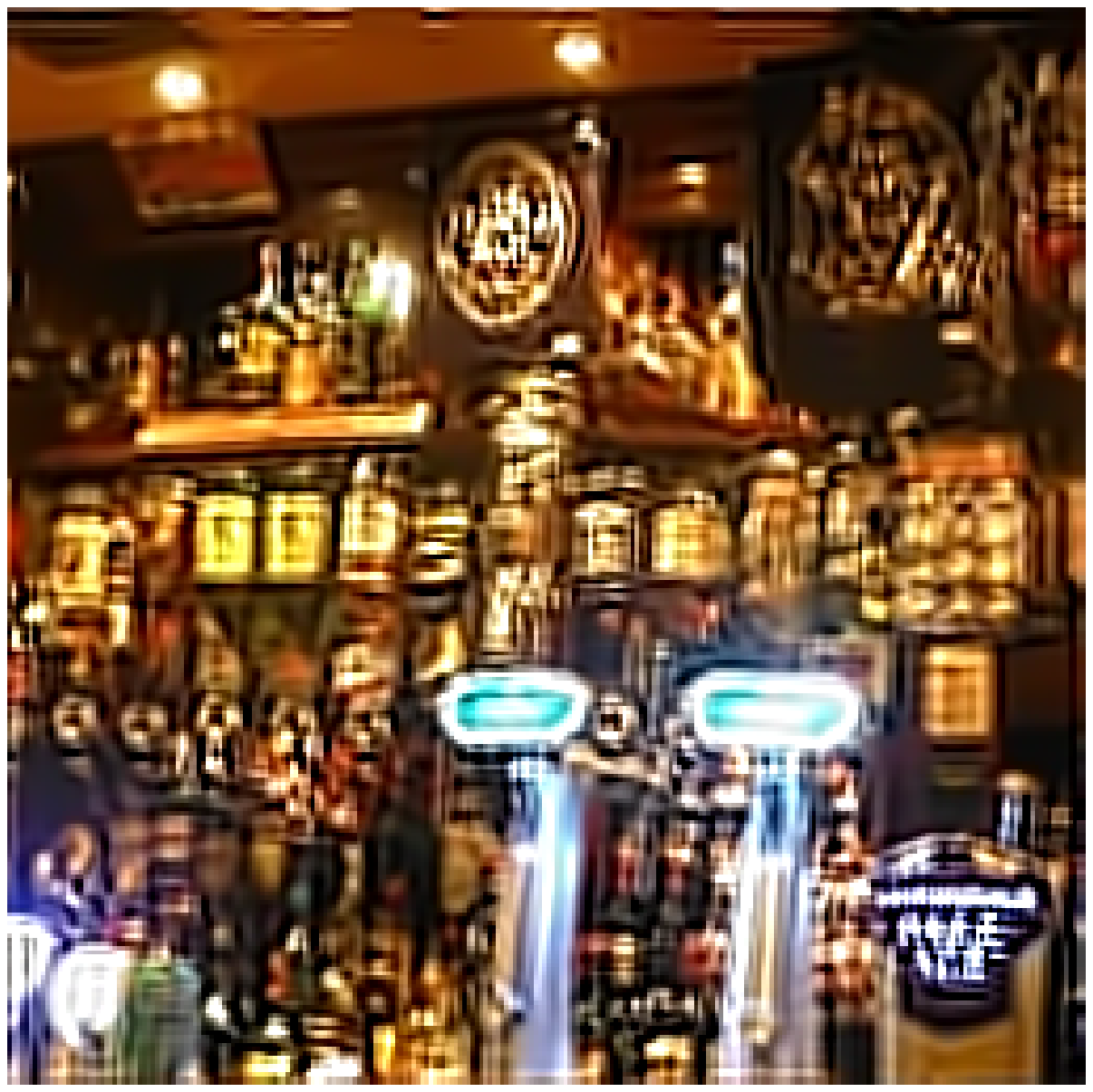}
\put(22,6)
{\includegraphics[height=1cm,trim={4cm 1.5cm 3.4cm 1cm},clip]{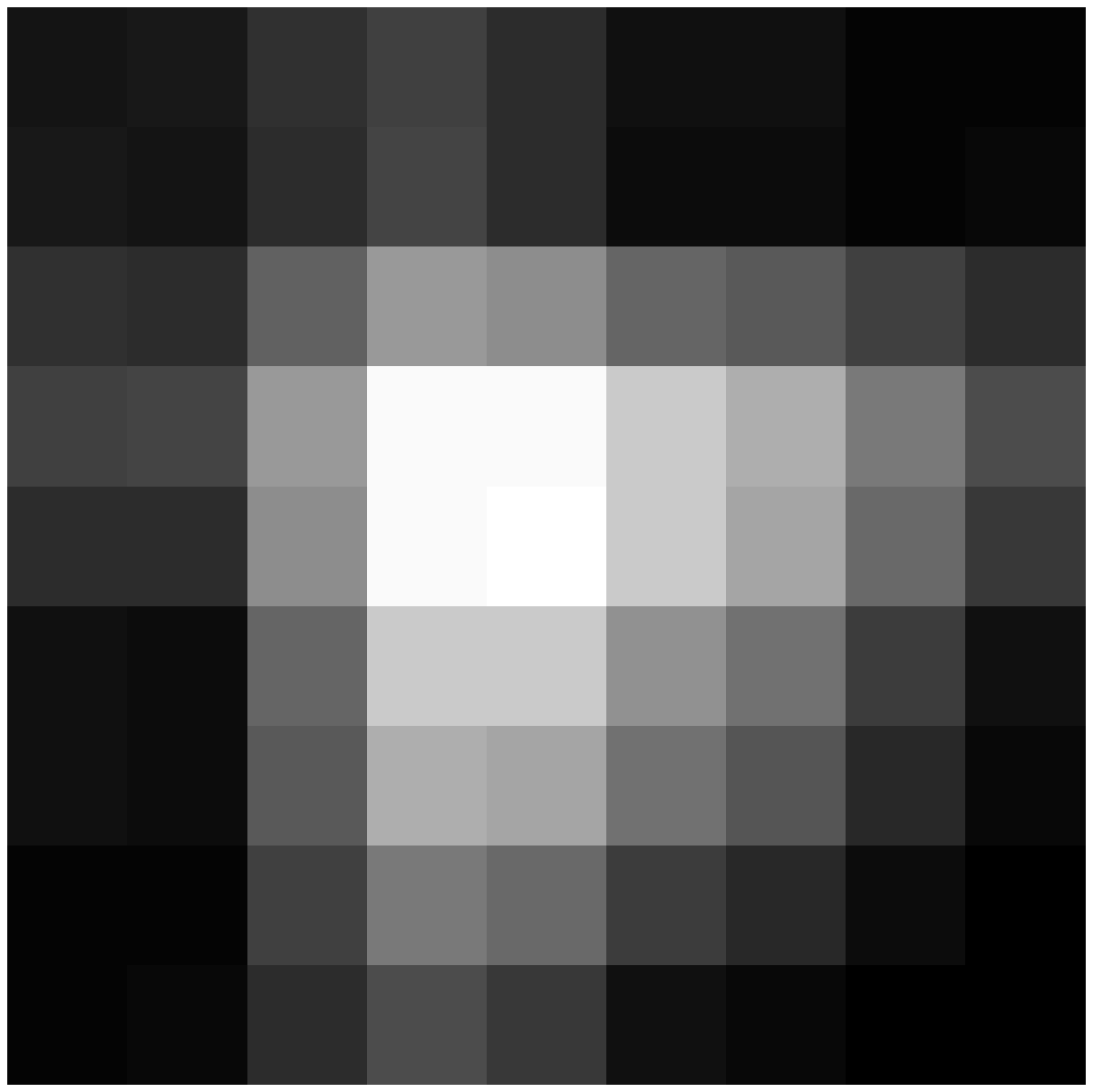}}
\end{overpic}&\hspace{-0.7cm}

\begin{overpic}
[height=3cm, trim = {1cm 1cm 1cm 1cm}, clip]{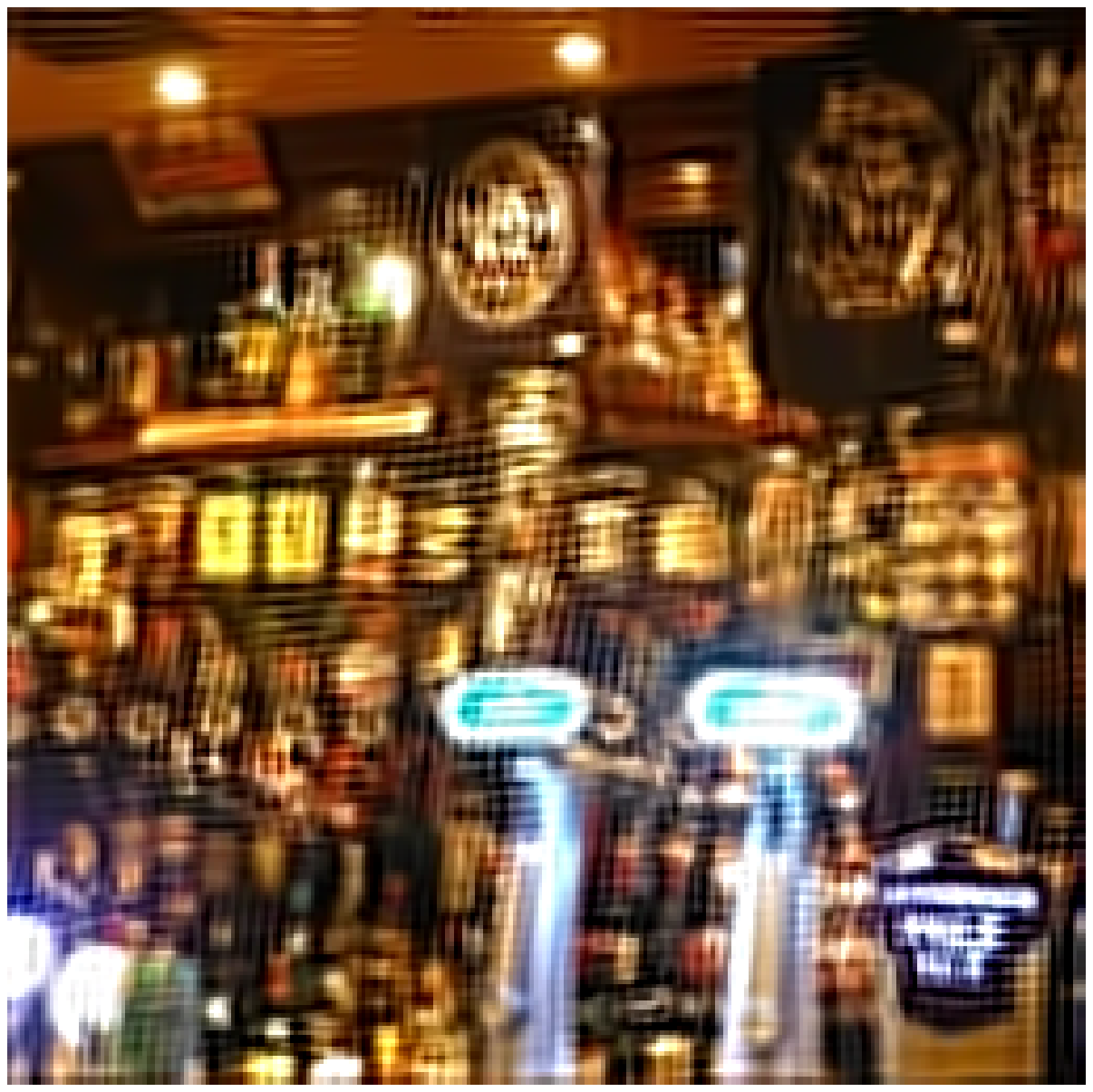}
\put(22,6)
{\includegraphics[height=1cm,trim={4cm 1.5cm 3.4cm 1cm},clip]{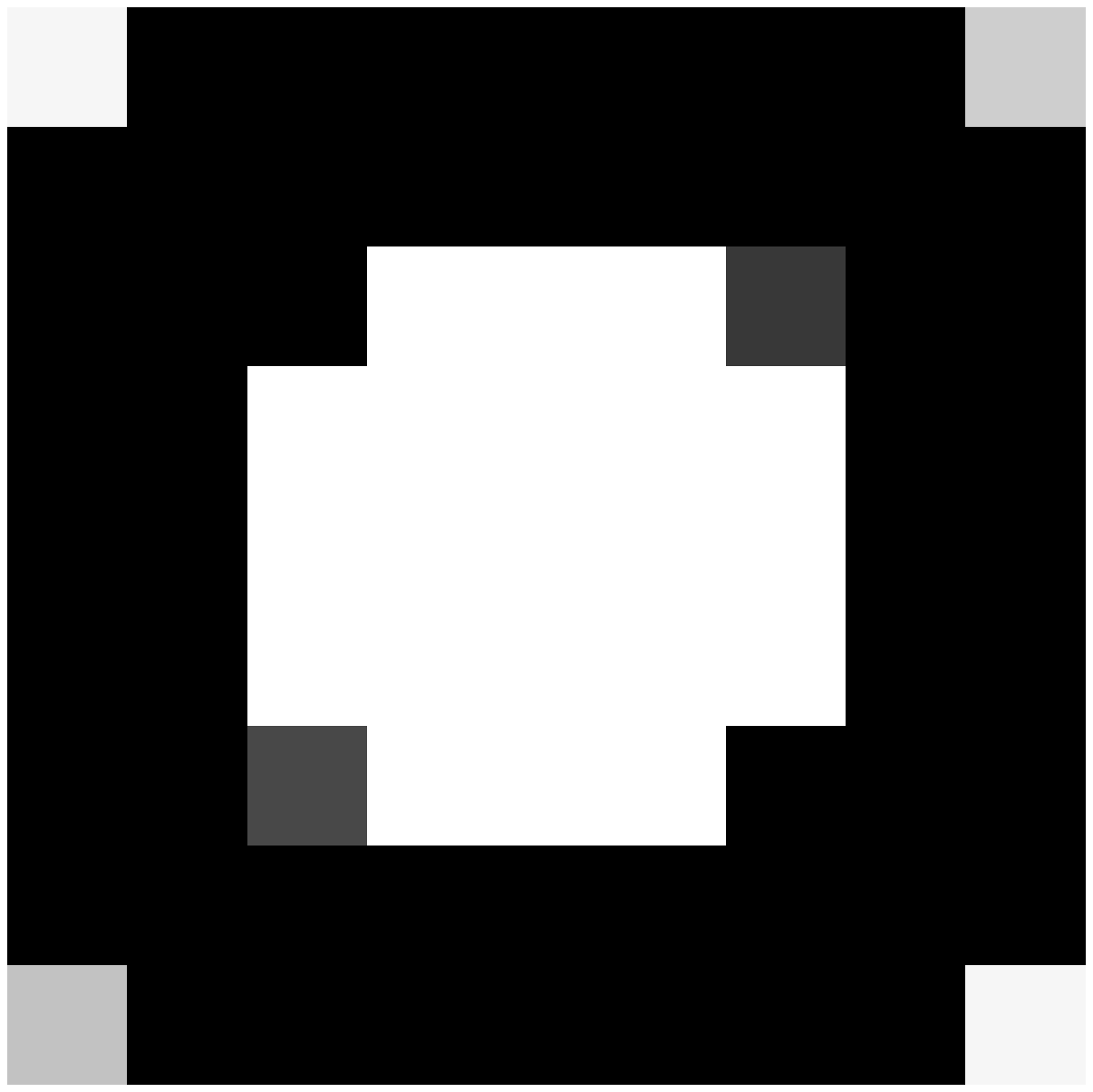}}
\end{overpic}&\hspace{-0.7cm}

\begin{overpic}
[height=3cm, trim = {1cm 1cm 1cm 1cm}, clip]{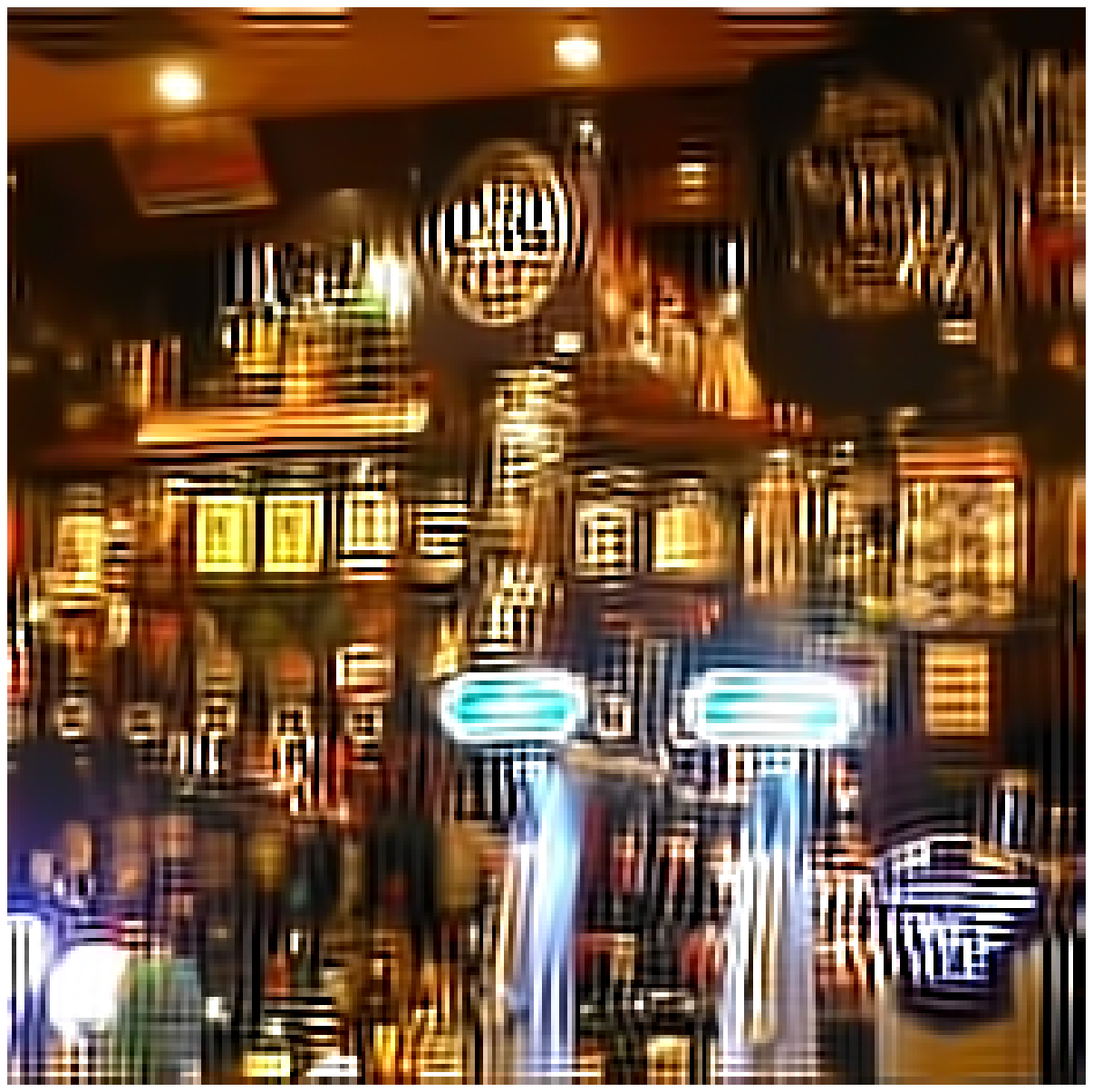}
\put(22,6)
{\includegraphics[height=1cm,trim={4cm 1.5cm 3.4cm 1cm},clip]{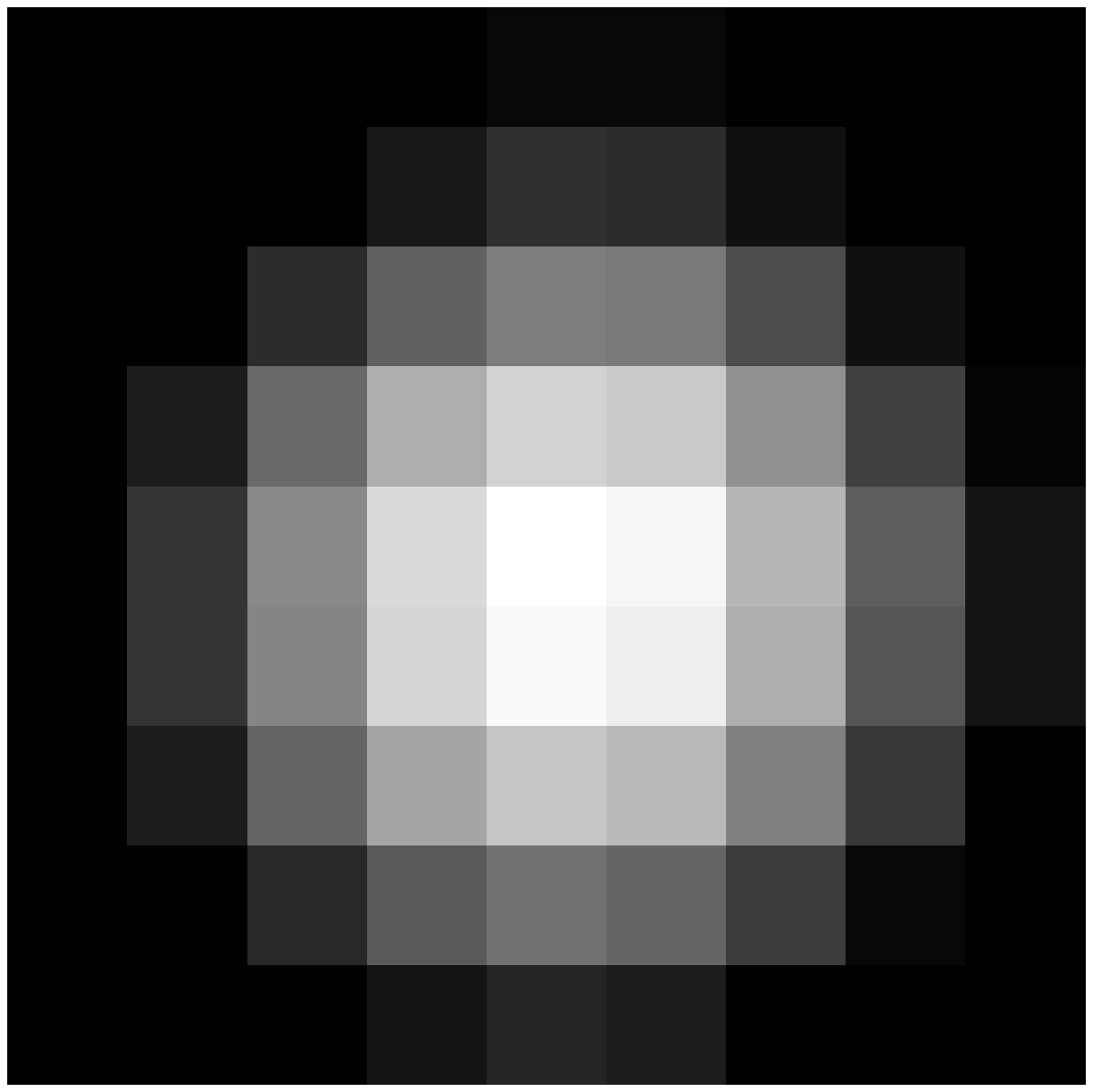}}
\end{overpic}\\

\textbf{Degraded} & \textbf{Original} & \textbf{VBA} & \textbf{deconv2D} & \textbf{blinddeconv}\\
& & MSE = 0.0115&MSE = 0.0190&MSE = 0.0075\\  
PieAPP = 3.8507 && PieAPP = \textbf{0.8278}& PieAPP = 1.6273&PieAPP = 1.0216 \\

\begin{overpic}
[height=3cm, trim = {1cm 1cm 1cm 1cm}, clip]{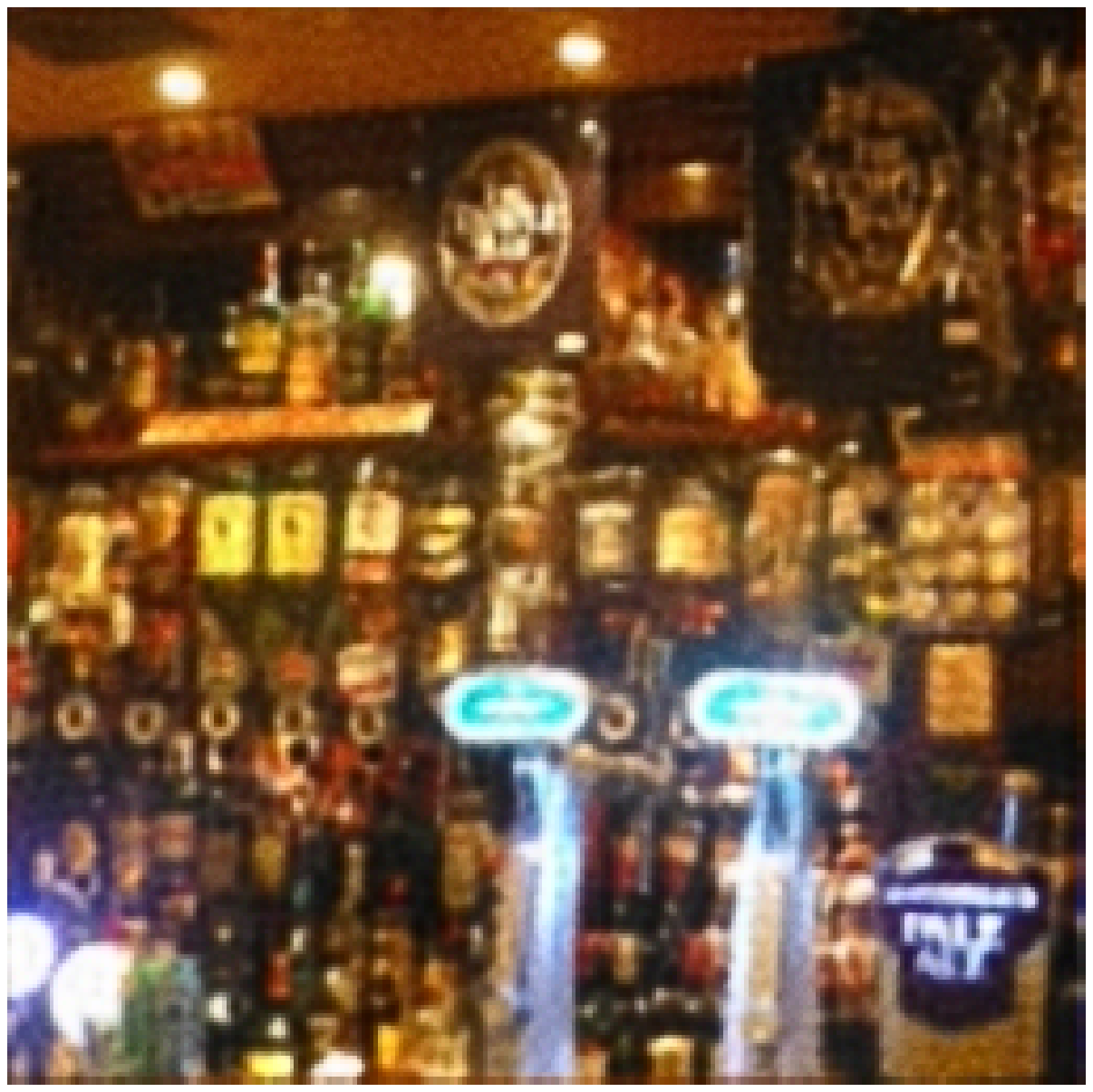}
\put(22,6)
{\includegraphics[height=1cm,trim={4cm 1.5cm 3.4cm 1cm},clip]{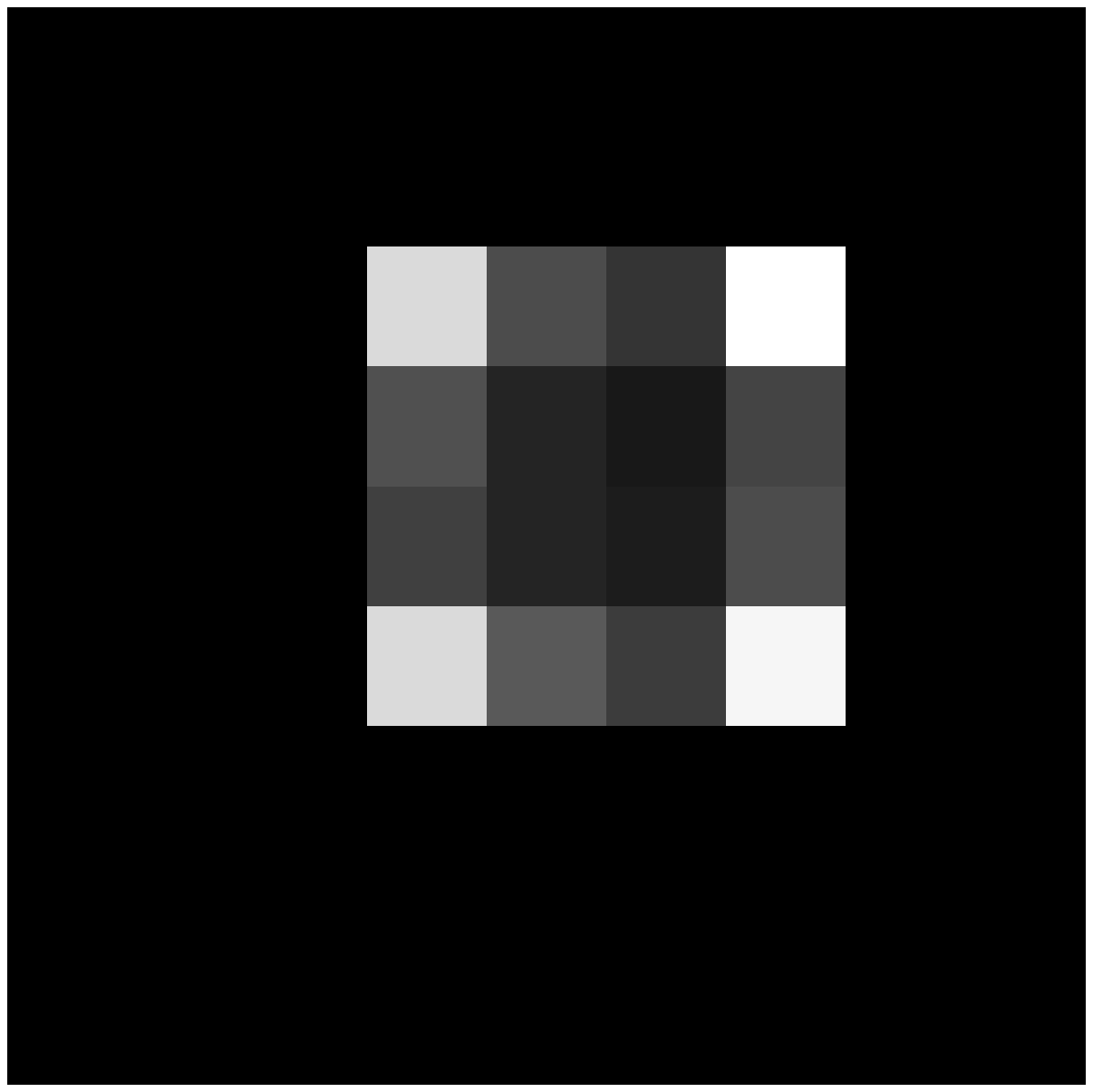}}
\end{overpic}&\hspace{-0.7cm}

\includegraphics[height=3cm, trim = {1cm 1cm 1cm 1cm}, clip]{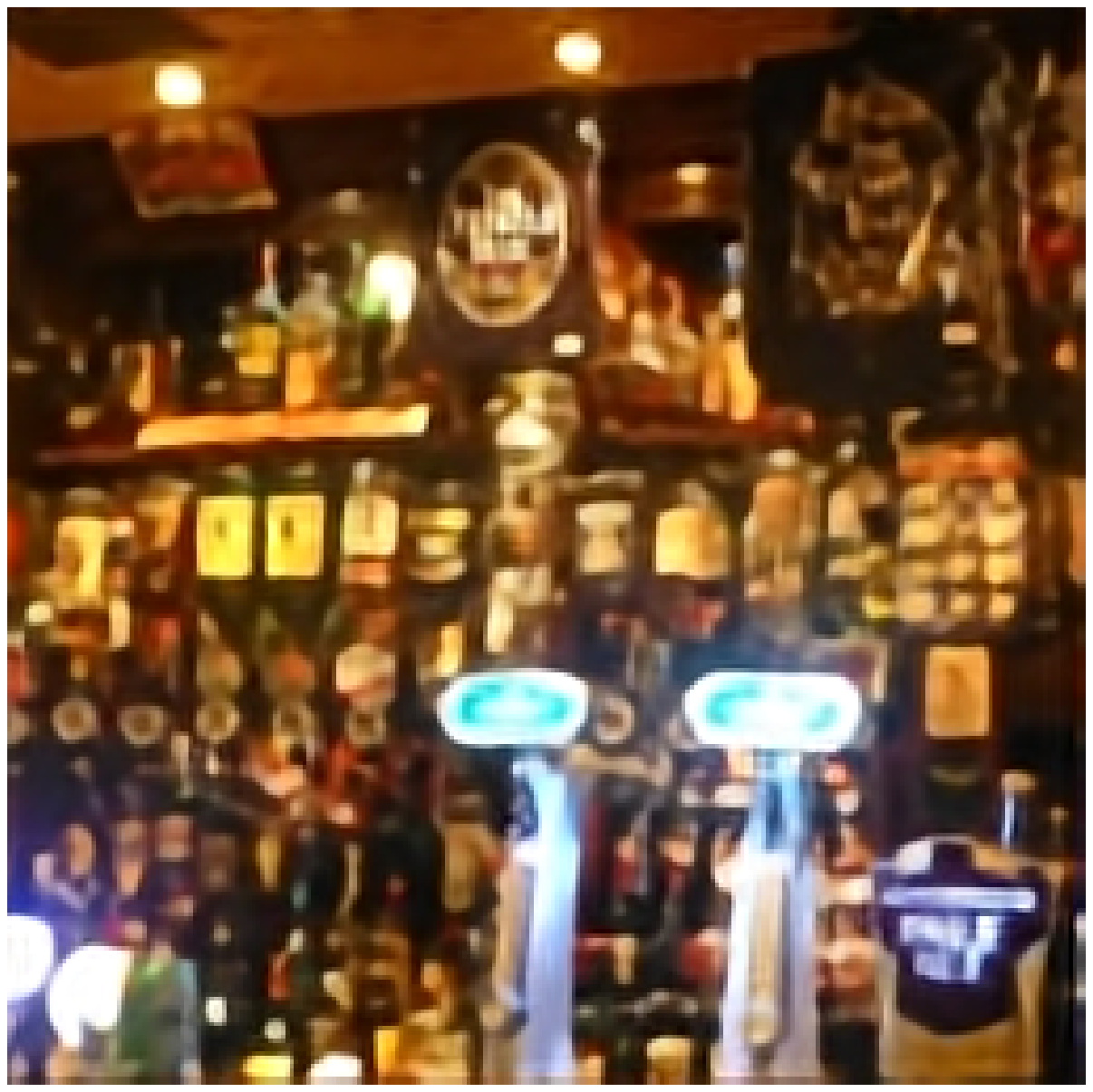}
&\hspace{-0.7cm}

\includegraphics[height=3cm, trim = {1cm 1cm 1cm 1cm}, clip]{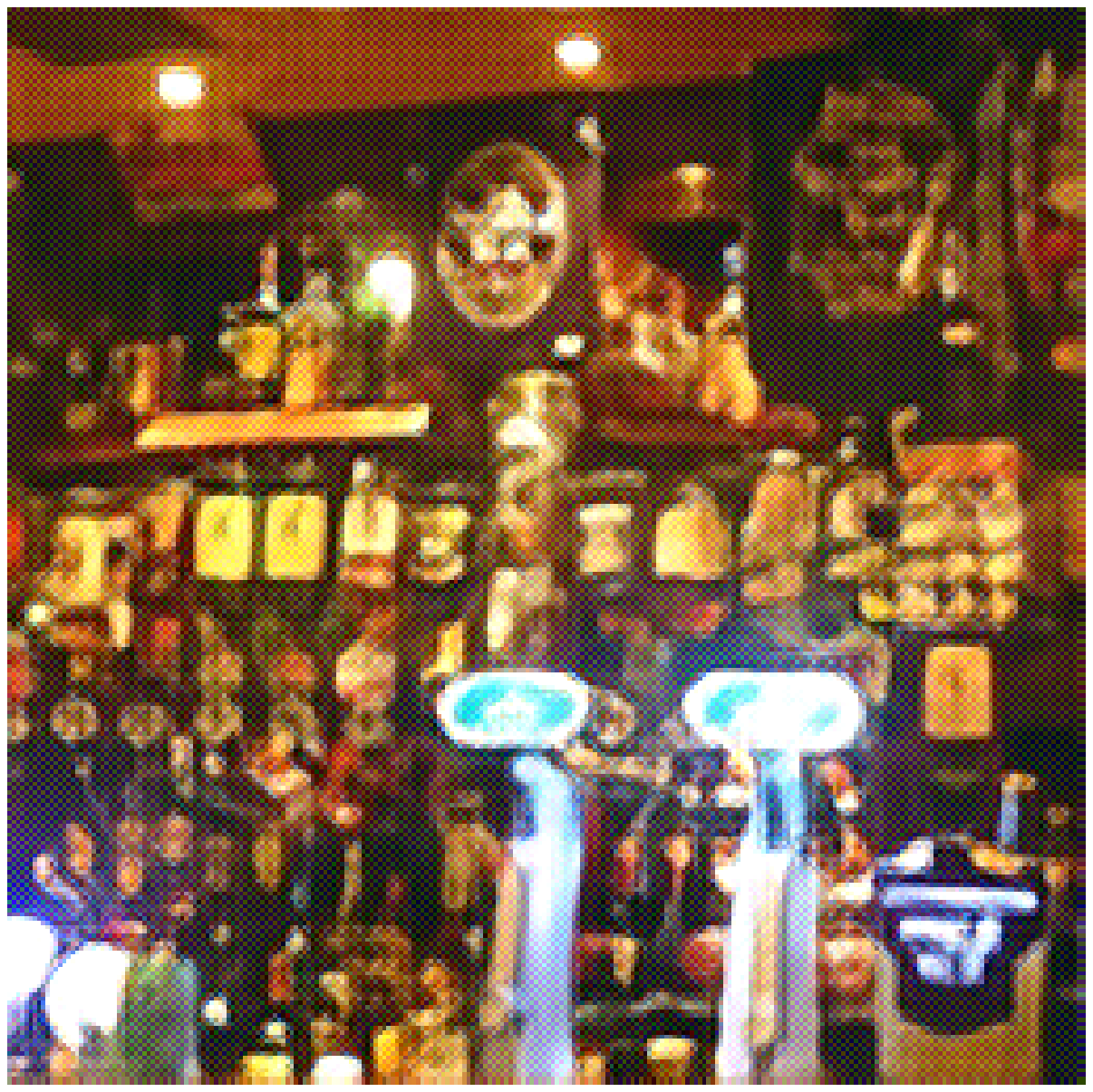}
&\hspace{-0.7cm}

\begin{overpic}
[height=3cm, trim = {1cm 1cm 1cm 1cm}, clip]{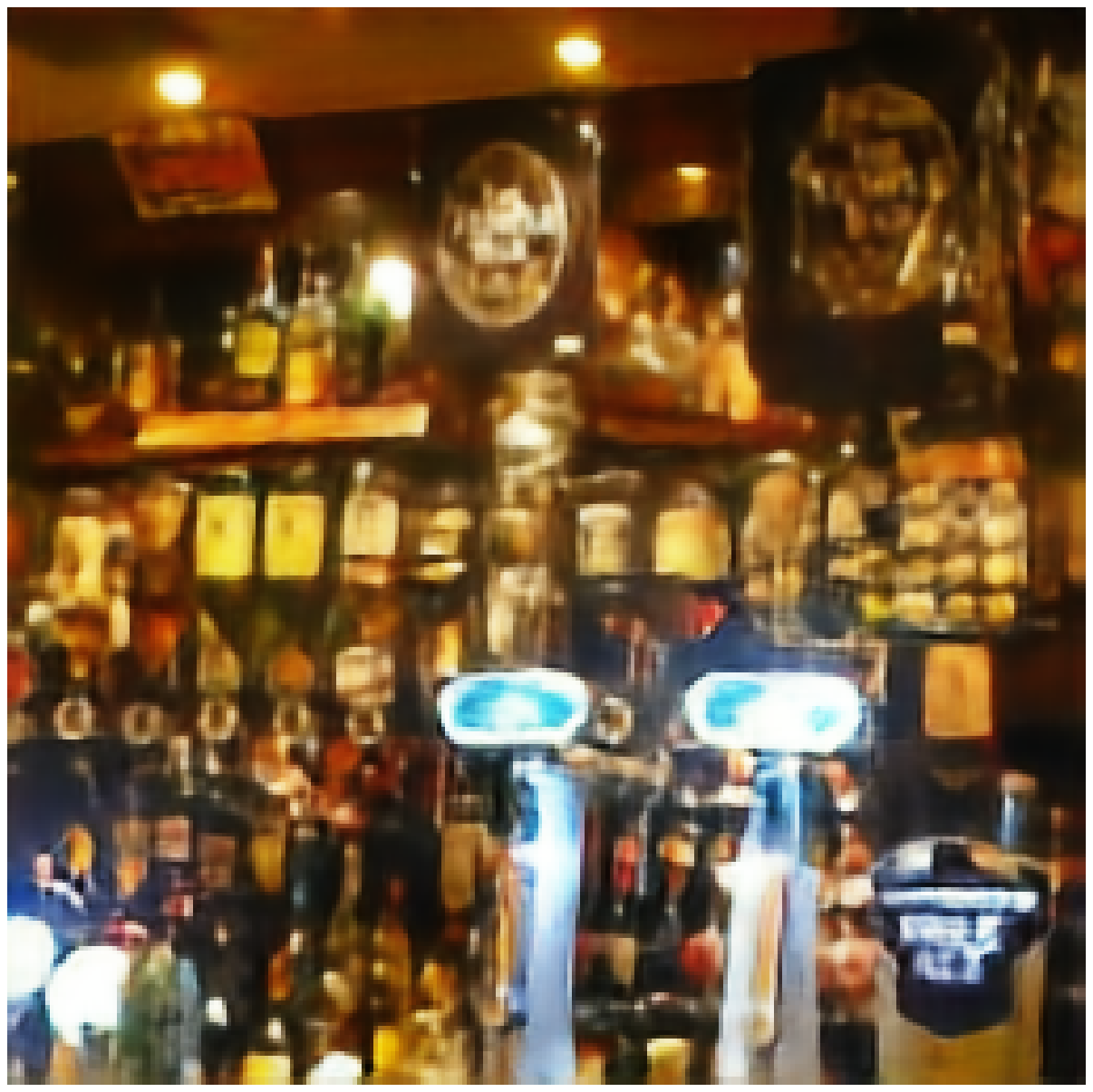}
\put(22,6)
{\includegraphics[height=1cm,trim={4cm 1.5cm 3.4cm 1cm},clip]{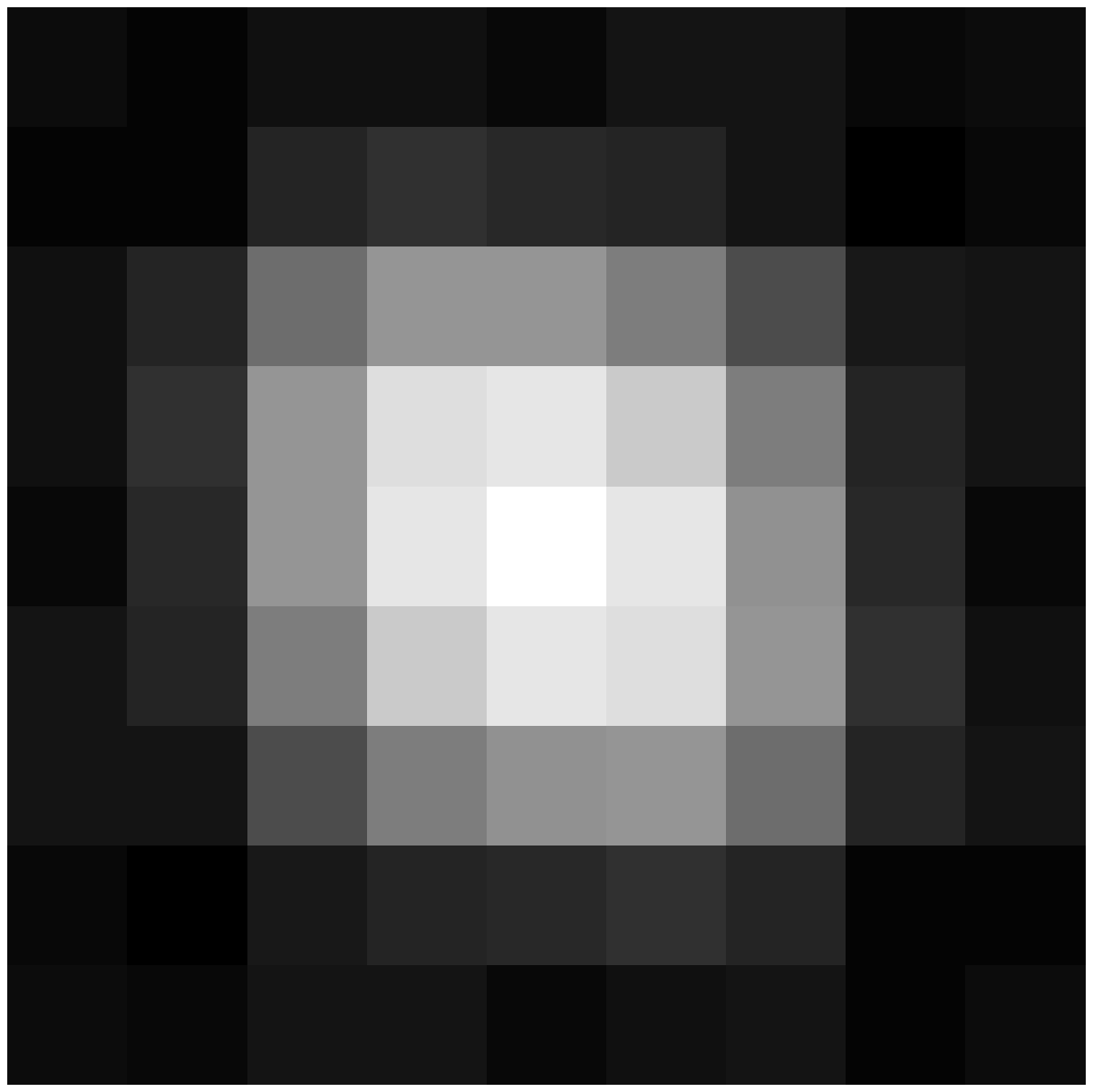}}
\end{overpic}&\hspace{-0.7cm}

\begin{overpic}
[height=3cm, trim = {1cm 1cm 1cm 1cm}, clip]{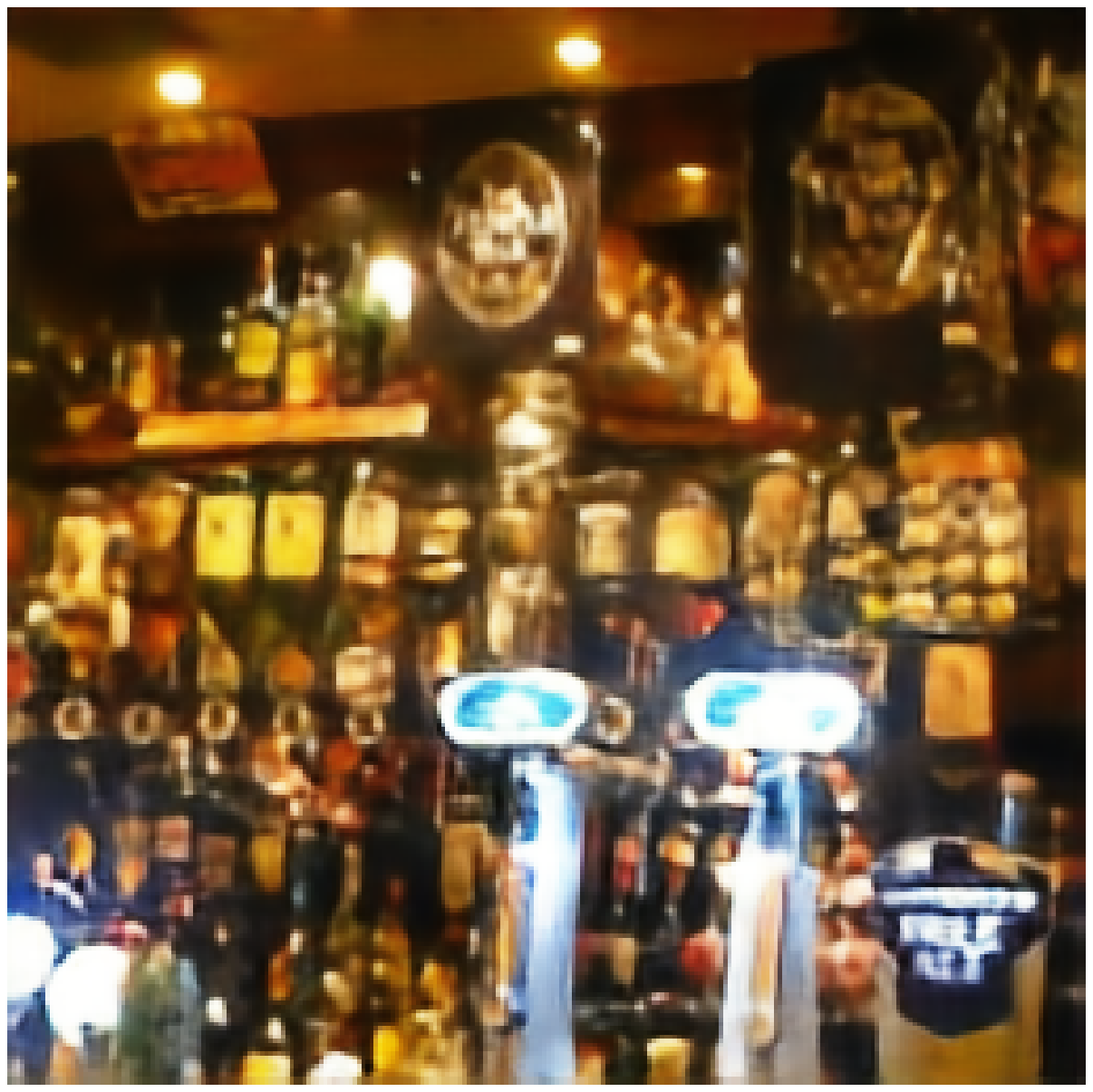}
\put(22,6)
{\includegraphics[height=1cm,trim={4cm 1.5cm 3.4cm 1cm},clip]{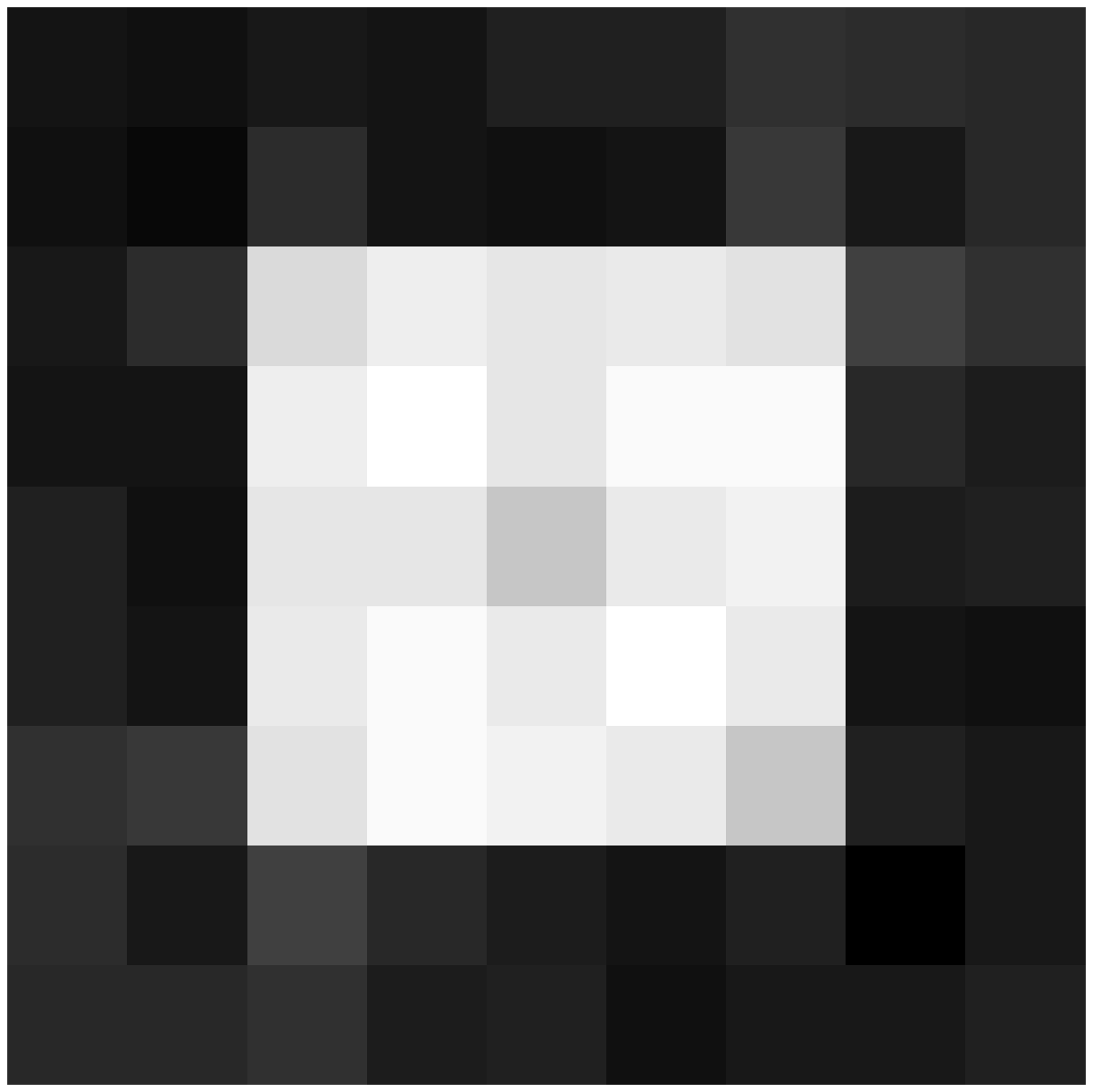}}
\end{overpic}\\

\textbf{SelfDeblur} & \textbf{DBSRCNN} & \textbf{DeblurGAN} & \textbf{proposed (greedy)} & \textbf{proposed (end-to-end)} \\
MSE = 3.6152& & &MSE = \textbf{0.0015}&MSE = 0.0020 \\
PieAPP = 8.1479 &PieAPP = 1.2058 &PieAPP = 2.6971&PieAPP = 1.2793&PieAPP = 1.1663 \\ 

%
\includegraphics[height=3cm, trim = {1cm 1cm 1cm 1cm}, clip]{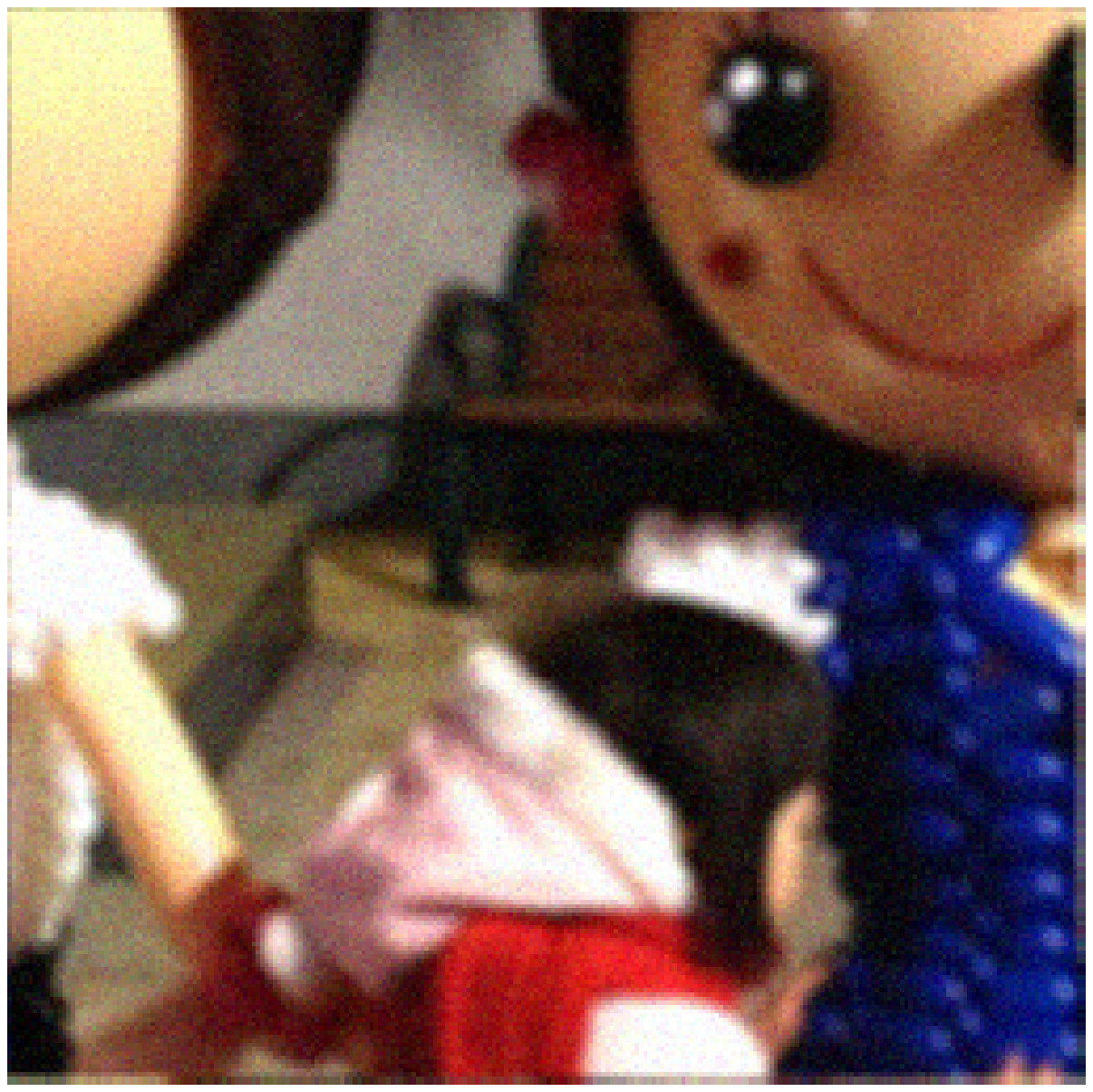} &\hspace{-0.7cm}

\begin{overpic}
[height=3cm, trim = {1cm 1cm 1cm 1cm}, clip]{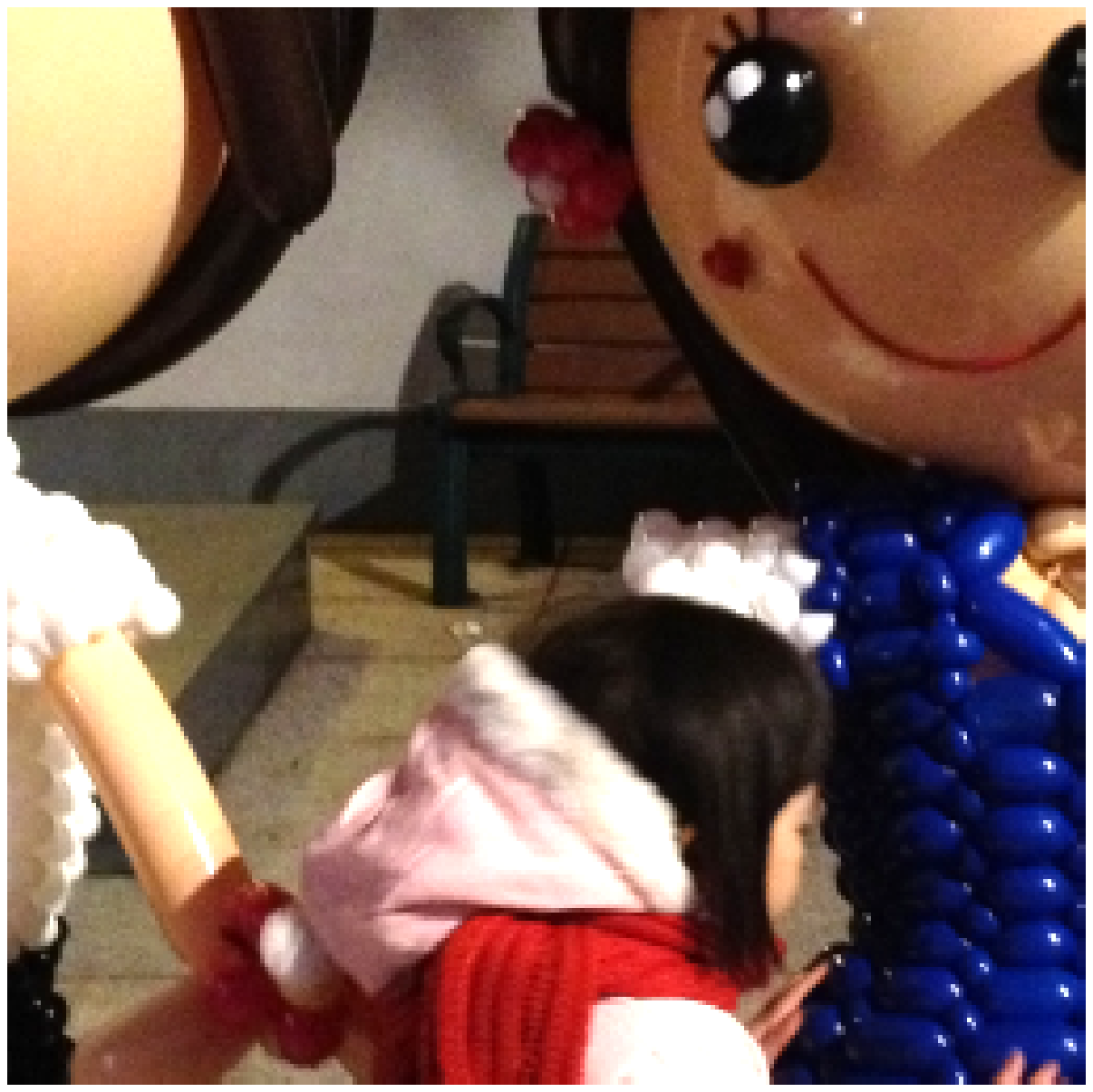}
\put(22,6)
{\includegraphics[height=1cm,trim={4cm 1.5cm 3.4cm 1cm},clip]{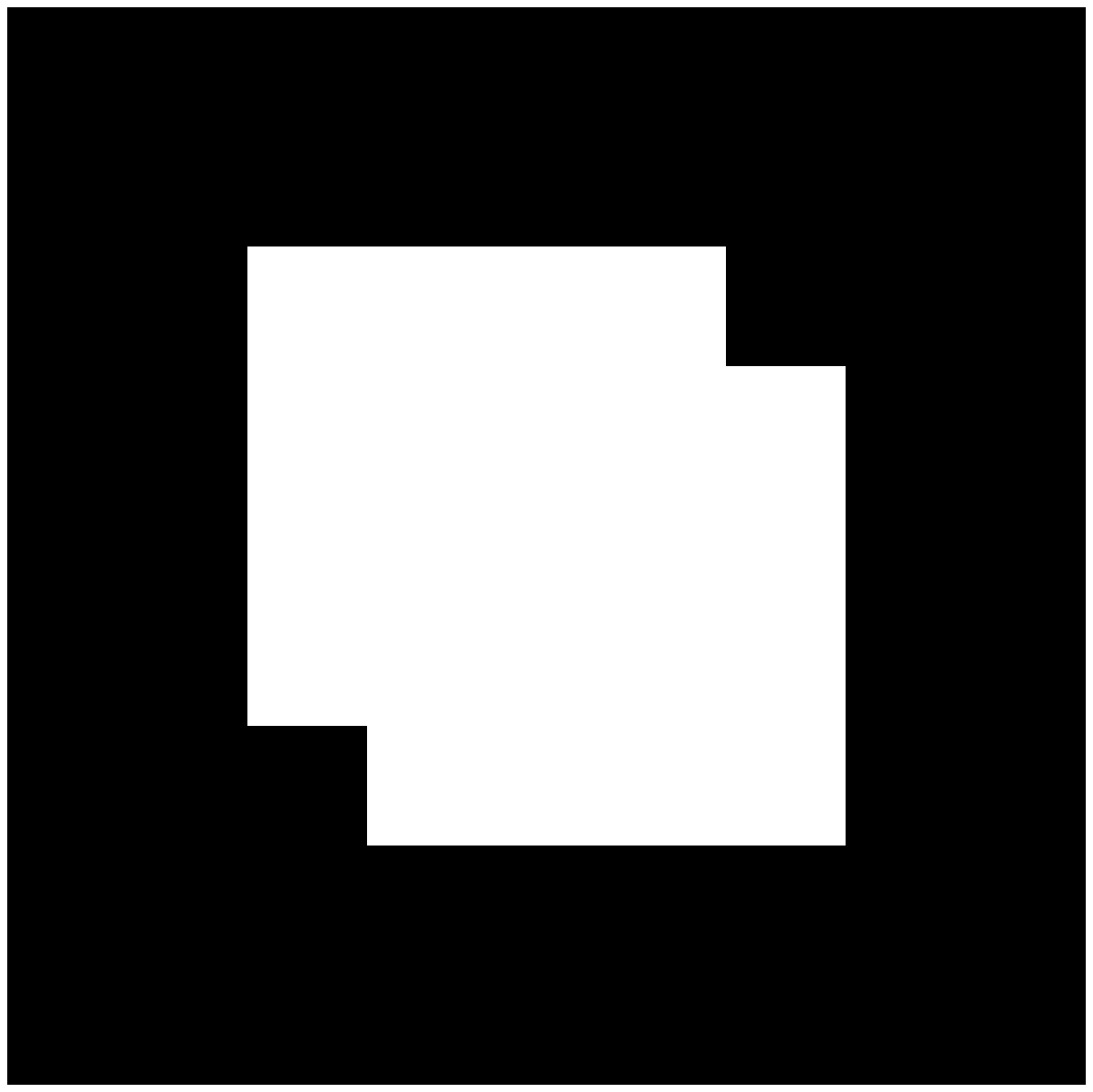}}
\end{overpic}&\hspace{-0.7cm}

\begin{overpic}
[height=3cm, trim = {1cm 1cm 1cm 1cm}, clip]{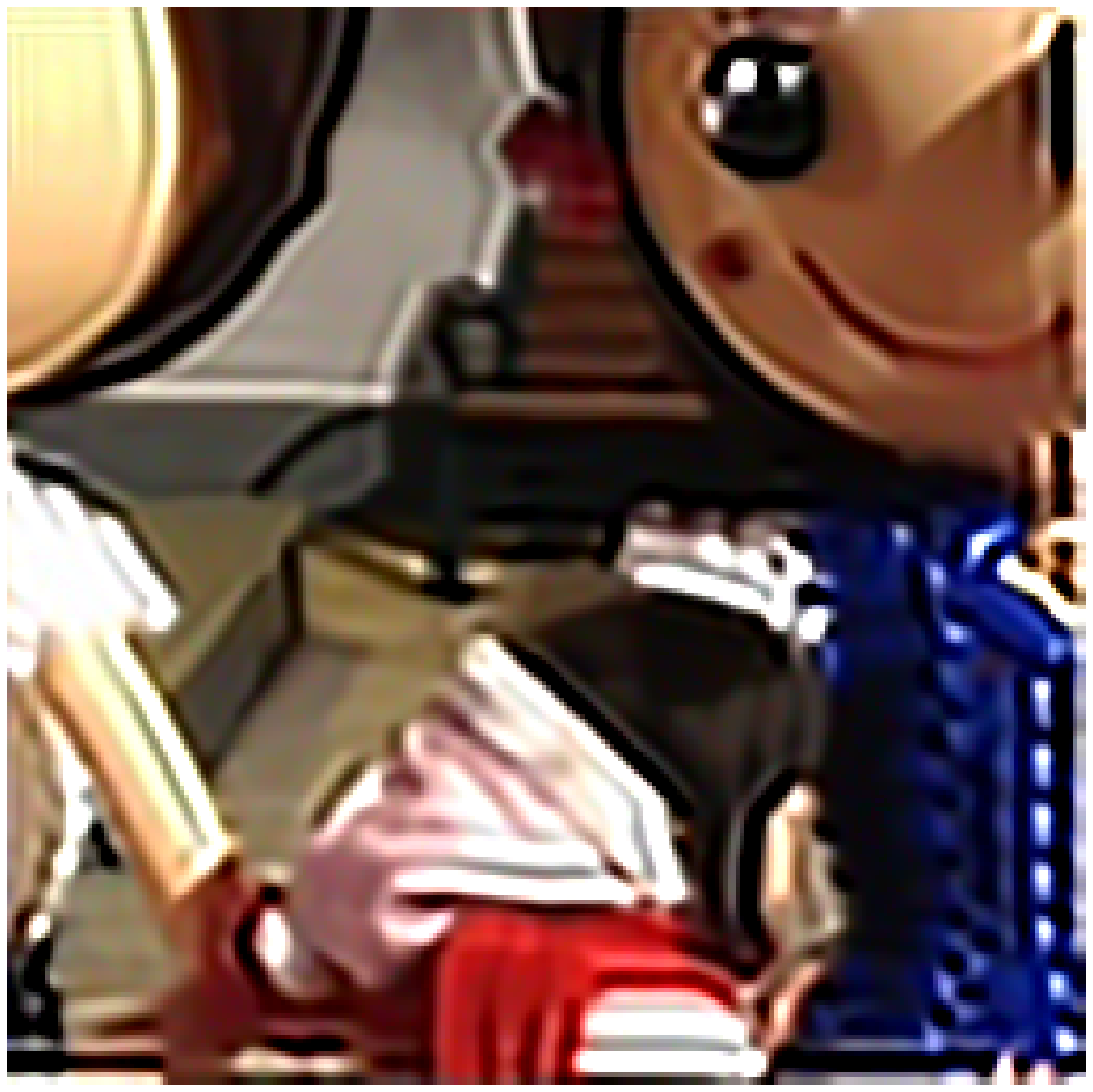}
\put(22,6)
{\includegraphics[height=1cm,trim={4cm 1.5cm 3.4cm 1cm},clip]{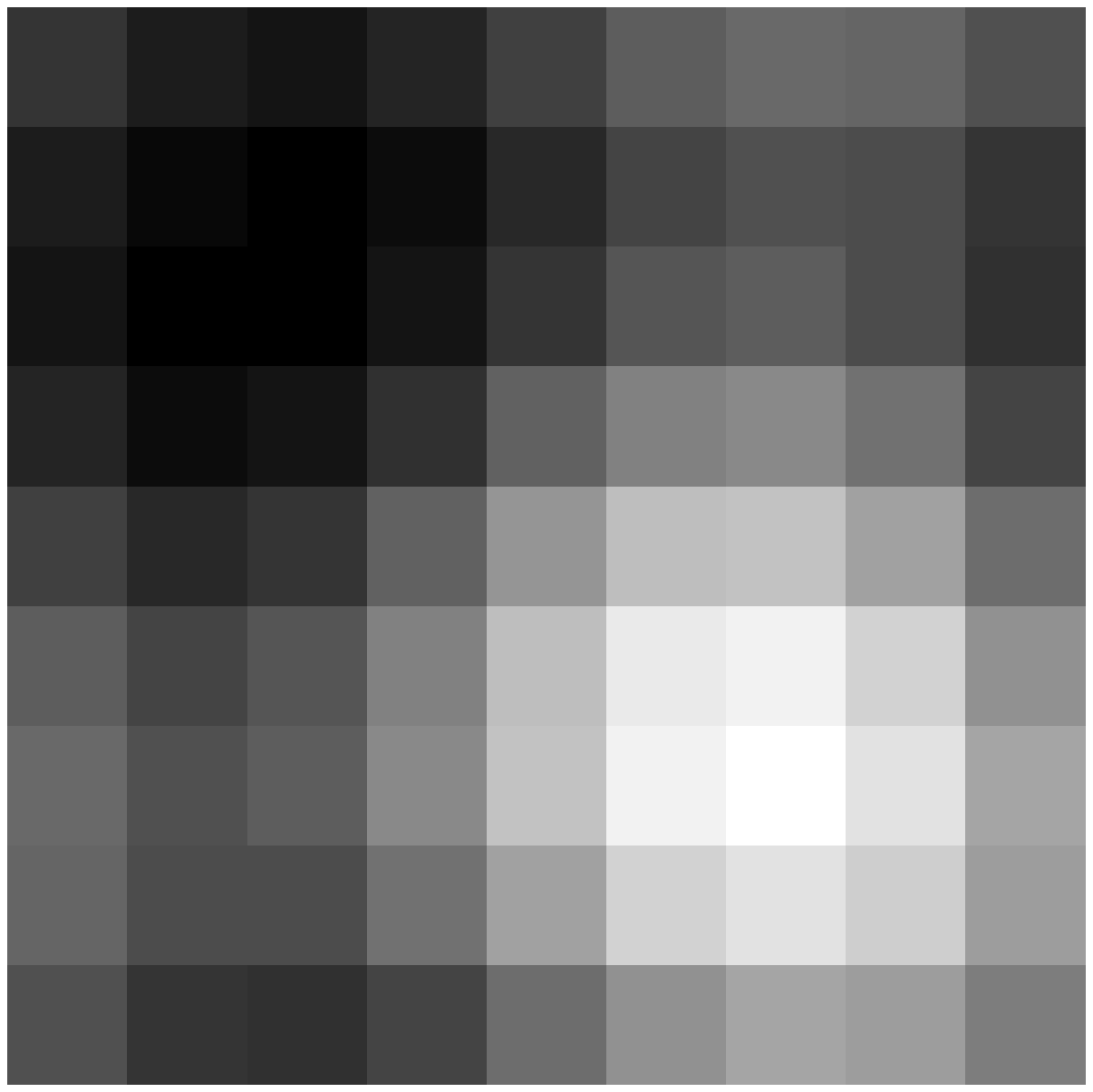}}
\end{overpic}&\hspace{-0.7cm}

\begin{overpic}
[height=3cm, trim = {1cm 1cm 1cm 1cm}, clip]{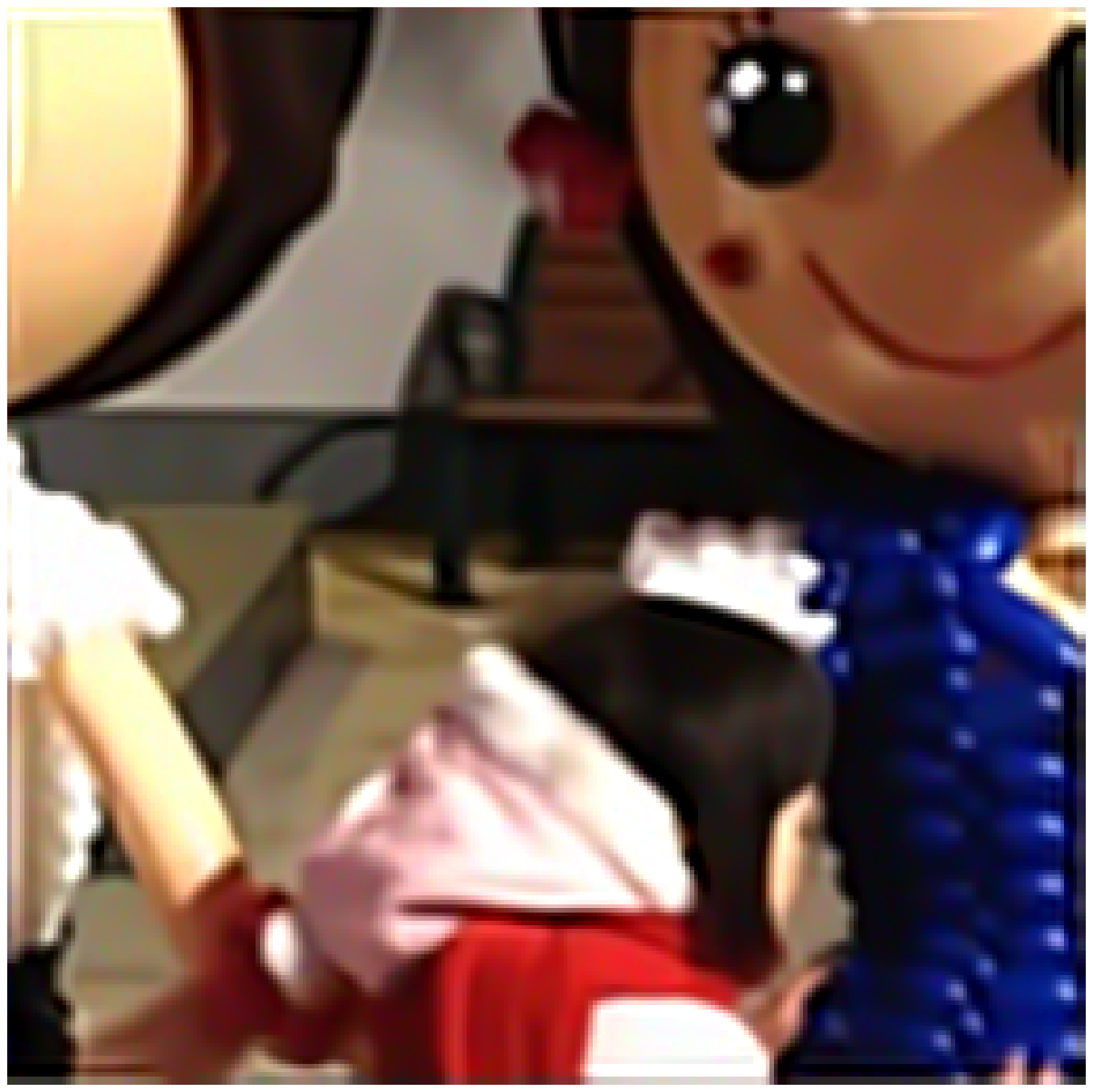}
\put(22,6)
{\includegraphics[height=1cm,trim={4cm 1.5cm 3.4cm 1cm},clip]{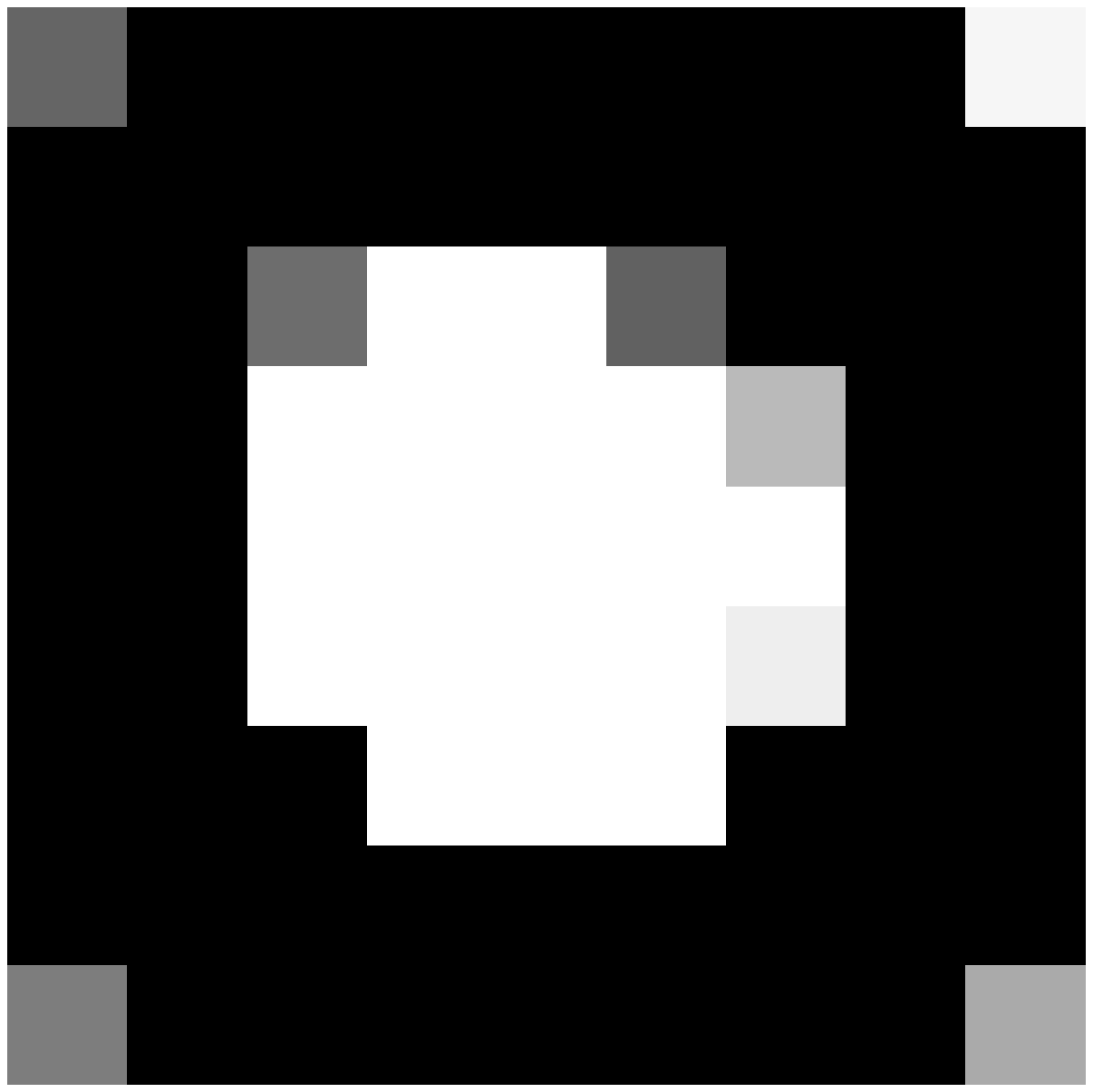}}
\end{overpic}&\hspace{-0.7cm}

\begin{overpic}
[height=3cm, trim = {1cm 1cm 1cm 1cm}, clip]{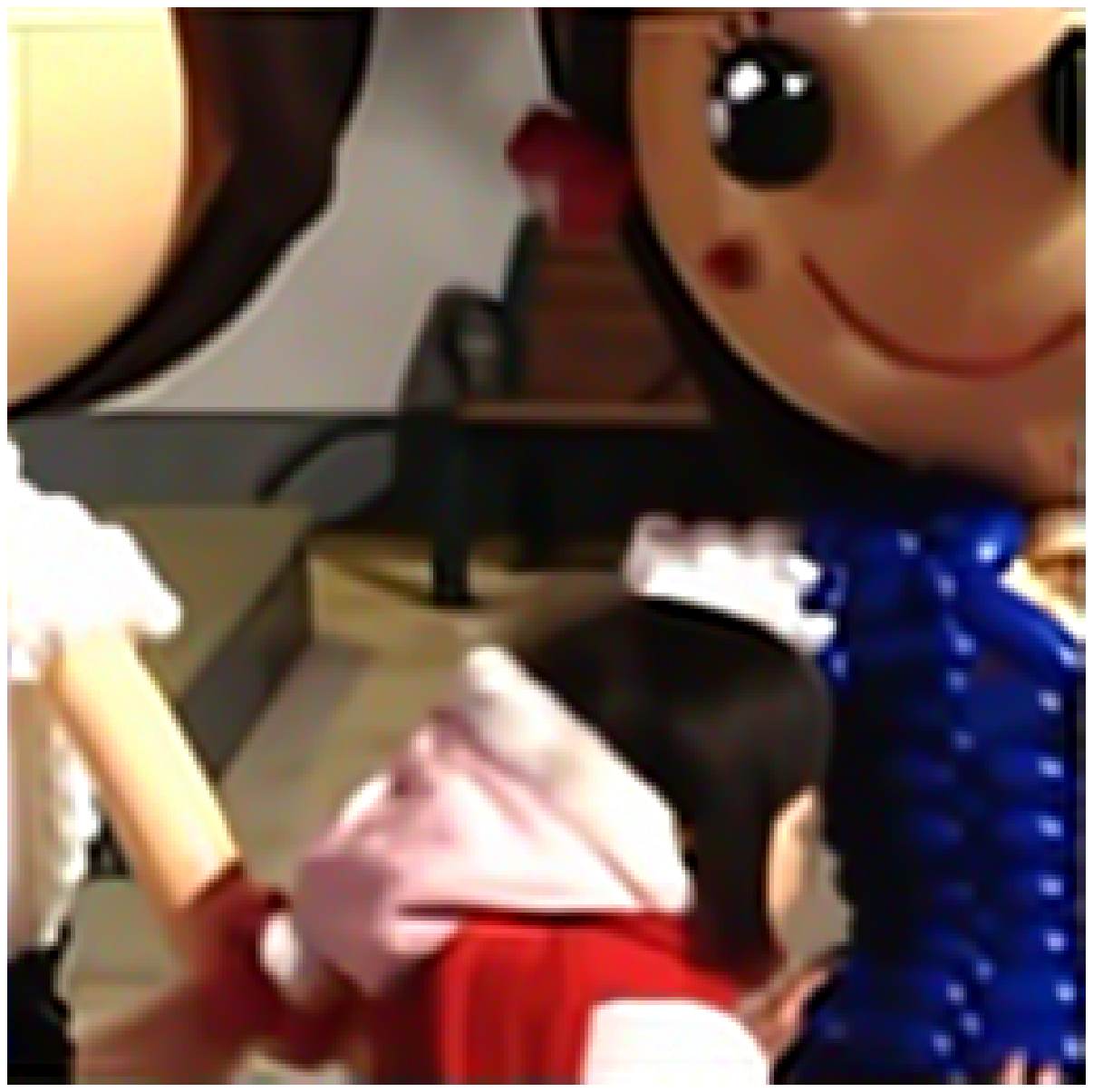}
\put(22,6)
{\includegraphics[height=1cm,trim={4cm 1.5cm 3.4cm 1cm},clip]{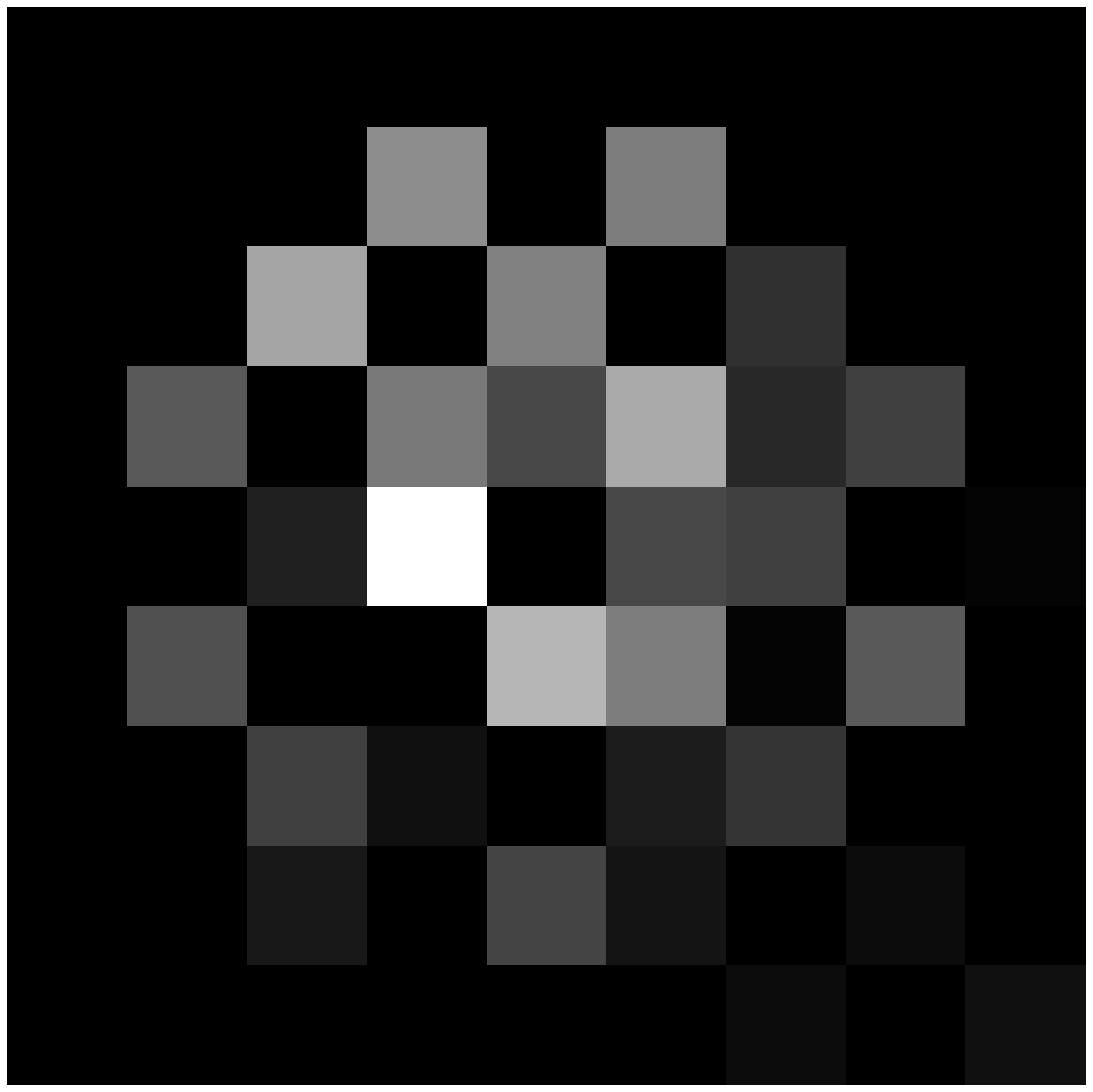}}
\end{overpic}\\

\textbf{Degraded} & \textbf{Original}& \textbf{VBA} & \textbf{deconv2D} & \textbf{blinddeconv}\\
& & MSE= 0.0293&MSE = 0.0184&MSE = 0.0388\\  
PieAPP = 2.8997 && PieAPP = 1.3625 & PieAPP = 1.0608&PieAPP = 1.0318 \\

\begin{overpic}
[height=3cm, trim = {1cm 1cm 1cm 1cm}, clip]{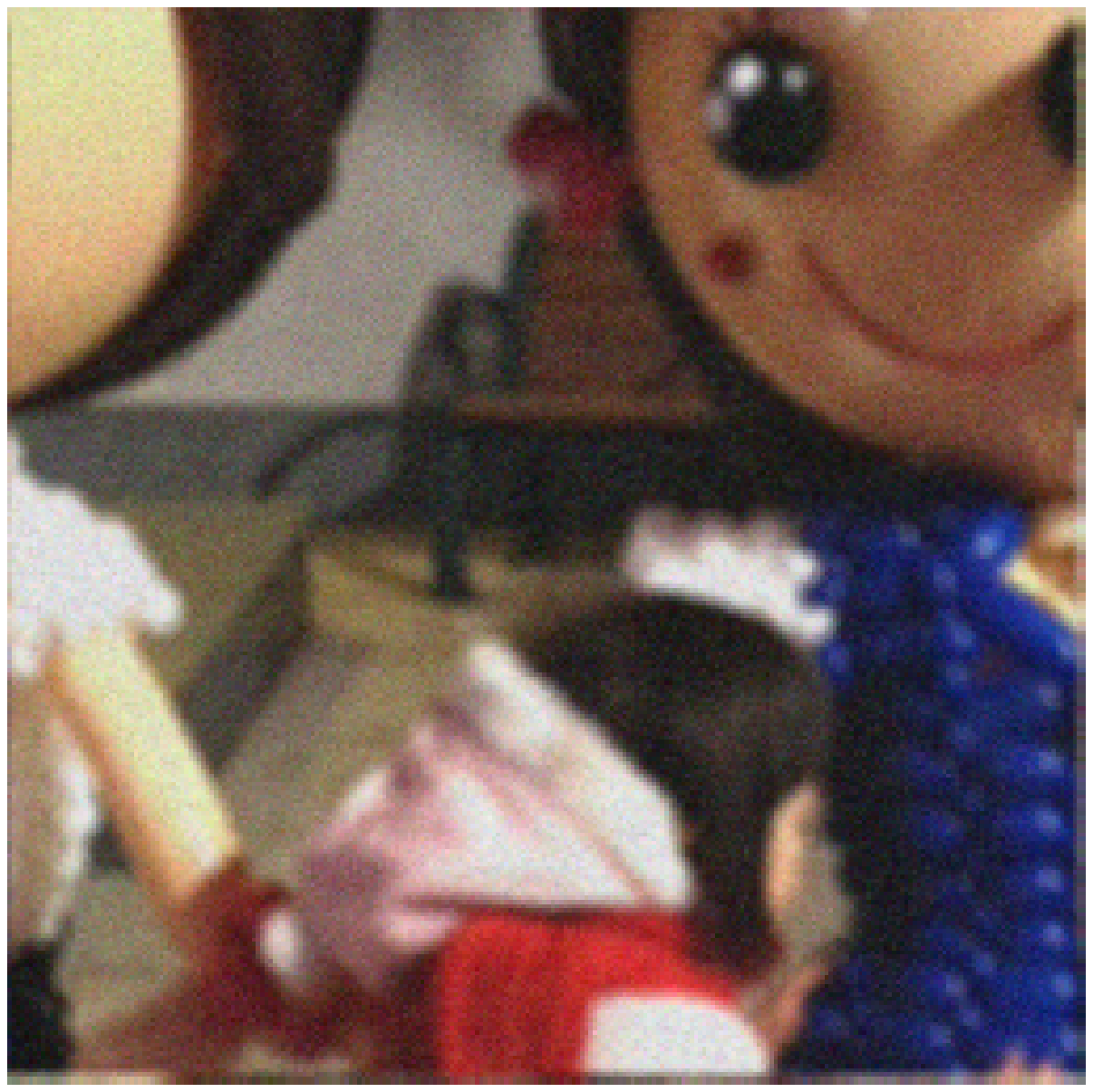}
\put(22,6)
{\includegraphics[height=1cm,trim={4cm 1.5cm 3.4cm 1cm},clip]{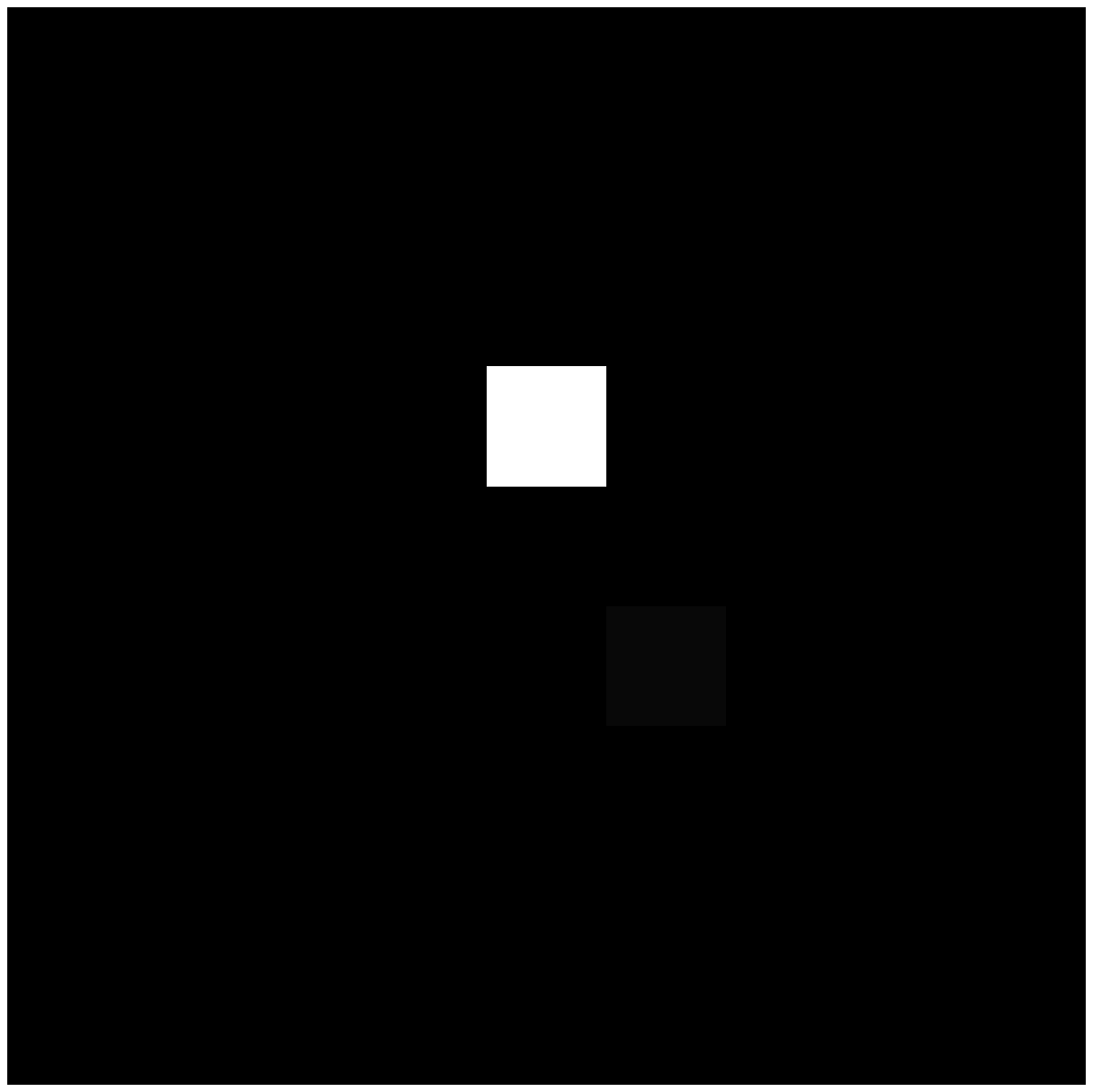}}
\end{overpic}&\hspace{-0.7cm}

\includegraphics[height=3cm, trim = {1cm 1cm 1cm 1cm}, clip]{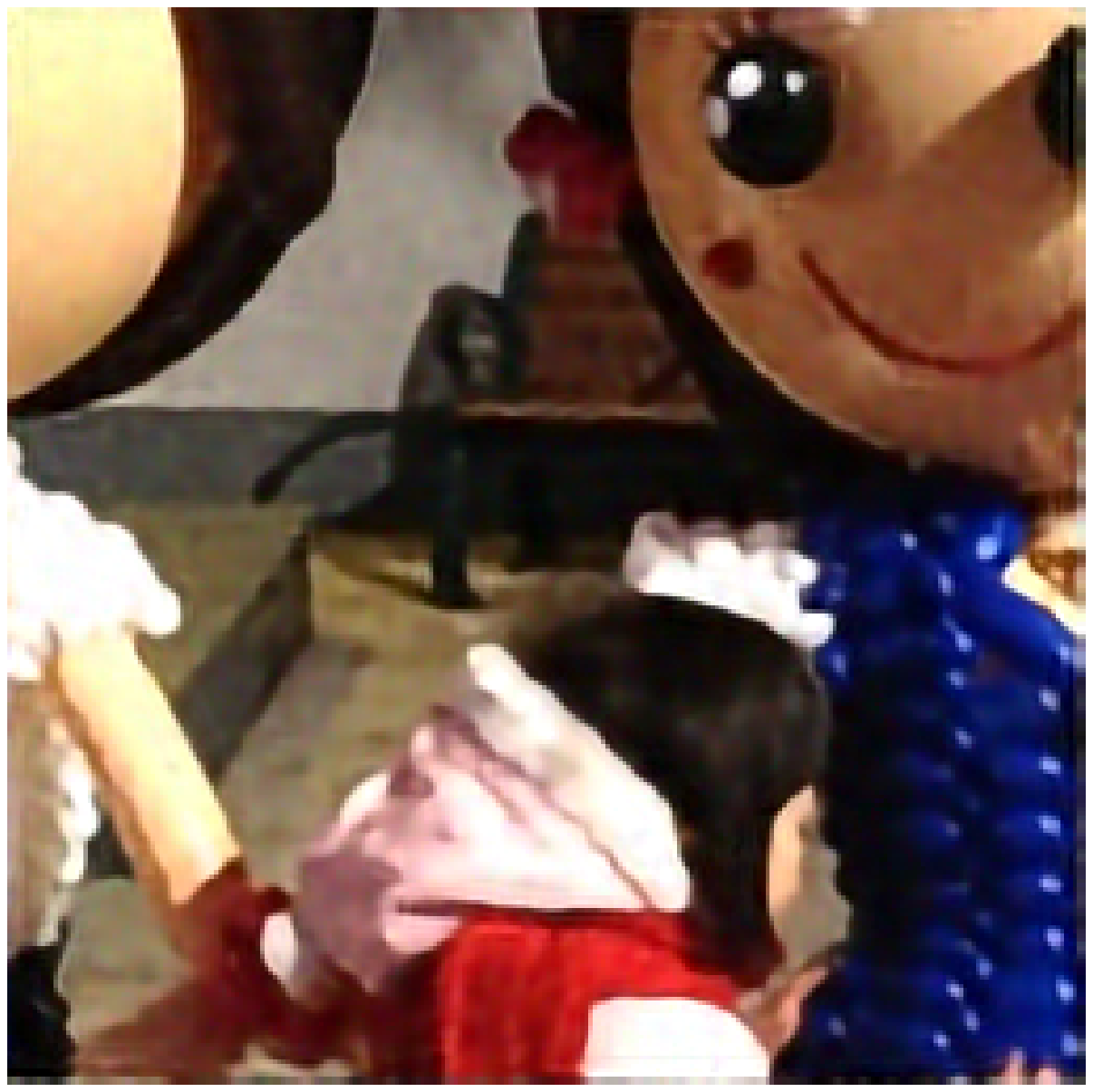}
&\hspace{-0.7cm}

\includegraphics[height=3cm, trim = {1cm 1cm 1cm 1cm}, clip]{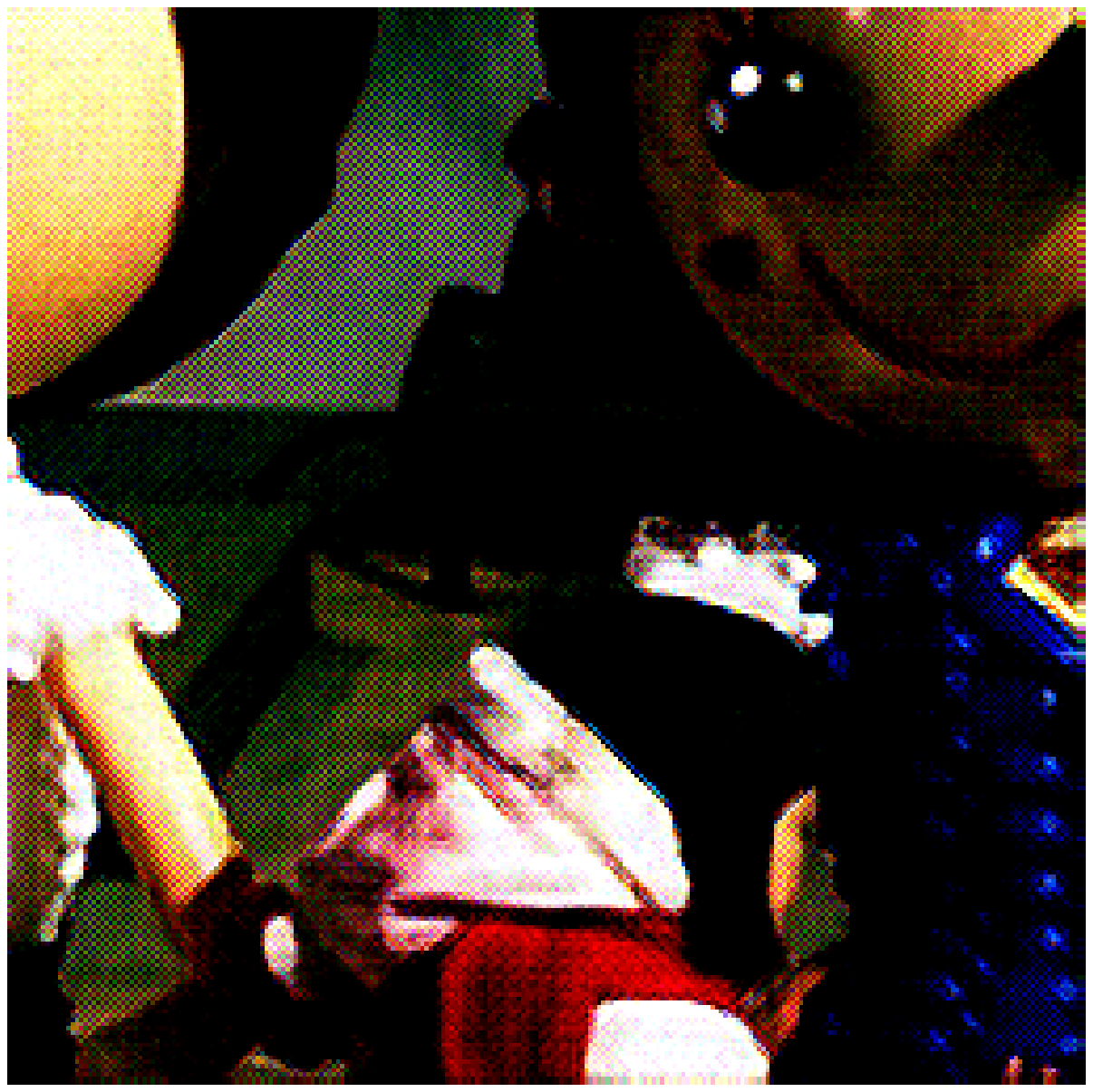}
&\hspace{-0.7cm}

\begin{overpic}
[height=3cm, trim = {1cm 1cm 1cm 1cm}, clip]{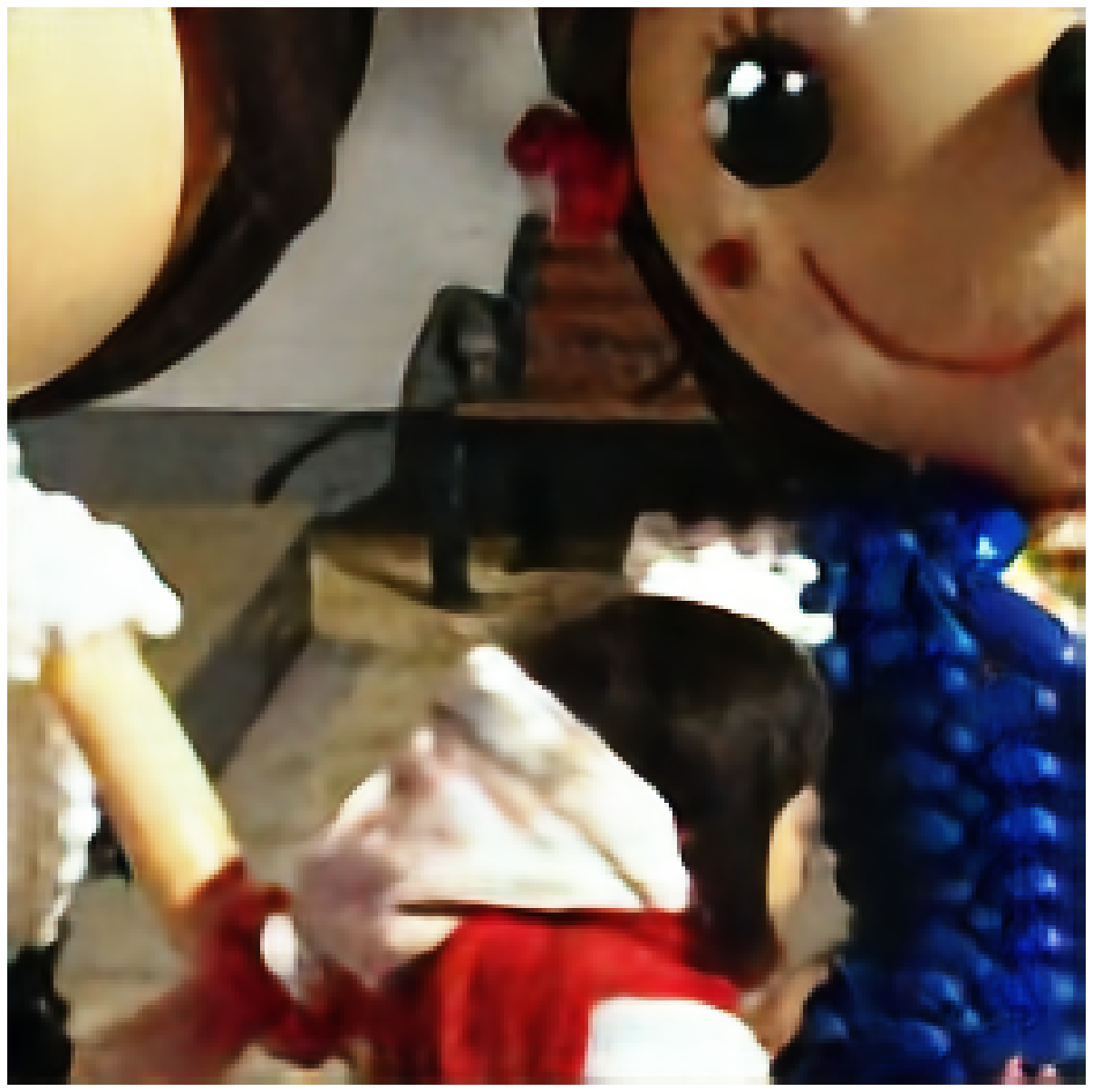}
\put(22,6)
{\includegraphics[height=1cm,trim={4cm 1.5cm 3.4cm 1cm},clip]{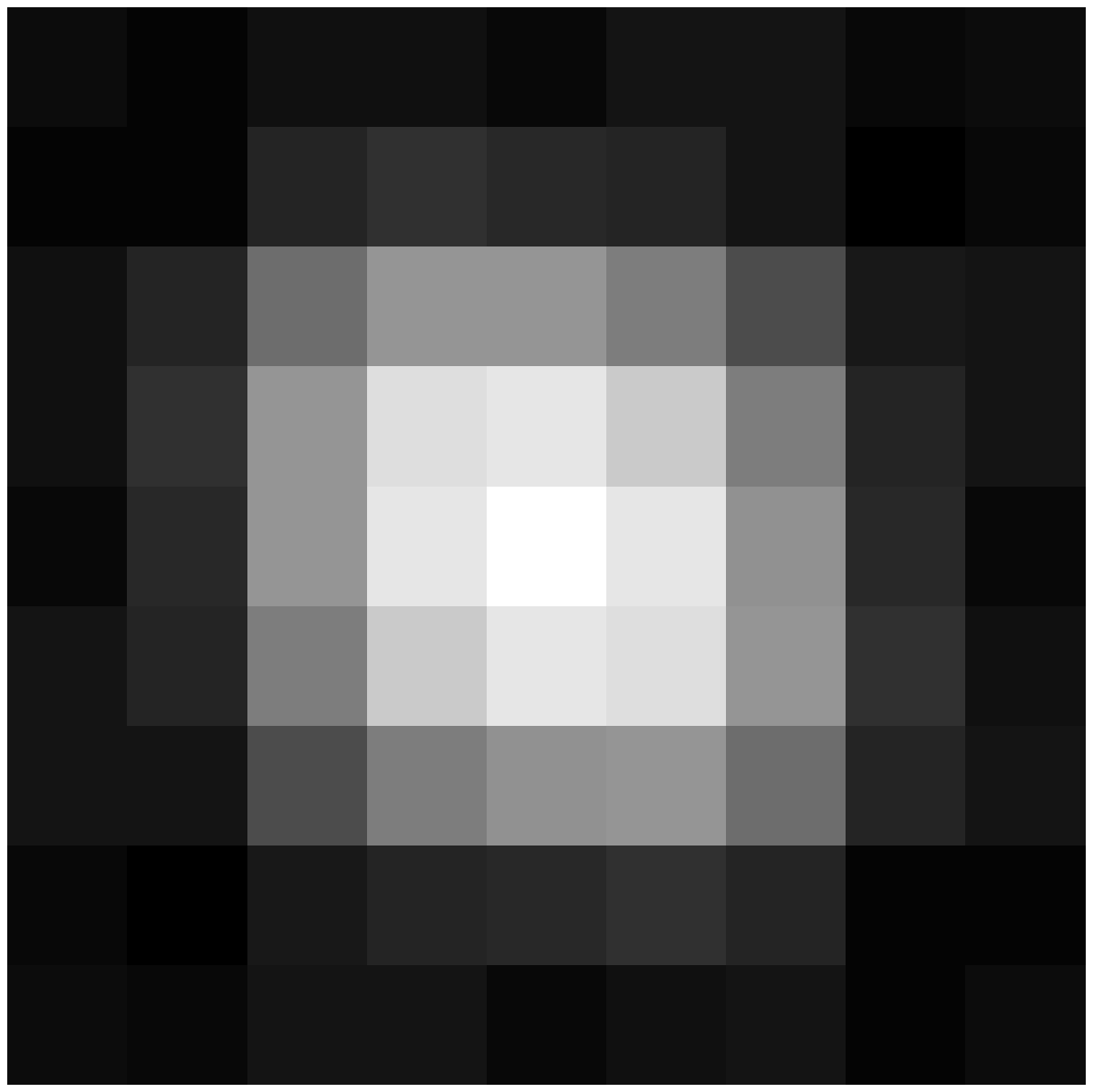}}
\end{overpic}&\hspace{-0.7cm}

\begin{overpic}
[height=3cm, trim = {1cm 1cm 1cm 1cm}, clip]{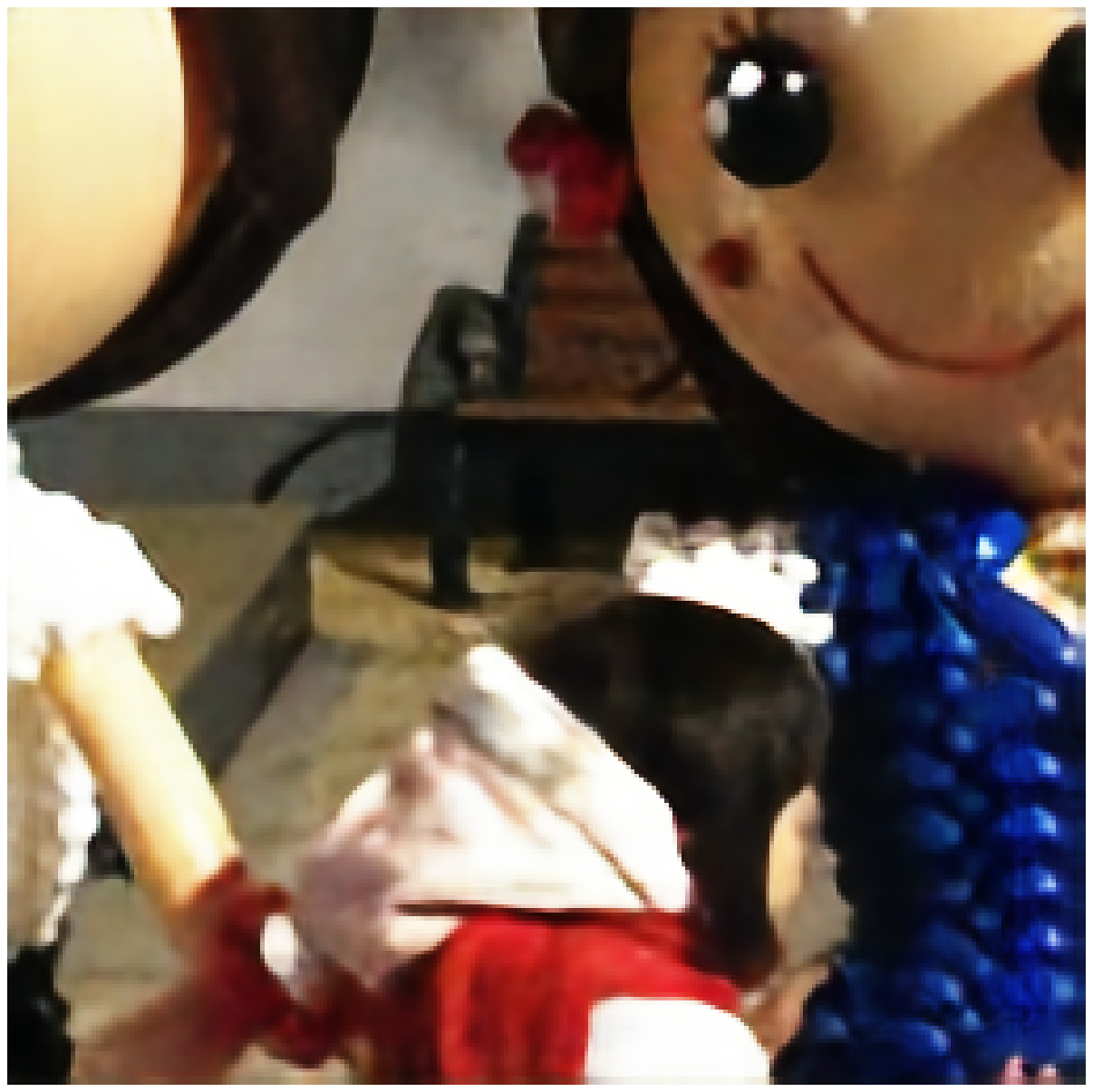}
\put(22,6)
{\includegraphics[height=1cm,trim={4cm 1.5cm 3.4cm 1cm},clip]{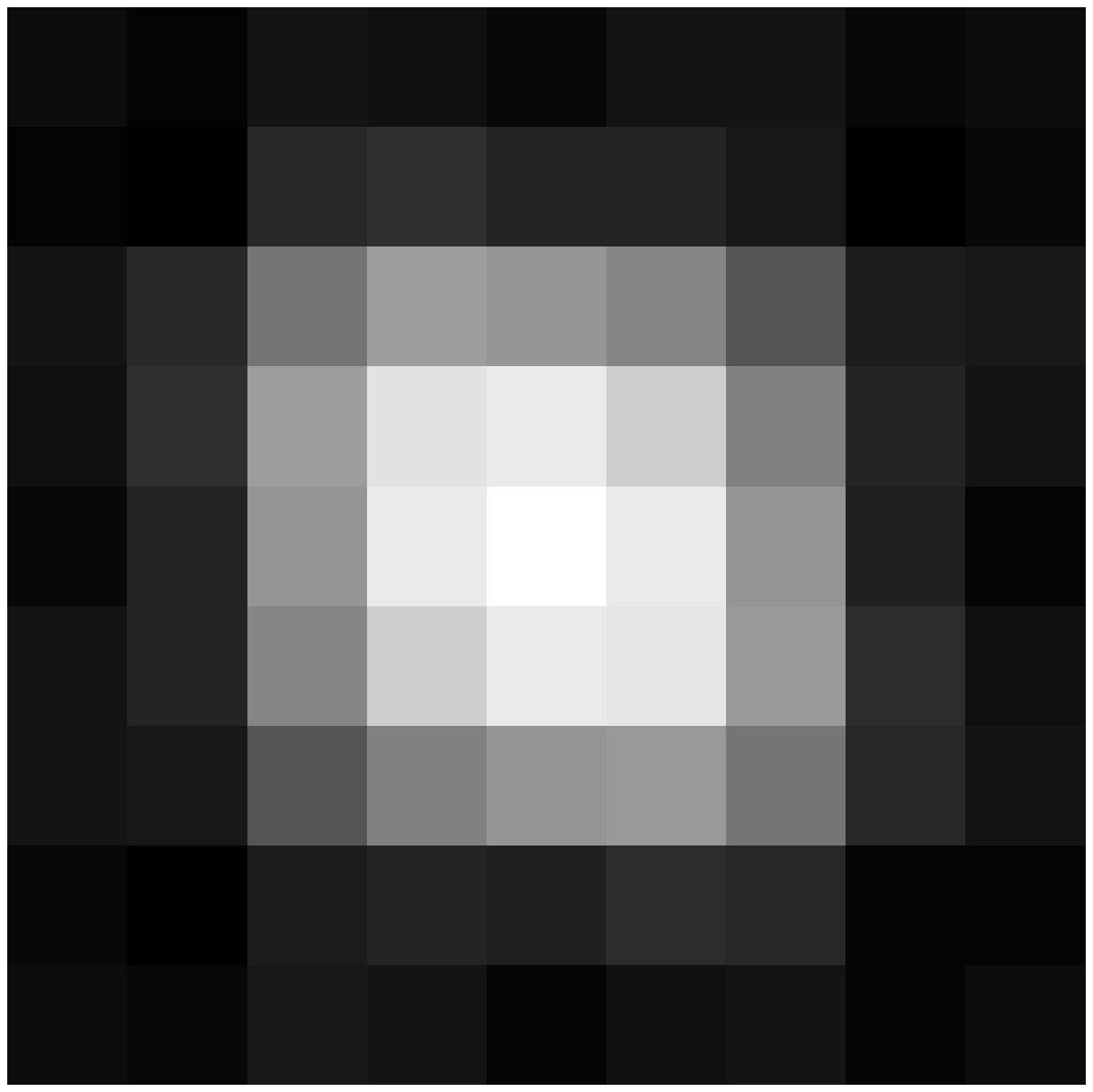}}
\end{overpic}\\

\textbf{SelfDeblur} & \textbf{DBSRCNN} & \textbf{DeblurGAN} & \textbf{proposed (greedy)} & \textbf{proposed (end-to-end)} \\
MSE = 0.9536& & &MSE = 0.0049&MSE = \textbf{0.0046} \\
PieAPP = 2.0619 &PieAPP = 0.8841&PieAPP = 2.7761&PieAPP = 0.7790&PieAPP = \textbf{0.6987}  \\ 

\end{tabular}
\caption{\footnotesize Ground-truth image/blur, degraded image, restored images (with PieAPP index) and estimated blurs (with MSE score) when available, for various methods, on two examples in the test set of \emph{Dataset 2}.}
\label{fig:Dataset2_case3}
\end{figure*}

\begin{figure}[t]
\centering
\begin{tabular}{@{}c@{}c@{}}
\includegraphics[height = 3cm]{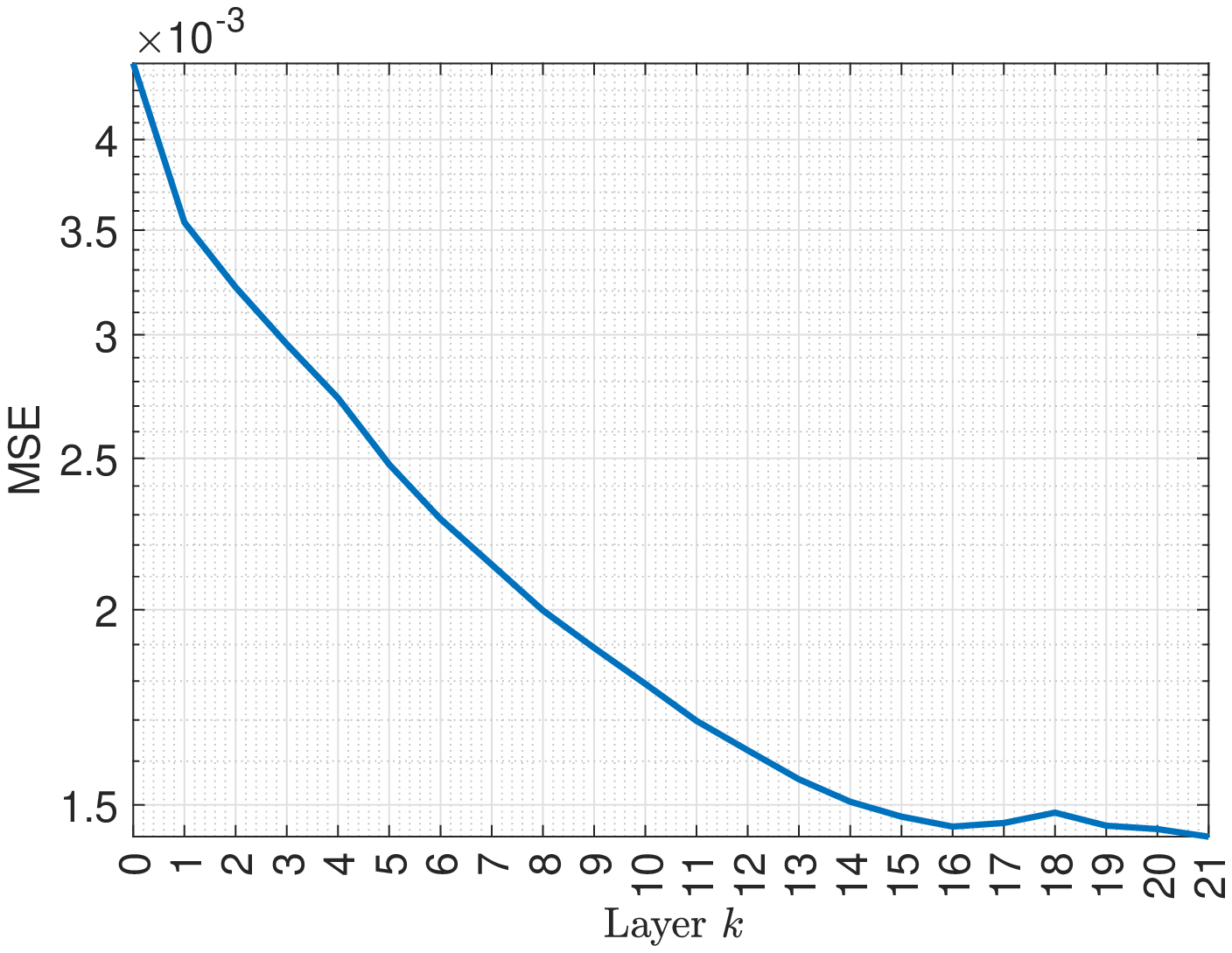} &
\includegraphics[height = 3cm]{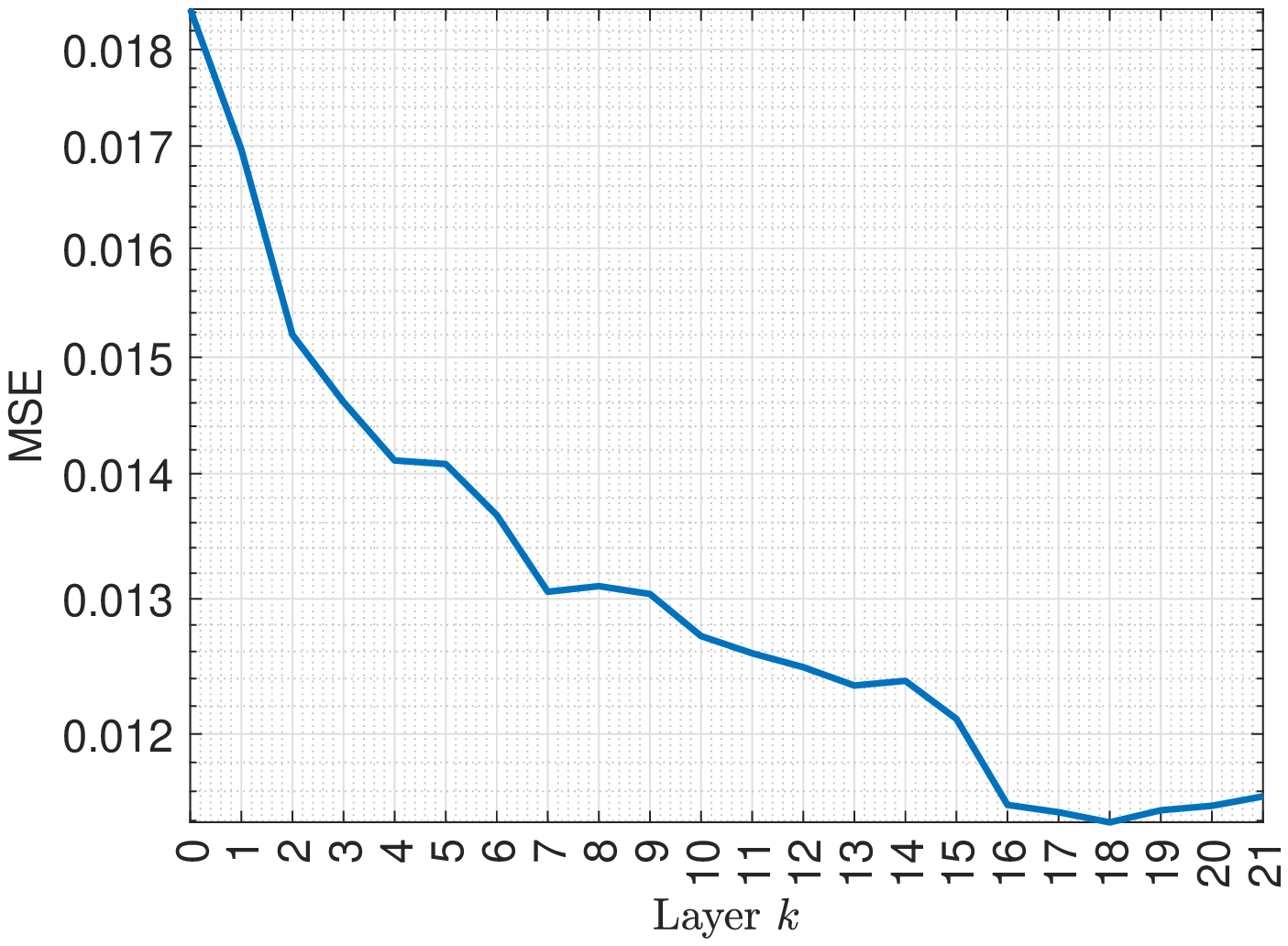}  
\end{tabular}
\caption{\footnotesize  MSE loss along the layers for proposed method using \emph{end-to-end training}. Average over \emph{Dataset 2} test examples involving either Gaussian (left), or out-of-focus (right) blur shapes.}
\label{fig:Gaussian_MSE}
\end{figure}

The results of kernel estimation and image restoration on Dataset 2 
using the various methods are shown in Tables~\ref{table:result_blur2} and \ref{table:result_image2}, respectively. This dataset is more challenging, as it includes color images, various blur shapes, and various noise levels. The latter are not assumed to be known anymore. Hereagain, we can observe that the \emph{greedy training} yields the best performance in terms of kernel estimation for the three considered metrics. In contrast, \emph{end-to-end training} tends to favor the restored image quality while still providing a good kernel quality compared to other methods. In this more complicated context, standard VBA does not perform very well, as setting $\xi$ becomes tedious for such an heterogeneous dataset. Let us note that the noise level is assumed to be known for this particular method, putting it in a quite favorable situation, compared to the other competitors, including our proposed approach. DBSRCNN provides again a good image recovery, but our proposed approach still outperforms it for both SSIM and PieAPP metrics. We display two examples of restoration in Fig~\ref{fig:Dataset2_case3}, when the sought blur is uniform, and out-of-focus, respectively. Such blur shapes are challenging and the MSE on the estimated blur might appear not excellent. Nevertheless, our method remains the best among the compared ones. The visual quality of the image generated by the proposed method is also very satisfying. We display in Fig.~\ref{fig:SSIM_loss}(right) the evolution of the SSIM loss during the \emph{end-to-end training}, witnessing the absence of any overfitting issue. Moreover, {we present in Fig~\ref{fig:Gaussian_MSE} the evolution of the MSE loss on the kernel estimate, along the $K = 21$ layers of the architecture trained in an \emph{end-to-end} manner.
The MSE was averaged on test set examples associated to either Gaussian or out-of-focus blurs, respectively.} These plots show that, for our choice of $K$ (finetuned on the validation set), the MSE values are close to minimal. Larger $K$ implied an increase of memory and training time, while not necessarily improving the results quality. One can also notice more fluctuations in the case of out-of-focus blur, which turns out to be more challenging to restore. A similar curve was obtained for uniform blurs, not shown by lack of space. Finally, Table~\ref{table:time1}(right) presents the average test time of the different methods. Again, our method appears competitive in terms of running time. 

%

\begin{table}
\scriptsize
\renewcommand{\arraystretch}{1.2}
\centering
\begin{tabular}{|c||c|c|c|c|}
\hline
Method & MSE &  $\mathcal{H}_{\infty}$ error & MAE\\
\hline
\hline
VBA &  0.0148 (0.0139)  & 0.4492 (0.1638)    & 0.1339 (0.0627)     \\
\hline
deconv2D & 0.0099 (0.0160) & 0.2796 (0.1692)  & 0.0869 (0.0576)  \\
\hline
blinddeconv& 0.0245 (0.0264)&0.3113 (0.1409) &0.1596 (0.1106) \\
\hline
SelfDeblur &1.7533 (1.4455)&2.5647 (2.7609)&1.3752 (0.5132) \\
\hline
Proposed (greedy) &\textbf{0.0037} (0.0079)&\textbf{0.1888} (0.1061)  &\textbf{0.0570} (0.0414)  \\
\hline
Proposed (end-to-end)& 0.0039 (0.0079)     &   0.1960 (0.1056) &  0.0588 (0.0411)   \\

\hline
\end{tabular}
\vspace{0.1cm}
\caption{\footnotesize Quantitative assessment of the restored kernels. Mean (standard deviation) values computed over the test sets of \emph{Dataset 2}.}
\label{table:result_blur2}
\end{table}

\begin{table}
\scriptsize
\renewcommand{\arraystretch}{1.2}
\centering
\begin{tabular}{|c||c|c|c|c|}
\hline
Method & SSIM & PSNR & PieAPP\\
\hline
\hline
Blurred & 0.5427 (0.1150)  &21.7994 (2.1679)& 4.2378 (0.8539) \\
\hline
VBA &0.4024 (0.1571) &16.0371 (4.1798)& 2.4218 (0.5545)\\
\hline
deconv2D &0.6880 (0.1065)&23.1940 (2.8986) & 2.2245 (0.6721) \\
\hline
blinddeconv& 0.6961 (0.1034) &23.2663 (2.7229)  &2.3259 (0.8080) \\
\hline
SelfDeblur &0.5107 (0.1305)&19.9943 (2.1467)&5.9269 (1.4066)  \\
\hline
DBSRCNN &0.6948 (0.1688)&\textbf{23.6041} (4.2073)&1.9474 (0.7171)\\
\hline
DeblurGAN &0.3370 (0.0740)&17.2781 (1.2909) &3.6581 (1.0040) \\
\hline
Proposed (greedy) &0.7454 (0.1015)  &23.2169 (2.4442)   &  \textbf{1.7250} (0.5324) \\
\hline
Proposed (end-to-end) & \textbf{0.7518} (0.1025) & \textbf{23.5631} (2.5959) &\textbf{1.7681} (0.5502)  \\
\hline
\end{tabular}
\vspace{0.1cm}
\caption{\footnotesize Quantitative assessment of the restored images. Mean (standard deviation) values computed over the test sets of \emph{Dataset 2}.}
\label{table:result_image2}
\end{table}

\section{Conclusion}
\label{sec:conclusion}
This paper proposes a novel method for blind image deconvolution that combines a variational Bayesian algorithm with a neural network architecture. Our experiments illustrate the excellent performance of this new method on two datasets, comprising grayscale and color images, and degraded with various kernel types. Compared to state-of-the-art variational and deep learning approaches, our method delivers a more accurate estimation of both the image and the blur kernels. It also includes an automatic noise estimation step, so that it requires little hyperparameter tuning. The proposed method is very competitive in terms of computational time during the test phase, while showing similar train time to its deep learning competitors. 
The main core of the proposed architecture is highly interpretable, as it implements unrolled iterates of a well sounded Bayesian-based blind deconvolution method. As a byproduct, it also outputs estimates for the covariance matrices of both sought quantities (image/kernel). This information could be of interest for uncertainty quantification and model selection tasks (see for instance \cite{Yunshi2021,Repetti2019}). More generally, our work demonstrates that unrolling VBA algorithms constitutes a promising research direction for solving challenging problems arising in Data Science.
\appendices

{\footnotesize
\bibliographystyle{ieeetr}
\bibliography{refs}
}


\end{document}